%% file: main.tex
\documentclass[sigconf,nonacm]{aamas}

\usepackage{balance}
\usepackage{pifont}
\usepackage{listings}
\usepackage{xcolor}
\usepackage{pgfplots}
\pgfplotsset{compat=1.18}
\usepgfplotslibrary{groupplots}
\usepackage{pgfplotstable}
\usepackage{longtable}
\usepackage{array}
\usepackage{placeins}
\usepackage{xurl}

\setcopyright{none}
\fancypagestyle{arxivplain}{\fancyhf{}\fancyfoot[C]{\thepage}}
\AtBeginDocument{\fancypagestyle{firstpagestyle}{\fancyhf{}\fancyfoot[C]{\thepage}}}

\title[MARL-BattleGrounds]{MARL-BattleGrounds: A Novel JAX-native Benchmark for Heterogeneous and Competitive Multi-agent RL}

\author{Ulixes Tariq Hawili}
\authornote{Corresponding author.}
\affiliation{
  \department{School of Engineering}
  \institution{University of Edinburgh}
  \city{Edinburgh}
  \country{United Kingdom}}
\email{t.hawili@sms.ed.ac.uk}
\author{Subramanian Ramamoorthy}
\affiliation{
  \department{School of Informatics}
  \institution{University of Edinburgh}
  \city{Edinburgh}
  \country{United Kingdom}}
\email{s.ramamoorthy@ed.ac.uk}
\author{Yoann Altmann}
\affiliation{
  \department{School of Engineering and Physical Sciences}
  \institution{Heriot-Watt University}
  \city{Edinburgh}
  \country{United Kingdom}}
\email{Y.Altmann@hw.ac.uk}
\author{Ignacio Carlucho}
\affiliation{
  \department{School of Engineering and Physical Sciences}
  \institution{Heriot-Watt University}
  \city{Edinburgh}
  \country{United Kingdom}}
\email{ignacio.carlucho@hw.ac.uk}

\input{sections/00_abstract}


\newif\ifshowtodos
\showtodostrue
\ifshowtodos
  \newcommand{\todo}[1]{\textcolor{red}{[TODO: #1]}}
\else
  \newcommand{\todo}[1]{??}
\fi

\newcommand{\weblink}[2]{\href{#1}{\textcolor{ACMDarkBlue}{#2}}}

\newcommand{\cmark}{\ding{51}}
\newcommand{\xmark}{\ding{55}}
\newcommand{\pmark}{$\circ$}

\newcommand{\bgs}{MARL-BGs}
\newcommand{\tdm}{Team Deathmatch}

\begin{document}


\maketitle
\pagestyle{arxivplain}

\input{sections/intro_IC}
\input{sections/02_related_work}
\input{sections/03_preliminaries}
\input{sections/04_game}
\input{sections/05_evaluation}
\input{sections/06_tools}
\input{sections/07_experiments}

\input{sections/08_limitations}
\input{sections/09_conclusion}


\emph{AI Disclosure:} This document was refined using AI assistance to optimize grammar, clarity, and structural flow. All content has been fully verified for accuracy and authenticity by the authors.

\begin{acks}
This work was supported by the Engineering and Physical Sciences Research Council and a UK government partner [EP/Y013859/1].
\end{acks}

\bibliographystyle{ACM-Reference-Format}
\bibliography{references}

\input{appendix}

\end{document}

%% file: sections/00_abstract.tex
\begin{abstract}
We present MARL-BattleGrounds (MARL-BGs), an easy-to-use JAX-native benchmark for heterogeneous and competitive multi-agent reinforcement learning (MARL), designed to be sample-efficient, computationally efficient, and researcher-centric. MARL benchmarks are typically simple and efficient, like SMAX, where winning reduces to unit micromanagement, or rich and costly, like Dota 2. MARL-BGs fills this missing middle. Its first mode, Team Deathmatch, is a simultaneous turn-based game between two teams of five heterogeneous agents. The five classes have distinct basic, passive, and ultimate abilities designed to synergize with allies and counter enemies. Our 52 hand-designed maps with diverse obstacles include a 12-map curriculum and ten held-out validation and test maps. MARL-BGs provides two canonical evaluation modes. First, eight endgame scenarios form a hard generalization challenge: each asks a team to win an unseen, nearly lost position against a strong reactive controller within 5--10 ticks, testing agent modeling, multi-step planning, and more. Second, a monthly round-robin tournament retrains each submission on equal compute, plays it against every member of the current ``Big Nine'' field, and publishes the Elo results on the official leaderboard. All entrants are publicly released as downloadable opponents. To help researchers train and evaluate their policies, MARL-BGs records 11,192 metrics per game in 28 topics, from damage and healing to team coordination and formation. We also include three strong scripted opponents, an analytical replay viewer, and off-the-shelf tools for self-play, cross-play, curriculum learning, and zero-shot coordination evaluation. We even provide an API for LLM-based agents and use it to evaluate Qwen3.5-9B against our strongest policy. In our BattleClient, human researchers can play alongside or against any mix of policies, enabling human-AI evaluation. We benchmark MARL-BGs with six additional deep MARL baselines, which we release. On one RTX~5090 with 1,024 parallel games, the simulator runs 50,000 5v5 game steps per second. To our knowledge, no other MARL benchmark offers this breadth and depth of research and evaluation tools.\footnote{Everything there is to know about MARL-BGs is available on our website (\weblink{https://marl-battlegrounds.com}{marl-battlegrounds.com}), including interactive demos and comprehensive tutorials, and the full GitHub repository and Hugging Face models will be made public upon manuscript acceptance.}
\end{abstract}

%% file: sections/intro_IC.tex
\section{Introduction}
\label{sec:introduction}

Many multi-agent reinforcement learning (MARL) benchmarks sit at two extremes. At one end are fast and simple benchmarks such as SMAX~\cite{rutherford2024jaxmarl}, which runs StarCraft-style unit battles on the GPU in JAX, but its units only move and shoot and differ only in stats such as health, speed, and range, so winning reduces to low-level unit micromanagement. At the other end are full games such as Dota~2. Dota~2 demands deep coordination among specialist agents, and it took OpenAI Five ten months of training on large GPU and CPU clusters to reach the final agent performance~\cite{berner2019dota}. Few research groups can run experiments at that scale, and iterating over results makes improvements difficult. That leaves a missing middle.

The middle matters because central questions in MARL need both rich team play and fast, cheap experiments: how agents with different abilities share what they learn, how a team assigns credit for a heal or a stun that pays off only through a teammate, how teammates time their actions together, and whether the resulting policies still work with opponents and teammates they never met. Evaluation has three gaps of its own. Fixed tasks get solved: by 2021, the best published win rates on most popular SMAC~\cite{samvelyan2019smac} maps were close to 100\%~\cite{gorsane2022standardised}, and SMACv2 found that policies which ignore their observations still win many SMAC tasks~\cite{ellis2023smacv2}. Results are hard to compare: re-analyzed~\cite{gorsane2022standardised} on 6 of SMAC's 14 maps, a common choice, QMIX~\cite{rashid2018qmix} lost its lead over VDN~\cite{sunehag2018vdn}, and in Google Research Football an agent trained against a frozen earlier agent beat it but played poorly against the built-in bot~\cite{kurach2020grf}, so one fixed opponent misleads. And a score hides what a team learned: in Overcooked, self-play agents worked well with each other but worse with humans, and watching their games showed why: they used one of the two pots, while humans used both~\cite{carroll2019overcooked}.

We present MARL-BattleGrounds (\bgs{}), an easy-to-use JAX-native benchmark for heterogeneous and competitive team-vs-team MARL that fills the missing middle. MARL-BGs was built on our own engine and designed to be sample-efficient, computationally efficient, and researcher-centric. Its three parts are our contributions: i) \emph{The Game}, ii) \emph{The Evaluation}, and iii) \emph{The Tools}.

\textbf{i) The Game}, \tdm{}, demands high-level team coordination from teams of up to five agents (Section~\ref{sec:game}).
We introduce five classes, inspired by World of Warcraft, with hand-designed abilities that require agents to synergize with allies and counter enemies (Figure~\ref{fig:classes}). The game has 52 maps with obstacles, and partial observability, requiring tight coordination from agents. 

\textbf{ii) The Evaluation} answers the three gaps (Section~\ref{sec:evaluation}). We provide a monthly tournament where researchers can upload their policies and compete against other MARL policies. 
Every method is rated with draw-aware Elo~\cite{elo1978rating,davidson1970ties} and confidence intervals on the same held-out test maps, under the same budget, in mirrored 5v5 games with spawn ends swapped, so results are comparable. The leaderboard also provides other researchers with harder policies to train against, improving the community's ability to progress. 
Eight hand-crafted endgame scenarios, each with a verified winning line, test skills such as agent modeling and multi-step planning, so a failed scenario points to a missing skill.
Behind every result, up to 11,192 metrics per game in 28 topics (Figure~\ref{fig:metrics}) allow us to analyze the teamwork, such as how much the Priest healed each ally, so methods can be judged by their coordination, not only by their win rates.

\textbf{iii) The Tools} cover the whole research workflow (Section~\ref{sec:tools}). Researchers bring their own learner through plain JAX \texttt{reset} and \texttt{step} or start from six tuned baselines, and one call evaluates any mix of policies, so cross-play and zero-shot coordination tests come out of the box.
We provide three strong scripted opponents and all checkpoints of our baselines, providing researchers with opponents of all strengths.  
Our LLM API turns any language model into a team of agents, enabling the evaluation of foundation models. In the BattleClient, researchers play alongside or against their policies for human-AI teaming evaluation, and in the analytical Replay Viewer they step through any game tick by tick. 

We demonstrate every part of \bgs{} on Season~0, its first tournament (Section~\ref{sec:experiments}). Each PPO baseline trains at about a million game steps per minute on one RTX~5090 with 512 parallel games. Recurrent IPPO, our strongest baseline, loses to all three scripted opponents at 40 million steps but wins 579 of 600 games against them at 250 million, on the held-out test maps (Figure~\ref{fig:elo}). Our results demonstrate that stronger teams combine their abilities, yet no system solves a scenario, an LLM team plays legally but not as a team, and learned agents partner poorly with strangers and with an expert human, who also beats the tournament winner. These open problems are exactly what \bgs{} is built to measure.

%% file: sections/02_related_work.tex

\section{Related Work}
\label{sec:related}

\input{tables/comparison}

Table~\ref{tab:comparison} compares \bgs{} with sixteen MARL benchmarks and games on twenty-two features of the game, the evaluation, and the research tools, graded from each benchmark's official releases. 
%
Five of the sixteen have JAX or GPU-native simulators: SMAX in JaxMARL~\cite{rutherford2024jaxmarl}, Gigastep~\cite{lechner2023gigastep}, Lux AI Season~3~\cite{tao2024luxs3}, VMAS~\cite{bettini2022vmas}, and OvercookedV2~\cite{gessler2025overcookedv2}. In all five, agents differ at most in stats or task roles, with no class abilities such as healing or crowd control. Only Honor of Kings Arena~\cite{wei2022hok}, Mini Honor of Kings~\cite{liu2025minihok}, and Dota~2~\cite{berner2019dota} give every hero its own abilities, and all three run on the CPU. The Honor of Kings Arena paper trained with 128 to 2,048 CPU cores; Mini Honor of Kings is light but not competitive, as its heroes fight a rule-controlled dragon. SMAC~\cite{samvelyan2019smac}, SMACv2~\cite{ellis2023smacv2}, Google Research Football~\cite{kurach2020grf}, MAgent2~\cite{zheng2018magent}, and MPE~\cite{mordatch2018emergence,lowe2017maddpg} also run on the CPU, and only Honor of Kings Arena and Dota~2 have full cover that blocks sight.

Statistical standards for reporting RL results exist~\cite{gorsane2022standardised,agarwal2021precipice}, and Elo-style ratings against fixed opponents are common in competitive benchmarks such as Gigastep and Pgx~\cite{koyamada2023pgx}, yet no benchmark ships a rating tool with confidence intervals. Five ship scripted opponents that some learned baselines fail to beat, such as SMAC's built-in StarCraft~II AI, and six release trained agents, such as Melting Pot~2.0's bots. Official competitions, such as the Lux AI challenges on Kaggle, mostly ran once or for a few years; only Honor of Kings Arena still holds a yearly one. None of the sixteen holds out evaluation maps, and verified scenarios are rare: give-way is VMAS's only fixed, designed situation with a tested solution. Dota~2 logs who attacked whom, but no benchmark turns such logs into a full set of team metrics.

Baseline learners and replay viewers are common (fourteen and nine of the sixteen); reward-shaping and curriculum tools are rare (three and two). Partner tests are the strength of coordination suites such as Overcooked-AI~\cite{carroll2019overcooked}, Melting Pot~2.0~\cite{agapiou2022meltingpot2}, and Hanabi~\cite{bard2020hanabi}, but only Overcooked-AI and Melting Pot~2.0 ship full zero-shot coordination tools, and all but Melting Pot~2.0 are purely cooperative. Four benchmarks ship a client for human play. None has a full interface for LLM agents, and none lets researchers fill each team slot with any mix of learned, scripted, LLM, hybrid, and human players without writing their own code.

No prior benchmark combines heterogeneous team-vs-team play in a fast JAX simulator with held-out maps, verified scenarios, a recurring rated tournament, and detailed teamwork metrics.

%% file: tables/comparison.tex
\begin{table*}[t]
\caption{Features of MARL benchmarks, grouped by family: \cmark{} yes, \pmark{} partial, \xmark{} no. Section~\ref{supp:s2} of the supplementary material gives the rule and sources for every mark. $^\dagger$Committed for the public release.}
\label{tab:comparison}
\centering\small
\resizebox{\textwidth}{!}{%
\begin{tabular}{lccccccccccccccccc}
\toprule
 & \multicolumn{6}{c}{\textit{Lightweight suites}} & \multicolumn{6}{c}{\textit{Games}} & \multicolumn{4}{c}{\textit{Coordination suites}} &  \\
\cmidrule(lr){2-7} \cmidrule(lr){8-13} \cmidrule(lr){14-17}
Feature & \rotatebox{70}{SMAX~\cite{rutherford2024jaxmarl}} & \rotatebox{70}{Gigastep~\cite{lechner2023gigastep}} & \rotatebox{70}{Lux S3~\cite{tao2024luxs3}} & \rotatebox{70}{VMAS~\cite{bettini2022vmas}} & \rotatebox{70}{MAgent2~\cite{zheng2018magent}} & \rotatebox{70}{MPE~\cite{mordatch2018emergence,lowe2017maddpg}} & \rotatebox{70}{SMAC~\cite{samvelyan2019smac}} & \rotatebox{70}{SMACv2~\cite{ellis2023smacv2}} & \rotatebox{70}{GRF~\cite{kurach2020grf}} & \rotatebox{70}{HoK Arena~\cite{wei2022hok}} & \rotatebox{70}{Mini HoK~\cite{liu2025minihok}} & \rotatebox{70}{Dota 2~\cite{berner2019dota}} & \rotatebox{70}{Overcooked~\cite{carroll2019overcooked}} & \rotatebox{70}{OvercookedV2~\cite{gessler2025overcookedv2}} & \rotatebox{70}{MP 2.0~\cite{agapiou2022meltingpot2}} & \rotatebox{70}{Hanabi~\cite{bard2020hanabi}} & \rotatebox{70}{\textbf{\bgs{}}} \\
\midrule
\multicolumn{18}{l}{\textit{Game}} \\
JAX / GPU-native simulator & \cmark & \cmark & \cmark & \cmark & \xmark & \xmark & \xmark & \xmark & \xmark & \xmark & \xmark & \xmark & \xmark & \cmark & \xmark & \xmark & \cmark \\
Classes with distinct abilities & \pmark & \pmark & \xmark & \pmark & \pmark & \pmark & \pmark & \pmark & \pmark & \cmark & \cmark & \cmark & \xmark & \xmark & \pmark & \xmark & \cmark \\
Healing, auras, crowd control, ultimates & \xmark & \xmark & \pmark & \xmark & \xmark & \xmark & \pmark & \pmark & \pmark & \cmark & \cmark & \cmark & \xmark & \xmark & \pmark & \xmark & \cmark \\
Line of sight and obstacle cover & \xmark & \pmark & \pmark & \pmark & \pmark & \pmark & \pmark & \xmark & \xmark & \cmark & \xmark & \cmark & \pmark & \pmark & \pmark & \xmark & \cmark \\
Body blocking & \cmark & \xmark & \xmark & \pmark & \pmark & \pmark & \cmark & \cmark & \cmark & \pmark & \xmark & \cmark & \cmark & \cmark & \cmark & \xmark & \cmark \\
\addlinespace
\multicolumn{18}{l}{\textit{Evaluation}} \\
Built-in train/validation/test maps & \xmark & \xmark & \xmark & \xmark & \xmark & \xmark & \xmark & \xmark & \xmark & \xmark & \xmark & \xmark & \xmark & \xmark & \xmark & \xmark & \cmark \\
Scenarios with verified solutions & \pmark & \xmark & \xmark & \pmark & \xmark & \xmark & \pmark & \xmark & \pmark & \pmark & \pmark & \xmark & \xmark & \xmark & \xmark & \xmark & \cmark \\
Strong scripted opponents & \cmark & \pmark & \pmark & \pmark & \pmark & \xmark & \cmark & \cmark & \cmark & \cmark & \pmark & \pmark & \xmark & \xmark & \pmark & \xmark & \cmark \\
Tournament ratings with confidence intervals & \xmark & \pmark & \pmark & \xmark & \xmark & \xmark & \xmark & \xmark & \pmark & \pmark & \xmark & \pmark & \xmark & \xmark & \xmark & \xmark & \cmark \\
Recurring official tournament & \xmark & \xmark & \xmark & \xmark & \xmark & \xmark & \xmark & \xmark & \xmark & \cmark & \xmark & \xmark & \xmark & \xmark & \xmark & \xmark & \cmark$^\dagger$ \\
Released agent population & \xmark & \cmark & \xmark & \xmark & \cmark & \xmark & \xmark & \cmark & \xmark & \cmark & \xmark & \xmark & \cmark & \xmark & \cmark & \xmark & \cmark$^\dagger$ \\
Who-did-what-to-whom metrics & \xmark & \xmark & \xmark & \xmark & \xmark & \xmark & \xmark & \xmark & \xmark & \xmark & \xmark & \pmark & \xmark & \xmark & \pmark & \pmark & \cmark \\
\addlinespace
\multicolumn{18}{l}{\textit{Research tools}} \\
Built-in baseline learners & \cmark & \cmark & \xmark & \cmark & \cmark & \cmark & \cmark & \cmark & \cmark & \cmark & \cmark & \xmark & \cmark & \cmark & \cmark & \cmark & \cmark \\
Built-in curriculum tools & \pmark & \pmark & \xmark & \pmark & \xmark & \cmark & \pmark & \pmark & \pmark & \pmark & \pmark & \pmark & \cmark & \pmark & \pmark & \pmark & \cmark \\
Built-in reward shaping tools & \pmark & \pmark & \xmark & \cmark & \cmark & \cmark & \pmark & \pmark & \pmark & \pmark & \xmark & \xmark & \pmark & \pmark & \pmark & \xmark & \cmark \\
Built-in competitive self-play tools & \pmark & \cmark & \pmark & \cmark & \cmark & \cmark & \xmark & \xmark & \pmark & \cmark & \xmark & \pmark & \xmark & \xmark & \cmark & \xmark & \cmark \\
Built-in cross-play tools & \pmark & \cmark & \cmark & \pmark & \pmark & \pmark & \pmark & \pmark & \cmark & \cmark & \pmark & \pmark & \cmark & \cmark & \cmark & \pmark & \cmark \\
Built-in zero-shot coordination tools & \xmark & \pmark & \xmark & \xmark & \pmark & \xmark & \xmark & \pmark & \pmark & \pmark & \xmark & \xmark & \cmark & \pmark & \cmark & \xmark & \cmark \\
Built-in LLM interface & \xmark & \xmark & \xmark & \xmark & \xmark & \xmark & \xmark & \xmark & \xmark & \xmark & \xmark & \xmark & \pmark & \xmark & \xmark & \pmark & \cmark \\
Human play client & \xmark & \cmark & \xmark & \pmark & \pmark & \pmark & \xmark & \xmark & \cmark & \xmark & \xmark & \cmark & \cmark & \pmark & \pmark & \xmark & \cmark \\
Replay viewer & \pmark & \pmark & \cmark & \pmark & \cmark & \pmark & \cmark & \cmark & \cmark & \cmark & \cmark & \cmark & \cmark & \pmark & \pmark & \xmark & \cmark \\
Built-in per-slot mixed teams & \pmark & \pmark & \pmark & \pmark & \pmark & \pmark & \pmark & \pmark & \pmark & \pmark & \pmark & \pmark & \pmark & \pmark & \pmark & \pmark & \cmark \\
\bottomrule
\end{tabular}}
\end{table*}

%% file: sections/03_preliminaries.tex

\section{Preliminaries}
\label{sec:preliminaries}

\tdm{} is a two-team, zero-sum, partially observable stochastic game~\cite{hansen2004posg}. Teams $A$ and $B$ have up to five agents each. At each tick $t$, every agent $i$ receives an observation $o^i_t = O^i(s_t)$ of the state $s_t$ and chooses a legal action $a^i_t \in \mathcal{A}^i(s_t)$; all agents act at once, and the tick resolves to $s_{t+1} = P(s_t, \boldsymbol{a}_t)$, where $\boldsymbol{a}_t$ is the joint action. $P$, every $O^i$, and resets are deterministic, so all randomness comes from the policies and from any random choice of map, roster, or spawn.
A team wins when it reaches 20 points with more points than its opponent. Equal scores at 20, and games still open after 300 ticks, are draws, so a team cannot win by taking the lead and then hiding; it must keep fighting. Rewards are sparse: when the game ends, team $A$ receives $r_A = +1$ for a win, $-1$ for a loss, and $0$ for a draw, and team $B$ receives $-r_A$. So team $A$'s expected reward is its win rate minus its loss rate.

Each action $a^i$ has three heads: a move (stay, or one of 8 compass directions), a target (none, or one of the 5 own-team or 5 enemy slots) that receives the agent's Basic ability or a targeted Ultimate, and Ultimate (none, or use). Exact legal-action masks define each agent's legal set $\mathcal{A}^i(s_t)$. Our baselines flatten the three heads into $9 \times 11 \times 2 = 198$ joint choices.

Each agent's action may depend only on its own history $h^i_t = (o^i_0, a^i_0, m^{-i}_0, \dots, o^i_t)$, through a policy $\pi^i(a^i_t \mid h^i_t)$, where $m^{-i}_k$ holds any messages its teammates sent on tick $k$; a message arrives one tick after it is sent. By default, $O^i$ returns observations within agent $i$'s own sight radius and line of sight, plus the sensor rows of its living teammates. Inputs describe other agents by role (self, ally, enemy), never by team identity. A centralized critic may use the full state during training only; actors never do.

A \emph{tick} is one simultaneous decision of both teams. A \emph{game step} is one tick of one game; in 5v5, one game step contains 10 \emph{agent steps}. Training budgets count game steps. A team's \emph{point margin} is its points minus its opponent's. A \emph{System} is any method that returns a team's actions, and so defines a team policy; for example, a neural network, a scripted controller, an LLM, or a hybrid. Each teammate can be a different System.

%% file: sections/04_game.tex

\section{The Game: Team Deathmatch}
\label{sec:game}

\subsection{Design Goals}
\label{sec:game-design}

Our main design goals were: 
\begin{itemize}
    \item \emph{Sample efficiency:} exact action masks remove illegal choices, and the factored action space stays small. Two optional built-in curricula start on smaller or shorter games, then grow the teams and add obstacles, or raise the winning score.
    \item \emph{Computational efficiency:} the simulator is pure JAX with fixed array shapes, and finished games reset without touching the others. Changing ordinary values, such as the map or the roster, does not force a new compilation, making training exceptionally fast (Section~\ref{sec:exp-speed}). 
\item \emph{Researcher centricity:} a familiar API (Section~\ref{sec:game-api}) and optional ready-made tools for the whole workflow (Section~\ref{sec:tools}).
\end{itemize}




\subsection{Rules}
\label{sec:game-rules}

\begin{figure}[t]
  \centering
  \includegraphics[width=\columnwidth]{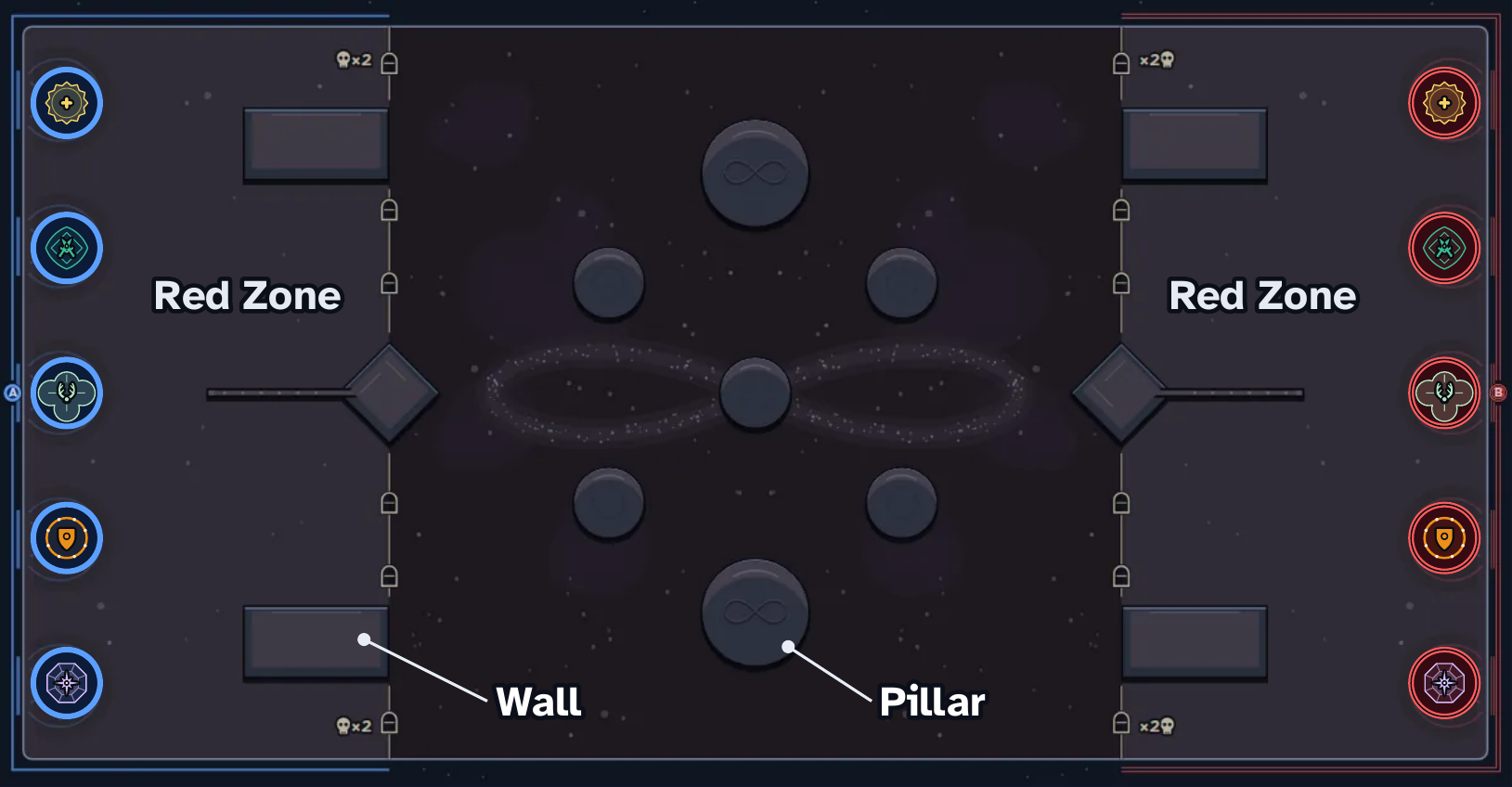}
  \caption{Training map 34, Infinity, with the official tournament's mirrored 5v5 on the spawn pads, from bottom to top: Mage, Warrior, Hunter, Rogue, and Priest. Watch real games and try our interactive demos at \weblink{https://marl-battlegrounds.com}{marl-battlegrounds.com}.}
  \label{fig:arena}
  \Description{A top-down drawing of a 20 by 10 unit map. Team A's five agents stand on spawn pads along the left edge and Team B's along the right edge. Each team's Red Zone, the strip at its own edge, is outlined and labeled. Labels point to a wall and a pillar.}
\end{figure}

\tdm{} is inspired by World of Warcraft's classic 5v5 arena. Each team has one to five agents of any classes, and the two teams may differ in size. The official tournament uses a mirrored 5v5 with one of each class (Section~\ref{sec:eval-tournament}).
Each kill gives the killing team one point, or two if the victim dies inside its own \emph{Red Zone}, a 5-unit-deep strip at its team's spawn end, so hiding at home is costly (Figure~\ref{fig:arena} shows a map with both Red Zones and the tournament's 5v5 on its spawn pads).
Obstacles block sight and attacks, and bodies block movement. Dead agents return in respawn waves every 5 ticks, with a 3-tick spawn shield against spawn camping: enemies cannot see or target them. Section~\ref{supp:s3.1} of the supplementary material gives the complete rules.


\subsection{Five Classes}
\label{sec:game-classes}

Each class has one Basic ability, one passive, and one Ultimate with a 30-tick cooldown. Every class is hand-designed for teamwork: each has a clear strength and a clear weakness, relies on its allies to cover that weakness, and has abilities that synergize with allies and counter enemies (Figure~\ref{fig:classes}; Section~\ref{supp:s3.2} of the supplementary material lists every class value): 
\begin{itemize}
\item  The \textbf{Mage} (80 HP) is the \emph{Glass-Cannon Artillery}: a ranged class with the highest Basic damage but the lowest health, so its team must keep it safe. Its passive aura raises its own and nearby allies' damage by 15\%, and its Ultimate, \emph{Burst}, raises its own damage by 50\% for 5 ticks.
\item  The \textbf{Warrior} (200 HP) is the \emph{Tank}: a melee class with the most health but a weak Basic, so it needs allies to finish fights. Its passive aura cuts the damage taken by itself and nearby allies by 15\%, and its Ultimate, \emph{Charge}, rushes it to a distant enemy, stuns them for 1 tick, and slows them for 5.
\item The \textbf{Hunter} (100 HP) is the \emph{Scout and Crowd Controller}: a ranged class with the longest reach and the widest sight but the weakest Basic attack, so it spots and slows enemies for its teammates to kill. Its Ultimate, \emph{Trap}, is the longest stun in the game, 4 ticks, but any damage breaks it, so teammates must hold their fire.
\item  The \textbf{Rogue} (100 HP) is the \emph{Assassin}: a melee class that moves 30\% faster than the others but has no dash, so it relies on teammates to engage first, letting it flank the enemy's backline. Its Ultimate, \emph{Poison}, is the biggest single hit in the game; it also stuns for 1 tick, slows for 5 ticks, and halves the healing its target receives for 4 ticks.
\item  The \textbf{Priest} (100 HP) is the \emph{Healer}: a ranged class that is the only healer but cannot deal damage, so its team must protect it while it keeps them alive. Each Basic heal briefly protects its target from strong slows, and its Ultimate, \emph{Salvation}, heals one ally to full.
\end{itemize}

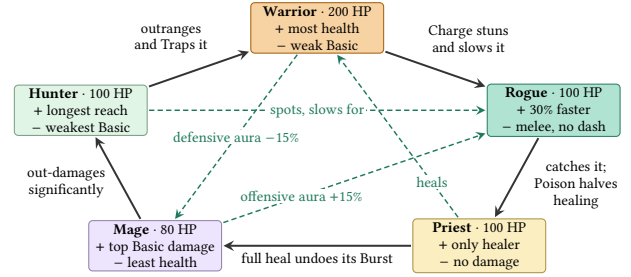
\begin{figure}[t]
  \centering
  \input{figures/classes}
  \caption{The five classes, each with its health, strength ($+$), and weakness ($-$). Solid arrows point from a class to the enemy class it counters; dashed arrows point to an ally it helps. Not every interaction is drawn to reduce clutter.}
  \label{fig:classes}
  \Description{Five boxes, one per class, arranged in a pentagon: Warrior at the top, Rogue at the right, Priest at the bottom right, Mage at the bottom left, and Hunter at the left. Each box gives the class's health, one strength, and one weakness. Solid arrows around the outside form a cycle of counters: the Warrior counters the Rogue (Charge stuns and slows it), the Rogue counters the Priest (it catches the Priest, and Poison halves healing), the Priest counters the Mage (a full heal undoes its Burst), the Mage counters the Hunter (it out-damages the Hunter at medium range), and the Hunter counters the Warrior (it outranges and Traps the Warrior). Dashed arrows inside show help between allies: the Warrior's aura cuts the Mage's damage taken by 15 percent, the Mage's aura raises the Rogue's damage by 15 percent, the Hunter spots and slows enemies for the Rogue, and the Priest heals the Warrior.}
\end{figure}

\subsection{Maps}
\label{sec:game-maps}

\bgs{} has 52 hand-designed maps, each a $20 \times 10$ arena with 0 to 32 obstacles: round pillars and straight or diagonal walls of different thicknesses. Every map's left and right halves mirror each other exactly, so both spawn ends start from the same geometry. Maps 0--41 are for training, and maps 0--11 form a curriculum that introduces, step by step, every kind of obstacle. We have held out maps 42--46 for validation, and maps 47--51 for testing. That way no policy is tested on a map it was trained or chosen on (every map in Section~\ref{supp:s3.3} of the supplementary material).

\subsection{API}
\label{sec:game-api}

We designed the API to be simple, obvious, and familiar by comparing our full research workflows with those of JaxMARL~\cite{rutherford2024jaxmarl}, whose interface is in turn inspired by PettingZoo~\cite{terry2021pettingzoo} and gymnax~\cite{lange2022gymnax}. \texttt{make}, \texttt{reset}, and \texttt{step} take and return the same things as in JaxMARL, and the game state is an explicit JAX tree with no hidden state or random-number generator, so the full game loop compiles under \texttt{jit}, \texttt{vmap}, and \texttt{lax.scan}: running 1,024 games in parallel with random legal actions, with finished games restarting in place, takes 13 lines of code (Listing~\ref{lst:supp-api} in Section~\ref{supp:s4} of the supplementary material, which also describes the engine and how we test it). On this familiar core, \bgs{} provides what a team-vs-team benchmark needs: batching through a single argument, \texttt{num\_envs}; \texttt{reset\_done}, which restarts only finished games; exact action masks and a legal random sampler; per-team inputs that hold only what each agent may see, plus the full state for a critic used only in training; and full metrics and replays recorded only for the games a researcher picks. Any method becomes a System through one wrapper, as a JAX function compiled with the game or a Python function run on the host, with no base class, model framework, or learner of ours required. The command line calls the same functions as Python, and thirty example scripts show it in use.


%% file: figures/classes.tex
\definecolor{clsMage}{HTML}{D2BCFA}\definecolor{clsWarrior}{HTML}{EB9019}%
\definecolor{clsHunter}{HTML}{B4E6C1}\definecolor{clsRogue}{HTML}{3DB892}%
\definecolor{clsPriest}{HTML}{F5D565}\definecolor{synergy}{HTML}{1E7A5E}%
\begin{tikzpicture}[x=1pt,y=1pt,font=\fontsize{5.6}{6.3}\selectfont,
  cls/.style={draw=#1!70!black, fill=#1!40, rounded corners=1.5pt, line width=0.4pt,
    inner sep=1.3pt, align=center, minimum width=50pt},
  counter/.style={-stealth, line width=0.75pt, black!80, shorten <=0.5pt, shorten >=0.5pt},
  helps/.style={-stealth, line width=0.55pt, synergy, dash pattern=on 2pt off 1.2pt,
    shorten <=0.5pt, shorten >=0.5pt},
  why/.style={inner sep=0.6pt, align=center},
  gift/.style={inner sep=0.6pt, fill=white, text=synergy, align=center}]
\node[cls=clsWarrior] (W) at (119,95) {\textbf{Warrior} \textperiodcentered{} 200 HP\\$+$ most health\\$-$ weak Basic};
\node[cls=clsRogue]   (R) at (208,64) {\textbf{Rogue} \textperiodcentered{} 100 HP\\$+$ 30\% faster\\$-$ melee, no dash};
\node[cls=clsPriest]  (P) at (180,13) {\textbf{Priest} \textperiodcentered{} 100 HP\\$+$ only healer\\$-$ no damage};
\node[cls=clsMage]    (M) at (58,13)  {\textbf{Mage} \textperiodcentered{} 80 HP\\$+$ top Basic damage\\$-$ least health};
\node[cls=clsHunter]  (H) at (30,64)  {\textbf{Hunter} \textperiodcentered{} 100 HP\\$+$ longest reach\\$-$ weakest Basic};
\draw[counter] (H) -- (W);
\node[why] at (64,91) {outranges\\and Traps it};
\draw[counter] (W) -- (R);
\node[why] at (176,91) {Charge stuns\\and slows it};
\draw[counter] (R) -- (P);
\node[why, anchor=west] at (200,36) {catches it;\\Poison halves\\healing};
\draw[counter] (P) -- (M);
\node[why, anchor=north] at (119,11) {full heal undoes its Burst};
\draw[counter] (M) -- (H);
\node[why, anchor=east] at (40,37) {out-damages\\significantly};
\draw[helps] (W) -- node[gift, pos=0.5] {defensive aura $-15\%$} (M);
\draw[helps] (M) -- node[gift, pos=0.3] {offensive aura $+15\%$} (R);
\draw[helps] (H) -- node[gift, pos=0.5] {spots, slows for} (R);
\draw[helps] (P) -- node[gift, pos=0.22] {heals} (W);
\end{tikzpicture}

%% file: sections/05_evaluation.tex

\section{The Evaluation: Scenarios, Tournament, Metrics}
\label{sec:evaluation}

\begin{figure}[t]
  \centering
  \input{figures/pipeline}
  \caption{The \bgs{} evaluation pipeline. Each small square is one map or scenario, and each stage has its own. Scripted opponents can play at every stage.}
  \label{fig:pipeline}
  \Description{A left-to-right flow in four stages: Train, Select, Test, and Analyze. Train shows the 42 training maps, 0 to 41, as small squares, with maps 0 to 11 lighter as the curriculum; agents train with self-play, opponent pools, and scripted opponents. An arrow leads to Select: the 5 held-out validation maps, 42 to 46, used to pick the best checkpoint. The arrow splits into the two test modes: Mode 1, the 8 endgame scenarios on their own arenas, where Team A plays TDM-Alpha or TDM-Beta; and Mode 2, the tournament of mirrored 5v5 games on the 5 held-out test maps, 47 to 51, rated with Elo and 95 percent intervals. Both modes lead to Analyze: metrics, 16 or 11,192 per game, including who did what to whom, and replays that show every tick from any agent's view.}
\end{figure}
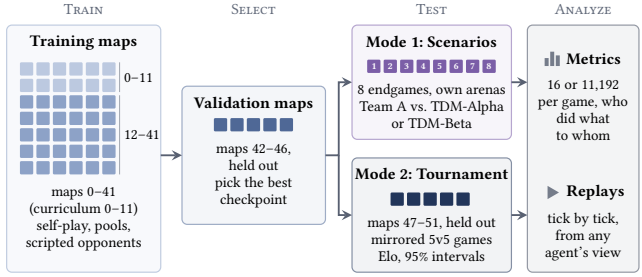

\bgs{} provides two canonical evaluation modes, challenging endgame scenarios and an official tournament. Both have comprehensive metrics that show the teamwork behind every result (Figure~\ref{fig:pipeline}). Researchers can run any other evaluation too, with one to five agents per team, and every results file names the systems, rosters, maps, and spawn ends that played.

\subsection{Mode 1: Endgame Scenarios}
\label{sec:eval-scenarios}

Each of the eight scenarios is a hand-designed, nearly lost position, built like a chess endgame puzzle. Each sits on its own arena, none of the 52 maps, so it is unseen by construction, and canonical results never use scenarios for training or selection. Team A must win within 5 to 10 ticks against a strong reactive controller, TDM-Alpha or TDM-Beta. Each winning line needs several of ten team skills (Table~\ref{tab:scenarios}), from teammate and opponent modeling to strategic sacrifice and multi-step team-level planning. In Scenario~6, for example, the Hunter traps the enemy Rogue and then holds its fire, because damage would free it. The Priest, at 1~HP, spends every heal on the Hunter, which becomes bait for the enemy Warrior's Charge. The Mage, powered by Burst on the first tick, kills the enemy Mage and the trapped Rogue, then finishes the Warrior after it charges. Section~\ref{supp:s5.1} of the supplementary material gives each scenario's rosters, score, controller, and solution length.
Every scenario has a verified winning line, re-checked tick by tick against the live controller, so a solved scenario shows a skill a team has, and a failed one a skill it still lacks. The score and the clock are part of each position, so the winning line may take risks that would be poor earlier in a game; a scenario therefore also shows whether a team trained with shaped rewards plays to win.

\input{tables/scenarios}

\subsection{Mode 2: The Official Tournament}
\label{sec:eval-tournament}

Scores in competitive games are relative, so \bgs{} rates entrants against a fixed field, the ``Big Nine'': the nine strongest entrants so far.

\emph{Games.} Every tournament game is a mirrored 5v5, with one of each class per team on mirrored spawn pads, on a test map (47--51). Each pair of entrants plays 200 games, 40 per test map, as 100 pairs that share a seed and swap spawn ends, so neither entrant gains from its starting end. Each challenger's seed is drawn only after its training run has finished, so no team can train on it, and is published with its results; games run in XLA's deterministic mode, so anyone can replay them exactly on the same GPU model and software (Sections~\ref{supp:s4.5} and~\ref{supp:s5.3} of the supplementary material).

\emph{Rating.} Each entrant gets one strength number, and the chance that it beats, draws with, or loses to another depends only on the gap between their numbers: the Bradley--Terry--Davidson model~\cite{bradley1952rank,davidson1970ties}, fit to all games at once and put on an Elo scale~\cite{elo1978rating} centered at 1,200. Its 95\% intervals show how far a rating would move if the same checkpoints played new games (Section~\ref{supp:s5.3} of the supplementary material gives the resampling). As in ranked online games, each Elo also maps to a tier, a label that never changes rank or admission. Eight tiers, from Iron to Grandmaster, step up every 200 Elo, with Gold starting at the field's average of 1,200, and each tier below Grandmaster splits into five divisions 40 Elo apart, numbered 5 (lowest) to 1 (highest). A 200-Elo lead means winning about 76\% of decisive games, those not drawn (Table~\ref{tab:tiers} in Section~\ref{supp:s5.2} of the supplementary material gives every tier's Elo range and win chance).


\emph{Entry.} Any System that runs as pure JAX may enter. Each team submits one training script per season, and the maintainers run it once on one RTX~5090, within 255 million game steps and 8 hours, validation included; the entrant must then play at least 5,000 game steps per second in at most 10\,GB of GPU memory, at 512 games, so that it stays practical as a training opponent. These limits keep a monthly tournament feasible, reward methods that are efficient in both samples and compute, and answer a call by Gorsane et al. for benchmark designers to track state-of-the-art results and vet submitted ones~\cite{gorsane2022standardised}. Challengers are evaluated one at a time: each plays every member of the Big Nine and replaces the lowest if its Elo is higher, with one seat per team, and every result is published, admitted or not.

\emph{Release.} Every entrant is released as a downloadable opponent with its code and all its training checkpoints, so the opponent pool grows every season. Later challengers can train against the incumbents, while each incumbent must beat challengers it never trained against: only teamwork that generalizes to unseen opponents keeps a place in the Big Nine. Season~0 holds our six baselines and three scripted opponents (Section~\ref{sec:exp-ladder}); from Season~1 in January~2027, a new table appears on the first day of each month (rulebook in Section~\ref{supp:s5.2} of the supplementary material).

\subsection{Metrics}
\label{sec:eval-metrics}

\begin{figure}[t]
  \centering
  \input{figures/metric_taxonomy}
  \caption{The 11,192 full metrics by topic. Bar length is the topic's number of columns; the dark part records who did what to whom. $^\dagger$Only enemies or allies already under a harmful effect, such as a slow or a stun.}
  \label{fig:metrics}
  \Description{A compact two-column list of the six metric sections and their 28 topics, each with its number of columns and a small bar. Damage and healing has 5,428 columns; kills, deaths, and respawns 2,108; status effects 1,532; results and actions 764; Ultimates 756; and teamwork and positioning 604. The largest topics are status applications (1,388), kills of enemies with a harmful effect (1,274), and damage to enemies with a harmful effect (1,204). Most of each large bar is dark: those columns record who did what to whom. The light part is the rest, mostly measures of one agent or one team.}
\end{figure}
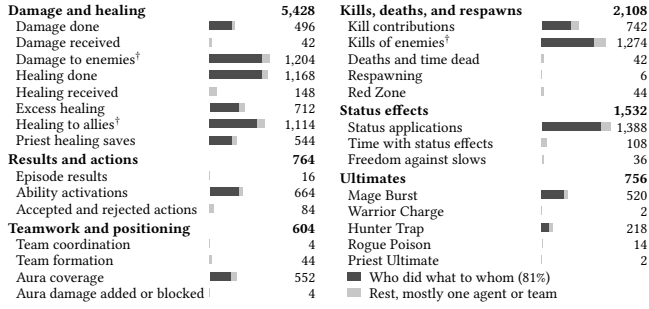

Beyond win rates, \bgs{} measures how agents cooperate, so researchers can see which teamwork a new method adds or an ablation removes. Researchers can choose between three metric storage options: \texttt{none}, \texttt{priority} (the default, 16 values per game), or \texttt{full} (11,192 values per game).
The full level covers 28 topics in six sections (Figure~\ref{fig:metrics}): results and actions; damage and healing; kills, deaths, and respawns; teamwork and positioning, such as focus fire, formation, and aura coverage; status effects; and Ultimates. Of these columns, 81\% record who did what to whom, such as which Priest healed which ally.
Metrics accumulate in JAX during play, with no stored trajectory, and training stays lean: full metrics are collected afterwards from frozen checkpoints and read in a pandas notebook or in the analytical Replay Viewer. One call, \texttt{system\_metrics}, compares every System from its own side, reading each game as if that System were Team~A, with 95\% intervals from resampled spawn-swapped pairs, as for Elo.

%% file: figures/pipeline.tex
\definecolor{plCur}{HTML}{BCC9E0}\definecolor{plTrain}{HTML}{8DA2C7}%
\definecolor{plVal}{HTML}{4F6896}\definecolor{plTest}{HTML}{22355A}%
\definecolor{plScen}{HTML}{7B5EA7}\definecolor{plInk}{HTML}{3A4458}%
\begin{tikzpicture}[x=1pt,y=1pt,font=\fontsize{5.6}{6.4}\selectfont,
  stage/.style={draw=#1!55, fill=#1!6, rounded corners=2pt, line width=0.45pt},
  head/.style={text=plInk!80, font=\fontsize{5.8}{6}\selectfont\scshape},
  title/.style={font=\fontsize{6.1}{6.6}\selectfont\bfseries, anchor=north, inner sep=0.5pt},
  body/.style={align=center, anchor=north, inner sep=0.5pt},
  flow/.style={-stealth, line width=0.7pt, plInk!85}]
\newcommand{\cell}[3]{\fill[#3, rounded corners=0.5pt] (#1,#2) rectangle ++(5,5);}
\node[head] at (29,103) {Train};
\node[head] at (93,103) {Select};
\node[head] at (160,103) {Test};
\node[head] at (217,103) {Analyze};
\draw[stage=plTrain] (0,4) rectangle (58,97);
\node[title] at (29,94) {Training maps};
\foreach \i in {0,...,41} {
  \pgfmathtruncatemacro{\r}{\i/6}\pgfmathtruncatemacro{\c}{\i-6*\r}
  \ifnum\i<12 \cell{5+6.2*\c}{78-6.2*\r}{plCur}\else\cell{5+6.2*\c}{78-6.2*\r}{plTrain}\fi}
\draw[plInk!45, line width=0.4pt] (42.5,71.8) -- (42.5,83) (42.5,40.8) -- (42.5,70.6);
\node[body, anchor=west] at (43.5,77.4) {0--11};
\node[body, anchor=west] at (43.5,55.7) {12--41};
\node[body] at (29,37) {maps 0--41\\(curriculum 0--11)\\self-play, pools,\\scripted opponents};
\draw[stage=plVal] (66,22) rectangle (120,74);
\node[title] at (93,71) {Validation maps};
\foreach \i in {0,...,4} {\cell{78+6.2*\i}{57}{plVal}}
\node[body] at (93,53) {maps 42--46,\\held out\\pick the best\\checkpoint};
\draw[stage=plScen] (130,54) rectangle (190,97);
\node[title] at (160,94) {Mode 1: Scenarios};
\foreach \i in {0,...,7} {\fill[plScen, rounded corners=0.5pt] (135.7+6.2*\i,79) rectangle ++(5,5);
  \node[text=white, font=\fontsize{3.6}{4}\selectfont\bfseries] at (138.2+6.2*\i,81.5) {\the\numexpr\i+1\relax};}
\node[body] at (160,75.5) {8 endgames, own arenas\\Team~A vs.\ TDM-Alpha\\or TDM-Beta};
\draw[stage=plTest] (130,4) rectangle (190,47);
\node[title] at (160,44) {Mode 2: Tournament};
\foreach \i in {0,...,4} {\cell{144.5+6.2*\i}{29}{plTest}}
\node[body] at (160,25.5) {maps 47--51, held out\\mirrored 5v5 games\\Elo, 95\% intervals};
\draw[stage=plInk] (196,4) rectangle (238,97);
\fill[plInk!70] (202.5,81.6) rectangle ++(1.5,3) (204.6,81.6) rectangle ++(1.5,5.4) (206.7,81.6) rectangle ++(1.5,4);
\node[title, anchor=west, inner sep=0] at (210.5,84.2) {Metrics};
\node[body] at (217,77) {16 or 11,192\\per game, who\\did what\\to whom};
\fill[plInk!70] (203.6,31.5) -- ++(0,5) -- ++(4.3,-2.5) -- cycle;
\node[title, anchor=west, inner sep=0] at (210.5,34) {Replays};
\node[body] at (217,27) {tick by tick,\\from any\\agent's view};
\draw[flow] (58,48) -- (66,48);
\draw[flow] (120,48) -- (125,48) |- (130,75.5);
\draw[flow] (125,48) |- (130,25.5);
\draw[flow] (190,75.5) -- (196,75.5);
\draw[flow] (190,25.5) -- (196,25.5);
\end{tikzpicture}

%% file: tables/scenarios.tex
\begin{table}[t]
\caption{The team skills each endgame scenario tests.}
\label{tab:scenarios}
\centering\footnotesize
\newcommand{\rot}[1]{\rotatebox{90}{#1}}
\setlength{\tabcolsep}{3.4pt}
\begin{tabular}{@{}c*{10}{c}c@{}}
\toprule
Scenario & \rot{Teammate modeling} & \rot{Opponent modeling} & \rot{Basic damage target timing} & \rot{Ultimate timing} & \rot{Tactical obstacle exploitation} & \rot{Tactical positioning} & \rot{Balancing (def/off)ensive play} & \rot{Strategic sacrifice} & \rot{Plan adaptation} & \rot{Multi-step team-level planning} & \rot{Skills tested} \\
\midrule
1 & {\normalsize$\bullet$} & {\normalsize$\bullet$} &  &  & {\normalsize$\bullet$} & {\normalsize$\bullet$} &  & {\normalsize$\bullet$} &  & {\normalsize$\bullet$} & 6 \\
2 & {\normalsize$\bullet$} & {\normalsize$\bullet$} &  & {\normalsize$\bullet$} & {\normalsize$\bullet$} & {\normalsize$\bullet$} &  &  &  & {\normalsize$\bullet$} & 6 \\
3 & {\normalsize$\bullet$} & {\normalsize$\bullet$} &  &  &  & {\normalsize$\bullet$} & {\normalsize$\bullet$} &  &  &  & 4 \\
4 & {\normalsize$\bullet$} & {\normalsize$\bullet$} &  &  &  & {\normalsize$\bullet$} &  &  &  & {\normalsize$\bullet$} & 4 \\
5 & {\normalsize$\bullet$} & {\normalsize$\bullet$} &  & {\normalsize$\bullet$} &  & {\normalsize$\bullet$} &  &  &  & {\normalsize$\bullet$} & 5 \\
6 & {\normalsize$\bullet$} & {\normalsize$\bullet$} & {\normalsize$\bullet$} & {\normalsize$\bullet$} & {\normalsize$\bullet$} & {\normalsize$\bullet$} &  &  & {\normalsize$\bullet$} & {\normalsize$\bullet$} & 8 \\
7 & {\normalsize$\bullet$} & {\normalsize$\bullet$} &  & {\normalsize$\bullet$} & {\normalsize$\bullet$} & {\normalsize$\bullet$} &  &  &  & {\normalsize$\bullet$} & 6 \\
8 & {\normalsize$\bullet$} & {\normalsize$\bullet$} &  & {\normalsize$\bullet$} & {\normalsize$\bullet$} & {\normalsize$\bullet$} &  &  &  & {\normalsize$\bullet$} & 6 \\
\addlinespace
Scenarios & 8 & 8 & 1 & 5 & 5 & 8 & 1 & 1 & 1 & 7 & \\
\bottomrule
\end{tabular}
\end{table}

%% file: figures/metric_taxonomy.tex
\begin{tikzpicture}[x=1pt,y=1pt,font=\fontsize{5.6}{6.15}\selectfont]
\node[anchor=base west,inner sep=0] at (0.0,0.0) {\textbf{Damage and healing}};
\node[anchor=base east,inner sep=0] at (116.0,0.0) {\textbf{5,428}};
\node[anchor=base west,inner sep=0] at (3.0,-6.2) {Damage done};
\fill[black!70] (76.00,-5.75) rectangle ++(8.43,3.2);
\fill[black!22] (84.43,-5.75) rectangle ++(0.86,3.2);
\node[anchor=base east,inner sep=0] at (116.0,-6.2) {496};
\node[anchor=base west,inner sep=0] at (3.0,-12.3) {Damage received};
\fill[black!22] (76.00,-11.90) rectangle ++(0.79,3.2);
\node[anchor=base east,inner sep=0] at (116.0,-12.3) {42};
\node[anchor=base west,inner sep=0] at (3.0,-18.5) {Damage to enemies$^\dagger$};
\fill[black!70] (76.00,-18.05) rectangle ++(19.67,3.2);
\fill[black!22] (95.67,-18.05) rectangle ++(2.88,3.2);
\node[anchor=base east,inner sep=0] at (116.0,-18.5) {1,204};
\node[anchor=base west,inner sep=0] at (3.0,-24.6) {Healing done};
\fill[black!70] (76.00,-24.20) rectangle ++(19.67,3.2);
\fill[black!22] (95.67,-24.20) rectangle ++(2.21,3.2);
\node[anchor=base east,inner sep=0] at (116.0,-24.6) {1,168};
\node[anchor=base west,inner sep=0] at (3.0,-30.8) {Healing received};
\fill[black!22] (76.00,-30.35) rectangle ++(2.77,3.2);
\node[anchor=base east,inner sep=0] at (116.0,-30.8) {148};
\node[anchor=base west,inner sep=0] at (3.0,-36.9) {Excess healing};
\fill[black!70] (76.00,-36.50) rectangle ++(11.24,3.2);
\fill[black!22] (87.24,-36.50) rectangle ++(2.10,3.2);
\node[anchor=base east,inner sep=0] at (116.0,-36.9) {712};
\node[anchor=base west,inner sep=0] at (3.0,-43.0) {Healing to allies$^\dagger$};
\fill[black!70] (76.00,-42.65) rectangle ++(17.98,3.2);
\fill[black!22] (93.98,-42.65) rectangle ++(2.88,3.2);
\node[anchor=base east,inner sep=0] at (116.0,-43.0) {1,114};
\node[anchor=base west,inner sep=0] at (3.0,-49.2) {Priest healing saves};
\fill[black!70] (76.00,-48.80) rectangle ++(8.43,3.2);
\fill[black!22] (84.43,-48.80) rectangle ++(1.76,3.2);
\node[anchor=base east,inner sep=0] at (116.0,-49.2) {544};
\node[anchor=base west,inner sep=0] at (0.0,-56.3) {\textbf{Results and actions}};
\node[anchor=base east,inner sep=0] at (116.0,-56.3) {\textbf{764}};
\node[anchor=base west,inner sep=0] at (3.0,-62.5) {Episode results};
\fill[black!22] (76.00,-62.10) rectangle ++(0.30,3.2);
\node[anchor=base east,inner sep=0] at (116.0,-62.5) {16};
\node[anchor=base west,inner sep=0] at (3.0,-68.6) {Ability activations};
\fill[black!70] (76.00,-68.25) rectangle ++(11.24,3.2);
\fill[black!22] (87.24,-68.25) rectangle ++(1.20,3.2);
\node[anchor=base east,inner sep=0] at (116.0,-68.6) {664};
\node[anchor=base west,inner sep=0] at (3.0,-74.8) {Accepted and rejected actions};
\fill[black!22] (76.00,-74.40) rectangle ++(1.57,3.2);
\node[anchor=base east,inner sep=0] at (116.0,-74.8) {84};
\node[anchor=base west,inner sep=0] at (0.0,-82.0) {\textbf{Teamwork and positioning}};
\node[anchor=base east,inner sep=0] at (116.0,-82.0) {\textbf{604}};
\node[anchor=base west,inner sep=0] at (3.0,-88.1) {Team coordination};
\fill[black!22] (76.00,-87.70) rectangle ++(0.07,3.2);
\node[anchor=base east,inner sep=0] at (116.0,-88.1) {4};
\node[anchor=base west,inner sep=0] at (3.0,-94.3) {Team formation};
\fill[black!22] (76.00,-93.85) rectangle ++(0.82,3.2);
\node[anchor=base east,inner sep=0] at (116.0,-94.3) {44};
\node[anchor=base west,inner sep=0] at (3.0,-100.4) {Aura coverage};
\fill[black!70] (76.00,-100.00) rectangle ++(8.24,3.2);
\fill[black!22] (84.24,-100.00) rectangle ++(2.10,3.2);
\node[anchor=base east,inner sep=0] at (116.0,-100.4) {552};
\node[anchor=base west,inner sep=0] at (3.0,-106.6) {Aura damage added or blocked};
\fill[black!22] (76.00,-106.15) rectangle ++(0.07,3.2);
\node[anchor=base east,inner sep=0] at (116.0,-106.6) {4};
\node[anchor=base west,inner sep=0] at (125.0,0.0) {\textbf{Kills, deaths, and respawns}};
\node[anchor=base east,inner sep=0] at (241.0,0.0) {\textbf{2,108}};
\node[anchor=base west,inner sep=0] at (128.0,-6.2) {Kill contributions};
\fill[black!70] (201.00,-5.75) rectangle ++(11.24,3.2);
\fill[black!22] (212.24,-5.75) rectangle ++(2.66,3.2);
\node[anchor=base east,inner sep=0] at (241.0,-6.2) {742};
\node[anchor=base west,inner sep=0] at (128.0,-12.3) {Kills of enemies$^\dagger$};
\fill[black!70] (201.00,-11.90) rectangle ++(19.67,3.2);
\fill[black!22] (220.67,-11.90) rectangle ++(4.20,3.2);
\node[anchor=base east,inner sep=0] at (241.0,-12.3) {1,274};
\node[anchor=base west,inner sep=0] at (128.0,-18.5) {Deaths and time dead};
\fill[black!22] (201.00,-18.05) rectangle ++(0.79,3.2);
\node[anchor=base east,inner sep=0] at (241.0,-18.5) {42};
\node[anchor=base west,inner sep=0] at (128.0,-24.6) {Respawning};
\fill[black!22] (201.00,-24.20) rectangle ++(0.11,3.2);
\node[anchor=base east,inner sep=0] at (241.0,-24.6) {6};
\node[anchor=base west,inner sep=0] at (128.0,-30.8) {Red Zone};
\fill[black!22] (201.00,-30.35) rectangle ++(0.82,3.2);
\node[anchor=base east,inner sep=0] at (241.0,-30.8) {44};
\node[anchor=base west,inner sep=0] at (125.0,-37.9) {\textbf{Status effects}};
\node[anchor=base east,inner sep=0] at (241.0,-37.9) {\textbf{1,532}};
\node[anchor=base west,inner sep=0] at (128.0,-44.0) {Status applications};
\fill[black!70] (201.00,-43.65) rectangle ++(22.48,3.2);
\fill[black!22] (223.48,-43.65) rectangle ++(3.52,3.2);
\node[anchor=base east,inner sep=0] at (241.0,-44.0) {1,388};
\node[anchor=base west,inner sep=0] at (128.0,-50.2) {Time with status effects};
\fill[black!22] (201.00,-49.80) rectangle ++(2.02,3.2);
\node[anchor=base east,inner sep=0] at (241.0,-50.2) {108};
\node[anchor=base west,inner sep=0] at (128.0,-56.3) {Freedom against slows};
\fill[black!22] (201.00,-55.95) rectangle ++(0.67,3.2);
\node[anchor=base east,inner sep=0] at (241.0,-56.3) {36};
\node[anchor=base west,inner sep=0] at (125.0,-63.5) {\textbf{Ultimates}};
\node[anchor=base east,inner sep=0] at (241.0,-63.5) {\textbf{756}};
\node[anchor=base west,inner sep=0] at (128.0,-69.6) {Mage Burst};
\fill[black!70] (201.00,-69.25) rectangle ++(8.43,3.2);
\fill[black!22] (209.43,-69.25) rectangle ++(1.31,3.2);
\node[anchor=base east,inner sep=0] at (241.0,-69.6) {520};
\node[anchor=base west,inner sep=0] at (128.0,-75.8) {Warrior Charge};
\fill[black!22] (201.00,-75.40) rectangle ++(0.04,3.2);
\node[anchor=base east,inner sep=0] at (241.0,-75.8) {2};
\node[anchor=base west,inner sep=0] at (128.0,-82.0) {Hunter Trap};
\fill[black!70] (201.00,-81.55) rectangle ++(2.81,3.2);
\fill[black!22] (203.81,-81.55) rectangle ++(1.27,3.2);
\node[anchor=base east,inner sep=0] at (241.0,-82.0) {218};
\node[anchor=base west,inner sep=0] at (128.0,-88.1) {Rogue Poison};
\fill[black!22] (201.00,-87.70) rectangle ++(0.26,3.2);
\node[anchor=base east,inner sep=0] at (241.0,-88.1) {14};
\node[anchor=base west,inner sep=0] at (128.0,-94.3) {Priest Ultimate};
\fill[black!22] (201.00,-93.85) rectangle ++(0.04,3.2);
\node[anchor=base east,inner sep=0] at (241.0,-94.3) {2};
\fill[black!70] (128.0,-100.00) rectangle ++(5,3.2);
\node[anchor=base west,inner sep=0] at (136.0,-100.4) {Who did what to whom (81\%)};
\fill[black!22] (128.0,-106.15) rectangle ++(5,3.2);
\node[anchor=base west,inner sep=0] at (136.0,-106.6) {Rest, mostly one agent or team};
\end{tikzpicture}

%% file: sections/06_tools.tex

\section{The Tools}
\label{sec:tools}

\emph{Opponents and learners.}
\bgs{} includes three strong scripted opponents, TDM-Alpha, TDM-Beta, and TDM-Gamma. These are deterministic JAX controllers that see only the inputs a learned actor may see. It also includes six deep MARL baselines: recurrent and feedforward MAPPO~\cite{yu2022mappo} and IPPO~\cite{dewitt2020ippo}, recurrent QMIX~\cite{rashid2018qmix}, and recurrent PQN~\cite{gallici2025pqn} with VDN~\cite{sunehag2018vdn} value mixing, adapted from Mava~\cite{dekock2023mava} (PPO and QMIX) and JaxMARL~\cite{rutherford2024jaxmarl} (PQN-VDN). Section~\ref{supp:s6} of the supplementary material gives full details and configuration options.

\emph{Ready-made research setups.}
We provide support and easy implementations of common MARL setups: self-play against current and past copies, weighted opponent pools, mixed teams per slot or per class through \texttt{team()}, per-game mixtures through \texttt{pool()}, curricula, frozen partners, and cross-play grids. Because evaluation accepts any mix of Systems in any slot, zero-shot coordination and ad hoc teamwork are natively supported~\cite{hu2020otherplay,mirsky2022survey}. Researchers can pass their own reward function or use the built-in potential-based and score-difference shaping, applied inside the compiled training loop; shaping changes only the learner's training signal, never the actors' inputs or the official score.

\emph{LLM-agent API.}
The game has no per-turn time limit, so slow methods such as LLM agents and planners can play it and test themselves against the released Big Nine; only official leaderboard entries must run as pure JAX. Our LLM API, \texttt{llm.make\_system(model, server\_url)}, turns any language model behind a standard Chat Completions endpoint, such as vLLM~\cite{kwon2023vllm}, into an ordinary System. Each actor's prompt holds only the public game rules, its own permitted view, and its legal actions, and its JSON reply is checked against the action mask (Section~\ref{supp:s7.1} of the supplementary material lists the API's other options, such as custom prompts and models run any other way).

\emph{Clients.}
In the BattleClient, humans play alongside or against any mix of policies, for human-AI studies such as how well agents coordinate with human teammates. The analytical Replay Viewer replays any saved game from the full view or from one agent's point of view (Figure~\ref{fig:supp-arena} in Section~\ref{supp:s3.1} of the supplementary material shows both views of the same moment), with analytical panels and image export: researchers step through a game tick by tick, see what each agent saw and chose, and read the metrics up to that moment, including who did what to whom. Our website has interactive demos of both on its \weblink{https://marl-battlegrounds.com/\#battleclient}{BattleClient} and \weblink{https://marl-battlegrounds.com/\#replay-viewer}{Replay Viewer} pages.

%% file: sections/07_experiments.tex

\section{Experiments}
\label{sec:experiments}


\subsection{Setup}
\label{sec:exp-setup}

Every baseline learned the same game from the same mix of opponents, so the ladder compares methods, not setups: as Team~A in mirrored 5v5 games on training maps 0--41, against its current self (60\%), one of its 12 newest saved copies (20\%), one of six fixed recurrent MAPPO models (15\%), or TDM-Alpha or TDM-Beta (5\%). Besides $+1$ for a win and $-1$ for a loss, agents earn small rewards for kills, damage, useful healing (healing that restores missing health), and keeping allies within the Priest's healing range, and lose reward for deaths and for standing in their own Red Zone; QMIX, PQN-VDN, and feedforward MAPPO get the same terms summed into one team reward. The full metrics shaped this reward: without the Red Zone terms, teams learned to hide at home, and most of the Priest's healing went to allies at full health. At routine checks, a run plays 128 games on four validation maps against TDM-Alpha, TDM-Beta, and two opponents it never trains against, TDM-Gamma and a 250-million-step recurrent MAPPO model, scored by its point margin averaged over the 16 map-and-opponent cells. Two studies of 84 runs of 40 million game steps, each recipe on three seeds, searched each method's settings; its chosen run then continued from its full saved state to 250.7 million steps, and the best of its two best checks and its last checkpoint, each replayed in 256 new games, became its Season~0 policy (every term, setting, and seed in Section~\ref{supp:s6} of the supplementary material).

\subsection{Speed and Cost}
\label{sec:exp-speed}

How fast is \bgs{}, and what does a baseline cost to train? Like JaxMARL~\cite{rutherford2024jaxmarl}, we time random legal actions in compiled 500-tick rollouts, compilation excluded. On one RTX~5090, throughput rises with the number of parallel 5v5 games, from 4,919 game steps per second at 32 games to 50,416 game steps, or 504,165 agent steps, per second at 1,024, still rising at the largest batch we measured (Figure~\ref{fig:supp-speed} and the timing setup in Section~\ref{supp:s6.5} of the supplementary material). Training adds the learner: recurrent MAPPO in pure self-play with 1,024 parallel games trains at 29,878 game steps per second with 5.0\,GB of peak JAX memory. With validation included, each PPO baseline reached 250 million game steps in 3.6 to 4.3 hours on one RTX~5090, and QMIX and PQN-VDN in 5.5 and 6.9 hours, so every method fits the tournament's budget of 8 hours on one card.


\subsection{Baselines and the Season 0 Ladder}
\label{sec:exp-ladder}

\begin{figure}[t]
  \centering
  \input{figures/elo}
  \caption{Season~0 Elo ratings on test maps 47--51 with 95\% intervals (filled: our baselines; open: scripted opponents). Dashed line: average Elo, 1,200; bands: tiers (Section~\ref{sec:eval-tournament}).}
  \label{fig:elo}
  \Description{A horizontal dot plot of Elo ratings from about 950 to 1,480 for nine entrants, from top to bottom: recurrent IPPO 1,441, PQN-VDN 1,358, recurrent MAPPO 1,304, QMIX 1,300, feedforward IPPO 1,240, feedforward MAPPO 1,140, TDM-Beta 1,043, TDM-Gamma 1,000, and TDM-Alpha 974. Each point has a 95 percent interval about 10 to 18 points wide on each side. Shaded bands mark the tiers: Bronze below 1,000, Silver from 1,000, Gold from 1,200, and Platinum from 1,400. A dashed line at 1,200 marks the field's average. All six learned baselines sit above all three scripted opponents.}
\end{figure}
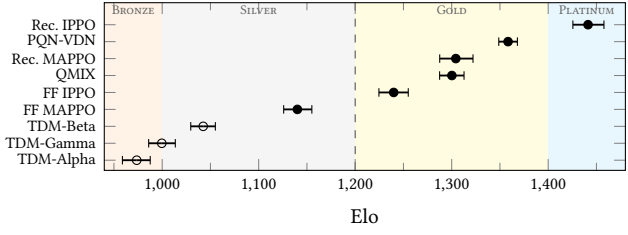

Can deep MARL learn this game, and which method leads? The scripted opponents set a real bar (Figures~\ref{fig:supp-curves} and~\ref{fig:supp-wins} in Section~\ref{supp:s6.3} of the supplementary material chart each baseline's point margin, kills, and wins at every check): after 40 million game steps, every baseline still lost to them on point margin, by 1.7 to 12.5 points per game, and none of the 54 runs of the first search study beat any of them. Each baseline first won half its games against them only after 51 to 166 million steps. The longer training taught fighting, not hiding: at its selected checkpoint, each baseline scored $+1.9$ to $+7.3$ points per game against the validation panel, with 16 to 19 kills per game, up from 6 to 13 at 40 million steps, while its deaths fell or held level. In the Season~0 round robin, every pairing of the six baselines and three scripted opponents played 200 games on the test maps; all 7,200 games took 8.2 minutes on one RTX~5090. All six baselines rank above all three scripted opponents (Figure~\ref{fig:elo}). Recurrent IPPO leads with 1,441 Elo (Platinum~4), so it would win about 80\% of decisive games against a policy of average strength, and it wins 579 of its 600 games against the scripted opponents it lost to at 40 million steps. It is also the only baseline still clearly improving at 250 million steps, so the top of the ladder is not yet the ceiling of its method.


\subsection{Scenario Results}
\label{sec:exp-scenarios}

Can any Season~0 system win a nearly lost position it has never seen? Each member of the Big Nine, and Random, which plays random legal actions, played all eight scenarios as Team~A, 256 games per scenario. None solves any scenario (Table~\ref{tab:scenario-results} in Section~\ref{supp:s5.1} of the supplementary material gives every system's wins, draws, and losses in each scenario): the Big Nine win 23 of their 18,432 games, all in Scenarios 1--3, and every system loses every game of Scenarios 5--8. No system wins more than 8 of its 256 games in any scenario, and none wins any scenario significantly more often than Random (two-sided Fisher's exact test, $p > 0.1$, so the few wins are consistent with luck). The full metrics show how every system loses. In every scenario but Scenario~1, Team~B starts one kill from victory, and in Scenarios~2 and 5--8 that kill comes at once: 12,767 of the 12,793 losses there end within two ticks. The victim is the weak teammate Team~B reaches first: the Priest, at 1~HP, in 97.0\% of the Scenario~5 losses and 99.9\% of the Scenario~8 losses, and the Hunter, at 7~HP beside the enemy Rogue, in 98.3\% of the Scenario~6 losses. The Priest's full heal is not ready before tick 3, so a winning team must save that teammate by where it stands and when it strikes. No system, learned or scripted, does: the ladder ranks the field, and the scenarios show what all of it still lacks.


\subsection{How Baseline Agents Cooperated}
\label{sec:exp-teamwork}

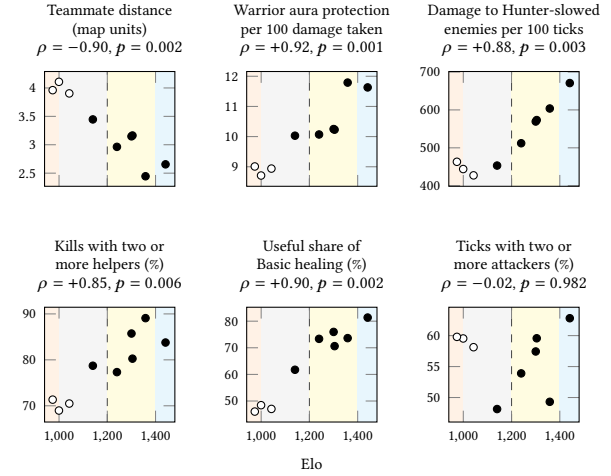
\begin{figure}[t]
  \centering
  \input{figures/teamwork}
  \caption{Six team measures against Elo for the nine Season~0 entrants. Points, tier bands, and the dashed line are as in Figure~\ref{fig:elo}; every 95\% interval hides behind its point. Each panel gives $\rho$, the rank correlation with Elo, and its exact $p$.}
  \label{fig:teamwork}
  \Description{Six small scatter plots in two rows of three, each with Elo from about 940 to 1,480 on the horizontal axis, shaded tier bands, a dashed line at 1,200, and one point per Season 0 entrant: filled points for the six learned baselines and open points for the three scripted opponents. Teammate distance falls as Elo rises (rank correlation -0.90). Warrior aura protection per 100 points of damage taken rises (+0.92). Damage to enemies a teammate Hunter had slowed rises (+0.88). The share of kills with two or more helpers rises (+0.85). The useful share of Basic healing rises (+0.90). The share of ticks with two or more attackers shows no trend (-0.02).}
\end{figure}

Which teamwork separates strong teams from weak ones? We replayed all 7,200 Season~0 games with full metrics, every game repeating exactly, and pooled each entrant's 1,600 games from its own side, counting events per 100 ticks because games differ in length. For each team-level metric, $\rho$, Spearman's rank correlation with Elo over the nine entrants, is $+1$ when every stronger entrant has more of the metric, and its exact $p$ is the chance that a random ordering of the nine ratings correlates at least as strongly. Corrected for testing 138 metrics at once (Holm's and Westfall-Young's methods), one trend survives: the three scripted opponents, the lowest-rated entrants, lean on the Warrior's Ultimate for 14\% of their kills, the baselines for 3--4\%; we read the other trends as exploratory (Sections~\ref{supp:s5.4} and~\ref{supp:s8} of the supplementary material). Strong teams do not kill faster ($\rho = +0.02$); they die less ($\rho = -0.83$, $p = 0.008$), and their abilities combine (Figure~\ref{fig:teamwork}). They stand closer together ($\rho = -0.90$), so their auras pay off: the Warrior's aura prevents 11.6 of every 100 points of damage recurrent IPPO's team takes, against 9.0 for the lowest-rated entrant ($\rho = +0.92$). They cash in each other's setups: damage to enemies a teammate Hunter has slowed rises from the lowest-rated entrant's 463 per 100 ticks to recurrent IPPO's 671 ($\rho = +0.88$), and the share of kills with two or more helpers, teammates who dealt damage or gave a useful heal on the killing tick, from 71\% to 84\% ($\rho = +0.85$). Their Priests heal the teammate who needs it: the share of Basic healing that restores missing health rises from 46\% to 81\% ($\rho = +0.90$). Yet attacking at the same time does not follow Elo ($\rho = -0.02$): strength lies in how abilities combine, not in how often agents attack together.

\subsection{An LLM Team}
\label{sec:exp-llm}

Can \bgs{} evaluate an LLM team with the same tools as any other System? Through our LLM API, Qwen3.5-9B~\cite{qwen2026qwen35}, served by vLLM with 8-bit weights on one RTX~5090, plays every agent once per tick, with greedy decoding, thinking off, no memory of earlier turns, and replies in a JSON schema the server enforces. It played each member of the Big Nine ten times, twice per test map with spawn ends swapped, and each scenario 256 times. Running it was practical: 99.97\% of its 62,629 replies against the Big Nine were legal, a reply took 1.7 seconds at the median, and the 90 games took 3.5 hours. Playing well was not: it won none of its 90 games (13 draws, 77 losses) and none of its 2,048 scenario games, made 5 kills in all while dying 10.8 times per game, and lost by 14.2 to 20.8 points per game. The same metrics show why: it used its Basic ability 31 times per game to its opponents' 271, had two or more agents dealing damage in 1.1 ticks per game to their 45.7, shared no kill to their 8.5, and kept its agents 6.8 map units apart to their 2.9. Each agent chose legal moves on its own but rarely attacked, and no team formed: following the rules is not playing as a team (prompt, settings, and results per opponent in Section~\ref{supp:s7} of the supplementary material).


\subsection{Zero-Shot Coordination}
\label{sec:exp-zsc}

How well does the Season~0 winner play with a teammate it never trained with?~\cite{mirsky2022survey} Every baseline trained only beside copies of itself, so an agent from another method is a stranger. With \texttt{team()}, we built 31 teams: recurrent IPPO alone, and recurrent IPPO with one class slot played by the same class's agent from one of the other five methods, or by Random. Each team played all nine Season~0 entrants 200 times on the test maps from the same game starts, so only the stranger differs. Strangers hurt coordination: 
24 of the 25 learned strangers cost 0.8 to 7.2 points per game, 
with four turning the winner into a team that loses more games than it wins. Section~\ref{supp:s9} of the supplementary material gives the method, charts every stranger's cost in Figure~\ref{fig:supp-strangers}, and compares how well each method's agents fit in. We found the Mage and Priest slots matter most, for learned and random strangers alike. The position of a stranger explains the cost incurred: within each class, the farther it stands from its teammates, the more points the team loses (Spearman's rank correlation averaged over the five classes, $-0.84$; shuffling strangers within each class almost never orders them this consistently, exact $p = 0.00002$). This is more prominent for the Mage, Priest, and Warrior, whose auras and heals reach only nearby teammates. 
The team does not adjust to the stranger either. Its other four agents stand where they always do, within 0.12 map units, and its Priest heals 19 of the 20 non-Priest strangers worse, for the damage they take, than the agents they replaced.
This analysis reinforces known information about multi-agent systems, that self-play builds conventions that outsiders do not share~\cite{carroll2019overcooked,hu2020otherplay}, which can significantly hurt coordination. As such, a stranger's value depends on how well it is able to coordinate with a team that will not adapt to it.


\subsection{Human-AI Evaluation}
\label{sec:exp-human}

How does the Season~0 winner fare against, and alongside, an expert human? Our Lead Game Developer, a MOBA and MMORPG PvP player for over 20 years who designed the game and whom we call the Expert Human, played six games on test maps 47--49 in the BattleClient's Agent POV, which shows only what their team sees, the same view recurrent IPPO's agents get (setup in Section~\ref{supp:s10} of the supplementary material).

\emph{Human vs AI.} Controlling all five agents, the Expert Human won all three games against recurrent IPPO, 21--4, 20--8, and 21--1, with 47 kills to 13 deaths. Comparatively, no method  beat recurrent IPPO by more than 14 points in 960 games on these maps. What separated them was timing (Section~\ref{supp:s10} of the supplementary material tabulates every count). Recurrent IPPO used 51\% of its Ultimates on the first tick it could, the Expert Human 25\%, and its Mage pressed none of its 16 Bursts with an enemy in reach, against 14 of 15. The Expert Human made gos, pushes in which a team spends its major abilities together to force a kill: 89\% of its Ultimates came within 4 ticks of a teammate's, where luck, each agent's Ultimates shifted by a random number of ticks, gives about 60\%, while recurrent IPPO's 57\% is what luck gives. Recurrent IPPO runs its shared network separately for each agent, so a go needs five separate decisions to line up; the Expert Human made them as one.

\emph{Human-AI zero-shot coordination.} With the Expert Human controlling the Mage beside four recurrent IPPO agents against PQN-VDN, the mixed team won two of its three games, with 45 kills to 36 deaths, each margin within the range of recurrent IPPO's own 20 games on that map and spawn end, yet the partnership failed. On questions adapted from Hoffman's fluency measures~\cite{hoffman2013fluency,hoffman2019fluency}, plus three of our own on predictability and adaptation, all from 1 (strongly disagree) to 7 (strongly agree), the Expert Human rated fluency, trust, and predictability 1 and their own share of the work 7, and said they adapted (5) while the team did not (1); their workload on the NASA Task Load Index~\cite{hart1988nasatlx}, from 0 to 100, rose from 22.5 against recurrent IPPO to 57.5 beside it. As in Hanabi, where people rated a learned teammate far below a rule-based one with no difference in score~\cite{siu2021hanabi}, a matched score can hide a partnership that does not work.

%% file: figures/elo.tex
\begin{tikzpicture}
\begin{axis}[width=\linewidth, height=3.8cm,
  xmin=940, xmax=1480, ymin=-0.3, ymax=9.6, y dir=reverse,
  xtick={1000,1100,1200,1300,1400}, minor xtick={950,1050,1150,1250,1350,1450},
  ytick={1,...,9}, yticklabels={Rec.\ IPPO, PQN-VDN, Rec.\ MAPPO, QMIX, FF IPPO, FF MAPPO, TDM-Beta, TDM-Gamma, TDM-Alpha},
  xlabel={Elo}, tick label style={font=\scriptsize}, label style={font=\small},
  axis on top]
  \fill[orange!10] (axis cs:940,-0.3) rectangle (axis cs:1000,9.6);
  \fill[gray!8] (axis cs:1000,-0.3) rectangle (axis cs:1200,9.6);
  \fill[yellow!14] (axis cs:1200,-0.3) rectangle (axis cs:1400,9.6);
  \fill[cyan!8] (axis cs:1400,-0.3) rectangle (axis cs:1480,9.6);
  \node[font=\tiny\scshape, text=black!60, anchor=center] at (axis cs:970,0.1) {Bronze};
  \node[font=\tiny\scshape, text=black!60, anchor=center] at (axis cs:1100,0.1) {Silver};
  \node[font=\tiny\scshape, text=black!60, anchor=center] at (axis cs:1300,0.1) {Gold};
  \node[font=\tiny\scshape, text=black!60, anchor=center] at (axis cs:1440,0.1) {Platinum};
  \draw[black!70, dashed] (axis cs:1200,-0.3) -- (axis cs:1200,9.6);
  \addplot[only marks, mark=*, mark size=1.6pt, black, error bars/.cd, x dir=both, x explicit,
    error bar style={line width=0.7pt}, error mark options={rotate=90, mark size=1.6pt}]
    coordinates {
      (1441.3,1) -= (15.7,0) += (16.5,0)
      (1358.4,2) -= (9.6,0) += (9.7,0)
      (1304.3,3) -= (16.7,0) += (17.8,0)
      (1300.2,4) -= (12.7,0) += (12.6,0)
      (1239.9,5) -= (15.3,0) += (15.1,0)
      (1140.1,6) -= (14.2,0) += (15.0,0)};
  \addplot[only marks, mark=o, mark size=1.6pt, black, error bars/.cd, x dir=both, x explicit,
    error bar style={line width=0.7pt}, error mark options={rotate=90, mark size=1.6pt}]
    coordinates {
      (1042.5,7) -= (12.9,0) += (12.6,0)
      (999.6,8) -= (13.6,0) += (14.0,0)
      (973.6,9) -= (14.8,0) += (14.0,0)};
\end{axis}
\end{tikzpicture}

%% file: figures/teamwork.tex
\begin{tikzpicture}
\begin{groupplot}[group style={group size=3 by 2, horizontal sep=0.95cm,
    vertical sep=1.6cm, xticklabels at=edge bottom},
  width=0.39\linewidth, height=3.1cm, xmin=940, xmax=1480,
  xtick={1000,1200,1400}, scaled y ticks=false,
  tick label style={font=\tiny}, title style={font=\scriptsize,
    align=center, yshift=-0.9ex}, axis on top]
\nextgroupplot[title={Teammate distance\\(map units)\\$\rho = -0.90$, $p = 0.002$}]
  \fill[orange!10] ({axis cs:940,0} |- {rel axis cs:0,0}) rectangle ({axis cs:999.5,0} |- {rel axis cs:0,1});
  \fill[gray!8] ({axis cs:999.5,0} |- {rel axis cs:0,0}) rectangle ({axis cs:1199.5,0} |- {rel axis cs:0,1});
  \fill[yellow!14] ({axis cs:1199.5,0} |- {rel axis cs:0,0}) rectangle ({axis cs:1399.5,0} |- {rel axis cs:0,1});
  \fill[cyan!8] ({axis cs:1399.5,0} |- {rel axis cs:0,0}) rectangle ({axis cs:1480,0} |- {rel axis cs:0,1});
  \draw[black!70, dashed] ({axis cs:1200,0} |- {rel axis cs:0,0}) -- ({axis cs:1200,0} |- {rel axis cs:0,1});
  \addplot[only marks, mark=*, black, mark size=1.4pt, mark options={line width=0.4pt},
    error bars/.cd, y dir=both, y explicit, error bar style={black}, error mark=none]
    coordinates {
      (1441.3,2.655) -= (0,0.009) += (0,0.010)
      (1358.4,2.445) -= (0,0.008) += (0,0.008)
      (1304.3,3.163) -= (0,0.015) += (0,0.016)
      (1300.2,3.148) -= (0,0.012) += (0,0.011)
      (1239.9,2.963) -= (0,0.012) += (0,0.013)
      (1140.1,3.446) -= (0,0.042) += (0,0.043)
    };
  \addplot[only marks, mark=*, fill=white, draw=black, mark size=1.4pt, mark options={line width=0.4pt},
    error bars/.cd, y dir=both, y explicit, error bar style={black}, error mark=none]
    coordinates {
      (1042.5,3.904) -= (0,0.011) += (0,0.012)
      (999.6,4.108) -= (0,0.014) += (0,0.013)
      (973.6,3.961) -= (0,0.013) += (0,0.013)
    };
\nextgroupplot[title={Warrior aura protection\\per 100 damage taken\\$\rho = +0.92$, $p = 0.001$}]
  \fill[orange!10] ({axis cs:940,0} |- {rel axis cs:0,0}) rectangle ({axis cs:999.5,0} |- {rel axis cs:0,1});
  \fill[gray!8] ({axis cs:999.5,0} |- {rel axis cs:0,0}) rectangle ({axis cs:1199.5,0} |- {rel axis cs:0,1});
  \fill[yellow!14] ({axis cs:1199.5,0} |- {rel axis cs:0,0}) rectangle ({axis cs:1399.5,0} |- {rel axis cs:0,1});
  \fill[cyan!8] ({axis cs:1399.5,0} |- {rel axis cs:0,0}) rectangle ({axis cs:1480,0} |- {rel axis cs:0,1});
  \draw[black!70, dashed] ({axis cs:1200,0} |- {rel axis cs:0,0}) -- ({axis cs:1200,0} |- {rel axis cs:0,1});
  \addplot[only marks, mark=*, black, mark size=1.4pt, mark options={line width=0.4pt},
    error bars/.cd, y dir=both, y explicit, error bar style={black}, error mark=none]
    coordinates {
      (1441.3,11.630) -= (0,0.060) += (0,0.060)
      (1358.4,11.790) -= (0,0.040) += (0,0.050)
      (1304.3,10.230) -= (0,0.060) += (0,0.050)
      (1300.2,10.250) -= (0,0.050) += (0,0.050)
      (1239.9,10.070) -= (0,0.070) += (0,0.060)
      (1140.1,10.030) -= (0,0.080) += (0,0.080)
    };
  \addplot[only marks, mark=*, fill=white, draw=black, mark size=1.4pt, mark options={line width=0.4pt},
    error bars/.cd, y dir=both, y explicit, error bar style={black}, error mark=none]
    coordinates {
      (1042.5,8.940) -= (0,0.040) += (0,0.030)
      (999.6,8.710) -= (0,0.040) += (0,0.030)
      (973.6,9.010) -= (0,0.040) += (0,0.040)
    };
\nextgroupplot[title={Damage to Hunter-slowed\\enemies per 100 ticks\\$\rho = +0.88$, $p = 0.003$}]
  \fill[orange!10] ({axis cs:940,0} |- {rel axis cs:0,0}) rectangle ({axis cs:999.5,0} |- {rel axis cs:0,1});
  \fill[gray!8] ({axis cs:999.5,0} |- {rel axis cs:0,0}) rectangle ({axis cs:1199.5,0} |- {rel axis cs:0,1});
  \fill[yellow!14] ({axis cs:1199.5,0} |- {rel axis cs:0,0}) rectangle ({axis cs:1399.5,0} |- {rel axis cs:0,1});
  \fill[cyan!8] ({axis cs:1399.5,0} |- {rel axis cs:0,0}) rectangle ({axis cs:1480,0} |- {rel axis cs:0,1});
  \draw[black!70, dashed] ({axis cs:1200,0} |- {rel axis cs:0,0}) -- ({axis cs:1200,0} |- {rel axis cs:0,1});
  \addplot[only marks, mark=*, black, mark size=1.4pt, mark options={line width=0.4pt},
    error bars/.cd, y dir=both, y explicit, error bar style={black}, error mark=none]
    coordinates {
      (1441.3,670.800) -= (0,5.380) += (0,5.230)
      (1358.4,603.730) -= (0,4.820) += (0,5.020)
      (1304.3,573.200) -= (0,6.770) += (0,6.680)
      (1300.2,569.390) -= (0,4.550) += (0,4.350)
      (1239.9,512.110) -= (0,5.660) += (0,5.490)
      (1140.1,453.250) -= (0,6.720) += (0,6.620)
    };
  \addplot[only marks, mark=*, fill=white, draw=black, mark size=1.4pt, mark options={line width=0.4pt},
    error bars/.cd, y dir=both, y explicit, error bar style={black}, error mark=none]
    coordinates {
      (1042.5,427.420) -= (0,3.600) += (0,3.430)
      (999.6,444.080) -= (0,3.440) += (0,3.410)
      (973.6,463.210) -= (0,3.390) += (0,3.470)
    };
\nextgroupplot[title={Kills with two or\\more helpers (\%)\\$\rho = +0.85$, $p = 0.006$}]
  \fill[orange!10] ({axis cs:940,0} |- {rel axis cs:0,0}) rectangle ({axis cs:999.5,0} |- {rel axis cs:0,1});
  \fill[gray!8] ({axis cs:999.5,0} |- {rel axis cs:0,0}) rectangle ({axis cs:1199.5,0} |- {rel axis cs:0,1});
  \fill[yellow!14] ({axis cs:1199.5,0} |- {rel axis cs:0,0}) rectangle ({axis cs:1399.5,0} |- {rel axis cs:0,1});
  \fill[cyan!8] ({axis cs:1399.5,0} |- {rel axis cs:0,0}) rectangle ({axis cs:1480,0} |- {rel axis cs:0,1});
  \draw[black!70, dashed] ({axis cs:1200,0} |- {rel axis cs:0,0}) -- ({axis cs:1200,0} |- {rel axis cs:0,1});
  \addplot[only marks, mark=*, black, mark size=1.4pt, mark options={line width=0.4pt},
    error bars/.cd, y dir=both, y explicit, error bar style={black}, error mark=none]
    coordinates {
      (1441.3,83.770) -= (0,0.480) += (0,0.450)
      (1358.4,89.080) -= (0,0.280) += (0,0.290)
      (1304.3,80.260) -= (0,0.470) += (0,0.460)
      (1300.2,85.730) -= (0,0.330) += (0,0.310)
      (1239.9,77.330) -= (0,0.570) += (0,0.530)
      (1140.1,78.720) -= (0,0.600) += (0,0.620)
    };
  \addplot[only marks, mark=*, fill=white, draw=black, mark size=1.4pt, mark options={line width=0.4pt},
    error bars/.cd, y dir=both, y explicit, error bar style={black}, error mark=none]
    coordinates {
      (1042.5,70.490) -= (0,0.460) += (0,0.450)
      (999.6,68.970) -= (0,0.420) += (0,0.420)
      (973.6,71.340) -= (0,0.430) += (0,0.440)
    };
\nextgroupplot[title={Useful share of\\Basic healing (\%)\\$\rho = +0.90$, $p = 0.002$}, xlabel={Elo}, label style={font=\scriptsize}]
  \fill[orange!10] ({axis cs:940,0} |- {rel axis cs:0,0}) rectangle ({axis cs:999.5,0} |- {rel axis cs:0,1});
  \fill[gray!8] ({axis cs:999.5,0} |- {rel axis cs:0,0}) rectangle ({axis cs:1199.5,0} |- {rel axis cs:0,1});
  \fill[yellow!14] ({axis cs:1199.5,0} |- {rel axis cs:0,0}) rectangle ({axis cs:1399.5,0} |- {rel axis cs:0,1});
  \fill[cyan!8] ({axis cs:1399.5,0} |- {rel axis cs:0,0}) rectangle ({axis cs:1480,0} |- {rel axis cs:0,1});
  \draw[black!70, dashed] ({axis cs:1200,0} |- {rel axis cs:0,0}) -- ({axis cs:1200,0} |- {rel axis cs:0,1});
  \addplot[only marks, mark=*, black, mark size=1.4pt, mark options={line width=0.4pt},
    error bars/.cd, y dir=both, y explicit, error bar style={black}, error mark=none]
    coordinates {
      (1441.3,81.360) -= (0,0.300) += (0,0.280)
      (1358.4,73.630) -= (0,0.410) += (0,0.430)
      (1304.3,70.620) -= (0,0.610) += (0,0.630)
      (1300.2,75.960) -= (0,0.460) += (0,0.420)
      (1239.9,73.360) -= (0,0.500) += (0,0.470)
      (1140.1,61.740) -= (0,0.720) += (0,0.690)
    };
  \addplot[only marks, mark=*, fill=white, draw=black, mark size=1.4pt, mark options={line width=0.4pt},
    error bars/.cd, y dir=both, y explicit, error bar style={black}, error mark=none]
    coordinates {
      (1042.5,47.030) -= (0,0.320) += (0,0.330)
      (999.6,48.430) -= (0,0.330) += (0,0.340)
      (973.6,46.030) -= (0,0.330) += (0,0.320)
    };
\nextgroupplot[title={Ticks with two or\\more attackers (\%)\\$\rho = -0.02$, $p = 0.982$}]
  \fill[orange!10] ({axis cs:940,0} |- {rel axis cs:0,0}) rectangle ({axis cs:999.5,0} |- {rel axis cs:0,1});
  \fill[gray!8] ({axis cs:999.5,0} |- {rel axis cs:0,0}) rectangle ({axis cs:1199.5,0} |- {rel axis cs:0,1});
  \fill[yellow!14] ({axis cs:1199.5,0} |- {rel axis cs:0,0}) rectangle ({axis cs:1399.5,0} |- {rel axis cs:0,1});
  \fill[cyan!8] ({axis cs:1399.5,0} |- {rel axis cs:0,0}) rectangle ({axis cs:1480,0} |- {rel axis cs:0,1});
  \draw[black!70, dashed] ({axis cs:1200,0} |- {rel axis cs:0,0}) -- ({axis cs:1200,0} |- {rel axis cs:0,1});
  \addplot[only marks, mark=*, black, mark size=1.4pt, mark options={line width=0.4pt},
    error bars/.cd, y dir=both, y explicit, error bar style={black}, error mark=none]
    coordinates {
      (1441.3,62.810) -= (0,0.240) += (0,0.230)
      (1358.4,49.310) -= (0,0.290) += (0,0.310)
      (1304.3,59.570) -= (0,0.560) += (0,0.570)
      (1300.2,57.440) -= (0,0.360) += (0,0.320)
      (1239.9,53.900) -= (0,0.380) += (0,0.360)
      (1140.1,48.140) -= (0,0.550) += (0,0.530)
    };
  \addplot[only marks, mark=*, fill=white, draw=black, mark size=1.4pt, mark options={line width=0.4pt},
    error bars/.cd, y dir=both, y explicit, error bar style={black}, error mark=none]
    coordinates {
      (1042.5,58.130) -= (0,0.170) += (0,0.170)
      (999.6,59.540) -= (0,0.180) += (0,0.170)
      (973.6,59.800) -= (0,0.180) += (0,0.180)
    };
\end{groupplot}
\end{tikzpicture}

%% file: sections/08_limitations.tex

\section{Limitations, Future Work, and Availability}
\label{sec:limitations}

Our LLM results come from one small model with greedy decoding, thinking off, and no memory of earlier turns (Section~\ref{sec:exp-llm}), so they show that LLM teams can be evaluated here, not how well larger reasoning models coordinate; future work should test those with thinking and memory. Our human-AI results come from six games and one set of ratings by one expert, our Lead Game Developer, who designed the game, so they do not show how novice or intermediate players would play. A full human-AI game takes about 60 minutes, as long as a long game of rapid chess, so the games are few, and they mainly demonstrate the human-AI studies that \bgs{} makes possible; the BattleClient is also turn-based, so the Expert Human could think as long as they liked before each move. \bgs{} has one mode today. Support and expansion of \bgs{} are planned for at least the next three years, to make the benchmark suite richer and more challenging, to prevent saturation and accelerate research in MARL: new modes, new classes, abilities, maps, and scenarios, mana, bushes that hide agents, neutral camps that give a buff, and much more are already designed. \bgs{} is released under the Apache-2.0 license; vendored Mava and JaxMARL code (Apache-2.0) and on-policy MAPPO code (MIT) keep their licenses. The code will be made public on GitHub upon manuscript acceptance, and all training checkpoints of our baselines and of every tournament entrant will be hosted on Hugging Face at the same time.

%% file: sections/09_conclusion.tex

\section{Conclusion}
\label{sec:conclusion}

\balance

We presented \bgs{}, a JAX-native benchmark for heterogeneous team competition that is rich yet cheap to run. \emph{The Game} runs 5v5 \tdm{} at about 50,000 game steps per second with 1,024 parallel games on one RTX~5090. \emph{The Evaluation} combines verified endgame scenarios, an equal-budget tournament, and 11,192 metrics. \emph{The Tools} include six deep MARL baselines, three strong scripted opponents, an LLM-agent API, and human-play and replay clients. Our first results show that strong teams combine their abilities, yet no system solves a scenario, learned agents partner poorly with strangers and humans, and an expert human beats the tournament winner.

%% file: appendix.tex
\clearpage
\onecolumn
\FloatBarrier
\setcounter{section}{0}\setcounter{figure}{0}\setcounter{table}{0}\setcounter{lstlisting}{0}
\renewcommand{\thesection}{S\arabic{section}}
\renewcommand{\thetable}{S\arabic{table}}
\renewcommand{\thefigure}{S\arabic{figure}}
\renewcommand{\thelstlisting}{S\arabic{lstlisting}}
\def\theHsection{supp.\arabic{section}}
\def\theHtable{supp.\arabic{table}}
\def\theHfigure{supp.\arabic{figure}}
\def\theHlstlisting{supp.\arabic{lstlisting}}
\lstset{language={}, keywordstyle={}, commentstyle={}, breaklines=true, framesep=3pt}

\begin{center}{\LARGE\bfseries Supplementary Material}\end{center}
\medskip

\noindent This appendix holds the detail behind every pointer in the paper, for readers who want to check it. Section~\ref{sec:contents} lists what is here.

\FloatBarrier
\section*{Start Here: The Website}
\addcontentsline{toc}{section}{Start Here: The Website}
\label{sec:website}

The quickest way to see \bgs{} at work is our website, \weblink{https://marl-battlegrounds.com}{marl-battlegrounds.com}. Its 91 pages run in the browser with nothing to install: live demos of the real tools, every rule, class, map, and scenario, the Season~0 results, the trained models, and a guide for every tool. Each item below names the page to open.

\begin{itemize}
\item \textbf{Watch real games} (\weblink{https://marl-battlegrounds.com/\#home}{Home}, \weblink{https://marl-battlegrounds.com/\#replay-viewer}{The Replay Viewer}). A live copy of the analytical Replay Viewer plays recorded games, including Season~0's top two systems, recurrent IPPO against PQN-VDN, and a game of the Expert Human against recurrent IPPO. Switch between the Oracle view and any agent's POV, step tick by tick, ask for a plain explanation of any frame, and open the metrics.
\item \textbf{Play with the BattleClient} (\weblink{https://marl-battlegrounds.com/\#battleclient}{The BattleClient}). What it is for, how to start it, and demos of building a turn and a match. The BattleClient runs on your own computer; Getting Started shows how to install it.
\item \textbf{Learn the game} (\weblink{https://marl-battlegrounds.com/\#rules}{How A Match Works}, \weblink{https://marl-battlegrounds.com/\#classes}{Classes}, \weblink{https://marl-battlegrounds.com/\#maps}{Map Atlas}, \weblink{https://marl-battlegrounds.com/\#visual-key}{Visual Key}, \weblink{https://marl-battlegrounds.com/\#modes}{Task Modes}). Every rule as a picture you can play with; a page per class with live clips of every ability and its exact numbers; all 52 maps, with a Sight Lab in which you drag a champion and see what it can see; every symbol the clients draw; and the modes, Team Deathmatch today with King of the Hills and Capture the Flag under construction.
\item \textbf{See the scenarios solved} (\weblink{https://marl-battlegrounds.com/\#scenarios}{Scenarios}). A page for each of the eight scenarios, with its exact start and a step-through of its verified winning line, recorded from the simulator against the scenario's scripted opponent, plus every Big Nine member's results in each scenario.
\item \textbf{Read the Season~0 results} (\weblink{https://marl-battlegrounds.com/\#leaderboard}{Official Leaderboard}, \weblink{https://marl-battlegrounds.com/\#leaderboard-rules}{Leaderboard Rules}, \weblink{https://marl-battlegrounds.com/\#leaderboard-submit}{How To Submit}, \weblink{https://marl-battlegrounds.com/\#hall-of-champions}{Hall of Champions}). The full Season~0 table with Elo intervals, tiers, and win, draw, and loss counts; the matchup matrix; how Season~0 was played; the tournament rules; and how to submit a method.
\item \textbf{Download the models} (\weblink{https://marl-battlegrounds.com/\#models}{Model Library}). The six Season~0 baselines and the seven fixed recurrent MAPPO models the paper's hyperparameter search trained and validated against, 56.1~MB in all.
\item \textbf{Get started} (\weblink{https://marl-battlegrounds.com/\#start}{Getting Started}, \weblink{https://marl-battlegrounds.com/\#tutorials}{Tutorials}). Seven short steps from install to watching your own method fight, and three complete lessons: train your first team, bring your own learner, and self-play with leagues.
\item \textbf{Use every tool} (\weblink{https://marl-battlegrounds.com/\#guides}{Guides}). Forty-seven guides in nine groups: basics, training (built-in learners, self-play, partners, heterogeneous teams, curriculum, reward shaping, checkpoint selection, resuming, and Hydra sweeps), evaluation (validation, mixed teams, cross-play, and scenario tests), metrics (with the Metric Explorer, a search over all 11,192 measures, and an analysis notebook), tournaments and Elo, replays, LLM agents, measured performance, and reference pages: the API, the command line, the pinned and licensed sources of each learner, and how to reproduce the paper's experiments.
\item \textbf{Find the rest} (\weblink{https://marl-battlegrounds.com/\#resources}{Resources}, \weblink{https://marl-battlegrounds.com/\#news}{News}). How to cite \bgs{}, papers that use it, questions and answers, bugs and feature requests, the changelog, and news.
\end{itemize}

\noindent The site's links to the GitHub repository and the Hugging Face models will open upon manuscript acceptance; until then they show a short notice.

\FloatBarrier
\section{What Is in This Appendix}
\label{supp:s1}
\label{sec:contents}

\begin{itemize}
\item \textbf{This appendix.} Section~\ref{sec:table1} explains how Table~\ref{tab:comparison} of the paper was graded. Section~\ref{sec:supp-game} lists every rule, class value, and map of Team Deathmatch. Section~\ref{sec:engine} describes the engine, how we test it, and how exactly games repeat. Section~\ref{sec:supp-eval} gives the setup of every endgame scenario, the full tournament rules, how the tournament's seeds work, and how the Season~0 runs were played. Section~\ref{sec:supp-training} gives how we trained and timed the six baselines. Section~\ref{sec:supp-llm} gives the setup of the LLM team, the full prompt one agent reads, and its results against each member of the Big Nine. Sections~\ref{sec:supp-teamwork}, \ref{sec:supp-zsc}, and~\ref{sec:supp-expert} give the methods behind Sections~\ref{sec:exp-teamwork}, \ref{sec:exp-zsc}, and \ref{sec:exp-human} of the paper: how teamwork was tested against strength, how the zero-shot teams were built and measured, and how the Expert Human's games were played and read.
\item \textbf{\texttt{table1\_evidence.csv}.} The evidence for every cell of Table~1 for the sixteen compared benchmarks (Section~\ref{sec:evidence}). It is an ancillary file of this arXiv submission and opens in any spreadsheet program.
\item \textbf{The source code}, to be made public on GitHub upon manuscript acceptance: the simulator, the tools, the six baseline learners, the examples, and the experiment stages and scripts behind the paper's results. Its \texttt{README.md} shows how to run it.
\end{itemize}

\FloatBarrier
\section{Table 1: How the Comparison Was Graded}
\label{supp:s2}
\label{sec:table1}

Table~\ref{tab:comparison} of the paper compares \bgs{} with sixteen MARL benchmarks and games on twenty-two features. This section gives the method (Section~\ref{sec:method}), the rule behind every mark (Section~\ref{sec:rules}), the releases graded for each benchmark (Section~\ref{sec:scope}), and where to find the evidence for each cell (Section~\ref{sec:evidence}).

\FloatBarrier
\subsection{Method}
\label{supp:s2.1}
\label{sec:method}

\begin{itemize}
\item Each benchmark is graded on its own official releases as of September 2026: its paper, its official repositories, including official companion repositories, and its official documentation. Dota~2 has no released benchmark, so its column grades the OpenAI Five paper and the game as Valve ships it (Table~\ref{tab:scope}).
\item Only what the benchmark ships counts. Third-party add-ons, re-implementations and checkpoints never count, not even as partial. Outside documentation may still show how an official game itself behaves.
\item Every cell has at least one source with a verbatim quote, except two cells (tournament ratings for SMAX and VMAS) whose evidence is a search of the repository that found nothing. Code links point to a fixed commit, and all but five also to a line; paper links point to a page. \input{supplement/table1_check}
\item Downloads, links and competitions were checked when the table was graded. A second review then checked every reason against its sources; Table~1 and this appendix include its corrections.
\item In the \bgs{} column, the marks for a recurring official tournament and a released agent population are committed for the public release (the dagger in Table~1). The zero-shot coordination mark has no dagger because its tools work today; the downloadable pool of partners it uses is the released agent population, which that row's dagger covers.
\end{itemize}

\FloatBarrier
\subsection{Rules for Each Row}
\label{supp:s2.2}
\label{sec:rules}

Marks: \cmark{} yes, \pmark{} partial, \xmark{} no. Table~\ref{tab:rules} gives the rule for each row, in the order of Table~1.

\input{supplement/table1_rules}

\FloatBarrier
\subsection{Releases Graded for Each Benchmark}
\label{supp:s2.3}
\label{sec:scope}

Table~\ref{tab:scope} lists what was graded for each benchmark.

\input{supplement/table1_scope}

\FloatBarrier
\subsection{Evidence for Each Cell}
\label{supp:s2.4}
\label{sec:evidence}

The file \texttt{table1\_evidence.csv} holds the evidence for all 352 cells of the sixteen compared benchmarks, one row per quoted passage. Its columns are: the benchmark; the row number in Table~1; the feature; the mark; a short reason for the mark; the source link; and the verbatim quote from that source. To check a cell, filter the file by benchmark and feature.

\FloatBarrier
\section{Team Deathmatch: Rules, Classes, and Maps}
\label{supp:s3}
\label{sec:supp-game}

This section lists every rule, class value, and map of Team Deathmatch as the simulator defines them, in the order Sections~\ref{sec:preliminaries} and~\ref{sec:game} of the paper introduce them. Distances are in map units, measured between agents' centers, and every map is 20 by 10 units. Speeds are in units per tick. World of Warcraft's battlegrounds give \bgs{} its name and inspire its next two modes, King of the Hills and Capture the Flag. \bgs{} is not a full commercial game, and it is not real time: there is no time limit per turn.

\FloatBarrier
\subsection{Rules}
\label{supp:s3.1}
\label{sec:supp-game-rules}

\begin{figure}[h]
  \centering
  \includegraphics[width=0.495\columnwidth]{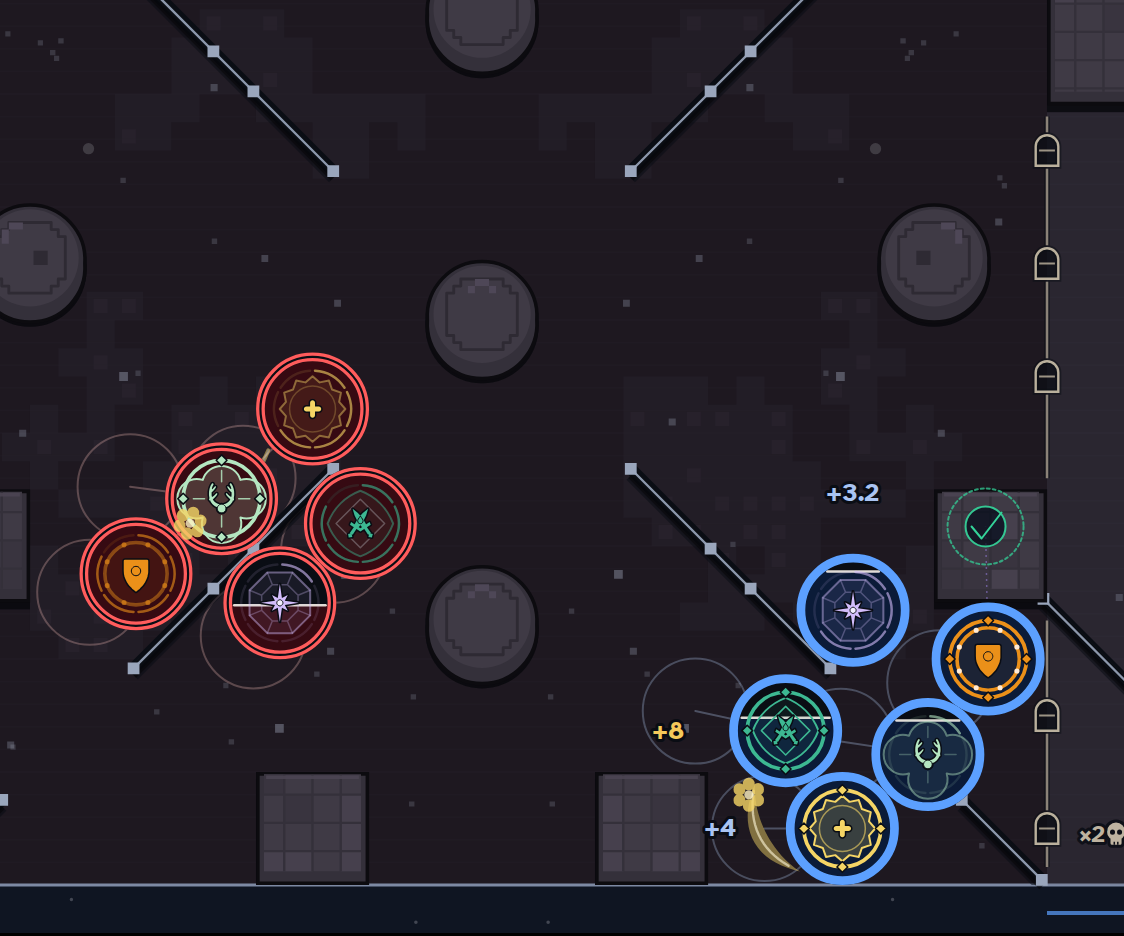}\hfill\includegraphics[width=0.495\columnwidth]{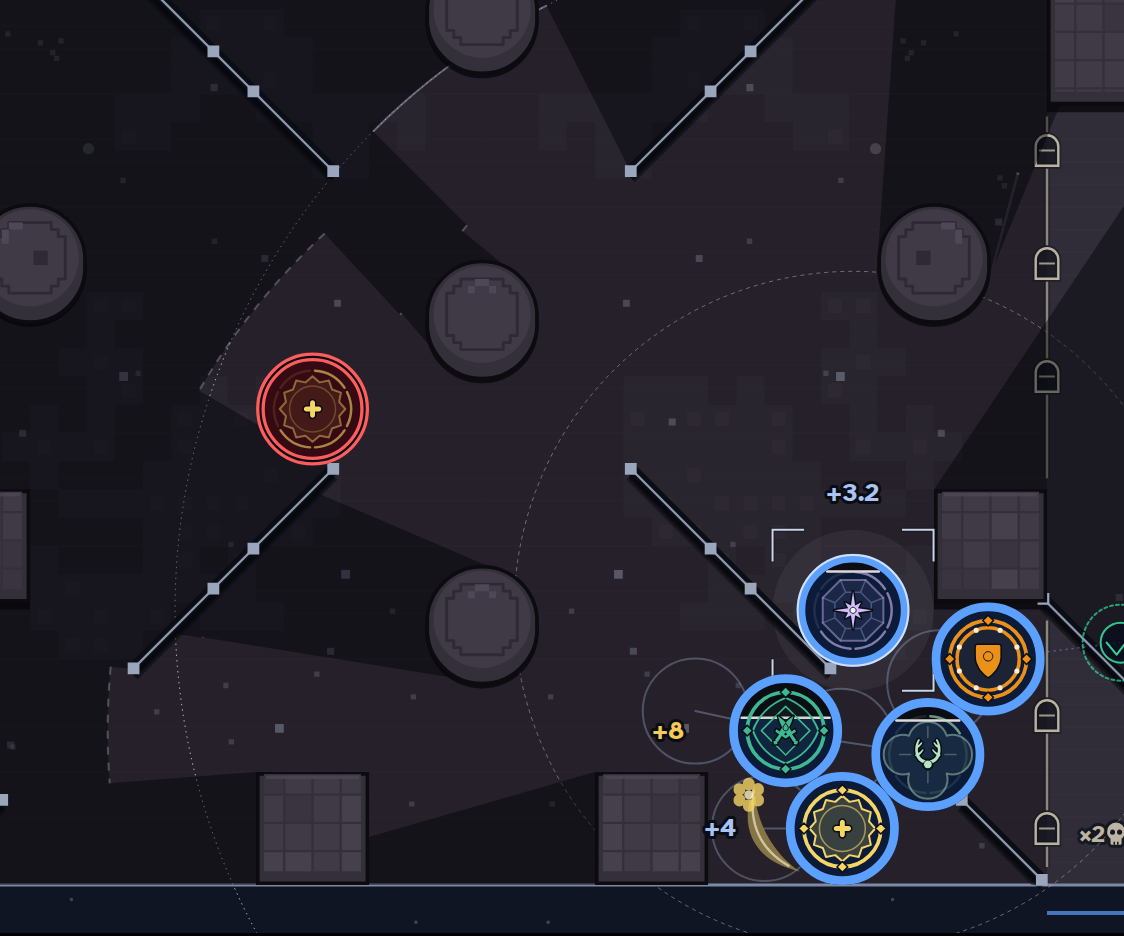}
  \caption{One moment of a test game in the Oracle view (left) and in Team A's Agent POV (right), where every enemy Team A cannot see vanishes.}
  \label{fig:supp-arena}
\end{figure}

Figure~\ref{fig:arena} of the paper shows training map 34 with the tournament's mirrored 5v5 on the spawn pads and each team's Red Zone. Figure~\ref{fig:supp-arena} shows one moment of a test game in the Oracle view and in Team~A's Agent POV, where every enemy Team~A cannot see vanishes.

\begin{itemize}
\item \textbf{Teams.} Each team has one to five agents of any classes, and the two teams may differ in size. The official tournament uses a mirrored 5v5 with one of each class, in which each class starts on the spawn pad that mirrors its opponent's. Every map has five spawn pads per team, along its left and right edges. In a batch of games, half of the games swap the two teams' spawn ends by default.
\item \textbf{Ticks.} A game is played in ticks. In each tick, every agent receives its observation and chooses an action, and all agents act at the same time. Transitions and resets are deterministic: all randomness comes from the policies and from any random choice of map, roster, or spawn made before a reset. A game lasts at most 300 ticks.
\item \textbf{Order within a tick.} Combat resolves first, from where agents stand at the start of the tick; the Warrior's Charge movement comes next, then ordinary movement. An agent's damage and healing in one tick are added together before deaths are checked, so a heal can save an ally from a blow that would otherwise kill it. Health never goes above the class maximum.
\item \textbf{Winning.} A team wins when it reaches 20 points with more points than its opponent. Equal scores at 20, and games still open after 300 ticks, are draws.
\item \textbf{Rewards.} When a game ends, every agent of the winning team receives $+1$ and every agent of the losing team $-1$; a draw gives every agent $0$. Dead agents share their team's result. All other ticks give $0$.
\item \textbf{Actions.} Each agent chooses a move (stay, or one of 8 compass directions), a target (none, or one of the 5 own-team or 5 enemy slots), and whether to use its Ultimate. Diagonal moves cover the same distance as straight ones. A Basic or a targeted Ultimate needs a target within range and within the agent's own sight, and a healer can target itself. An agent that uses its Ultimate does not also use its Basic that tick. A stunned, dead, or shielded agent cannot use abilities, and a stunned or dead agent cannot move.
\item \textbf{Legal actions.} Exact masks give each agent's legal choices every tick. An illegal move becomes stay, and an illegal target and Ultimate pair becomes no target and no Ultimate.
\item \textbf{Observations.} Each agent sees within its sight radius (Table~\ref{tab:classes}); obstacles block sight, and agents do not. By default, it also receives what its living teammates see. Every agent knows the map's obstacles, the current tick and the 300-tick limit, both teams' sizes and scores, the winning score, and the Red Zone depth, and it sees the last actions of the agents it can see. Inputs describe other agents as allies or enemies, never by team.
\item \textbf{Points.} Each death gives the other team 1 point, or 2 if the agent stood inside its own team's Red Zone at the start of the tick. A team's Red Zone is the strip within 5 units of its spawn edge.
\item \textbf{Obstacles and bodies.} Obstacles block sight, so Basics and targeted Ultimates cannot pass them; auras can. Every agent's body has a radius of 0.5. Obstacles and other agents' bodies block movement, except that the Warrior's Charge passes over obstacles and shielded agents pass through bodies.
\item \textbf{Respawn and shield.} Dead agents return to their own spawn pads in team waves every 5 ticks. For their first 3 ticks back, enemies cannot see or target them, allies cannot heal them, they cannot use abilities, auras neither come from them nor reach them, and they move at 2 units per tick and pass through other agents' bodies, ending clear of them when the shield ends. Agents start a game unshielded. The shield stops spawn camping, that is, killing agents as they respawn.
\end{itemize}

\FloatBarrier
\subsection{Classes}
\label{supp:s3.2}
\label{sec:supp-game-classes}

Each class has one Basic ability, one passive, and one Ultimate, and every Ultimate has a 30-tick cooldown. This section gives each class's full description, as the analytical Replay Viewer shows it, and Table~\ref{tab:classes} gives its values. The paper shortens three Ultimate names: Trap is Freezing Trap, Poison is Crippling Poison, and Salvation is Holy Word: Salvation. The classes and their abilities are inspired by real World of Warcraft classes and abilities. Figure~\ref{fig:classes} of the paper draws who counters and who helps whom: the counters form a cycle, so no class beats all the others.

\paragraph{Mage (Glass-Cannon Artillery, ranged).} The Mage has the strongest Basic in the game but the lowest health, so its team must keep it safe. The Mage and allies close to it deal more damage, and its Ultimate makes its own blows hit harder for the next few ticks.
\begin{itemize}
\item \textbf{Primary strength:} highest Basic damage among the damage-dealing classes. \textbf{Primary weakness:} lowest maximum health.
\item \textbf{Basic, Arcane Blast:} hits one enemy in reach. It is the strongest Basic in the game. Damage bonuses and reductions affect the damage dealt.
\item \textbf{Passive, Sorcerer's Empowerment:} while alive and without a spawn shield, the Mage gives off an aura that strengthens the attacks of living teammates within 2 units who also have no spawn shield, the Mage included. Each aura gives a 15\% damage bonus; overlapping auras multiply, up to 1.3225 times normal damage.
\item \textbf{Ultimate, Burst:} for 5 ticks, starting on the next tick, the Mage's damage rises by 50\% ($\times 1.5$).
\end{itemize}

\paragraph{Warrior (Tank, melee).} The Warrior has the most health in the game, and allies close to it take less damage. Its Basic is weak, so it needs allies to finish fights. Its Ultimate rushes it in one straight line up to a visible enemy, then hurts, stuns, and slows that enemy without pushing it back.
\begin{itemize}
\item \textbf{Primary strength:} highest maximum health. \textbf{Primary weakness:} second-lowest Basic damage among the damage-dealing classes.
\item \textbf{Basic, Shield Bash:} hits one enemy close to the Warrior. Damage bonuses and reductions affect the damage dealt.
\item \textbf{Passive, Guardian's Barrier:} while alive and without a spawn shield, the Warrior gives off an aura that protects living teammates within 2 units who also have no spawn shield, the Warrior included. Each barrier cuts incoming damage by 15\%; overlapping barriers each cut the damage left after the others, down to 72.25\% of the original.
\item \textbf{Ultimate, Charge:} moves the Warrior toward an enemy before ordinary movement, passing over obstacles. It deals 20 base damage, stuns for 1 tick, and slows the enemy to 50\% speed ($\times 0.5$) for 5 ticks. Damage bonuses and reductions affect the damage dealt.
\end{itemize}

\paragraph{Hunter (Scout and Crowd Controller, ranged).} The Hunter sees the farthest and its Basic reaches the farthest, but its arrows do the least damage and only slow their target a little. Its Ultimate holds one enemy in place for the longest stun in the game. Any damage on a later tick breaks the stun early, so allies must choose: wait, or strike.
\begin{itemize}
\item \textbf{Primary strength:} longest stun. \textbf{Primary weakness:} lowest Basic damage among the damage-dealing classes.
\item \textbf{Basic, Serrated Arrows:} hits one enemy in reach and slows it to 85\% speed ($\times 0.85$) for 1 tick. A new hit starts the slow again.
\item \textbf{Passive, Far Sight:} the Hunter sees everything within 6.5 units, farther than any other class. Obstacles still block its sight.
\item \textbf{Ultimate, Freezing Trap:} deals 10 base damage to one enemy and stuns it for 4 ticks. Damage bonuses and reductions affect the damage dealt. Damage breaks an existing Trap.
\end{itemize}

\paragraph{Rogue (Assassin, melee).} The Rogue moves faster than any other class, but it cannot turn invisible, blink, or teleport: it wins with speed, cover, and timing, and it relies on teammates to engage first, letting it flank the enemy's backline. Its Ultimate is the biggest single hit in the game; it also stuns, slows, and cuts the healing its target gets for a few ticks. It does no damage over time.
\begin{itemize}
\item \textbf{Primary strength:} highest base movement speed. \textbf{Primary weakness:} short attack range with no way to close the gap quickly.
\item \textbf{Basic, Backstab:} hits one enemy close to the Rogue. Damage bonuses and reductions affect the damage dealt.
\item \textbf{Passive, Phantom's Quickness:} a base movement speed of 1.3, the highest in the game, and out-of-combat healing after 3 ticks instead of 5.
\item \textbf{Ultimate, Crippling Poison:} deals 36 base damage to one enemy, stuns it for 1 tick, slows it to 50\% speed ($\times 0.5$) for 5 ticks, and halves its healing and out-of-combat healing for 4 ticks. Damage bonuses and reductions affect the damage dealt.
\end{itemize}

\paragraph{Priest (Healer, ranged).} The Priest is the only class that can heal, and it cannot deal damage at all, so its team must protect it. Its Basic mends one ally or itself and briefly lets that ally shake off most of a strong slow; stuns still hold. Its Ultimate is a huge heal that can save an ally from a killing blow in the same tick, but it never brings back the dead.
\begin{itemize}
\item \textbf{Primary strength:} the only class with healing abilities. \textbf{Primary weakness:} cannot deal damage.
\item \textbf{Basic, Lesser Heal:} heals one ally in reach, or the Priest itself, by 8 and gives it Blessing of Freedom. Healing reductions and the maximum-health limit apply.
\item \textbf{Passive, Blessing of Freedom:} for 1 tick after each Lesser Heal, slows cannot push the healed agent below 85\% of its base speed; stuns still apply.
\item \textbf{Ultimate, Holy Word: Salvation:} heals one ally, or the Priest itself, by 200, before healing reductions and the maximum-health limit, so it fills any class (100 under Poison).
\end{itemize}

All classes share these rules:

\begin{itemize}
\item \textbf{Damage and healing.} Damage and healing values are base values: Burst and the two auras multiply damage, and Poison halves healing.
\item \textbf{Auras.} Auras need no line of sight. Overlapping auras of the same kind multiply, up to 1.3225 times damage for Mage auras and down to 0.7225 times damage taken for Warrior auras; neither the tournament nor any scenario puts two agents of the same class on one team, so this applies only to custom rosters.
\item \textbf{Stuns and slows.} A stunned agent cannot move or use abilities. Slows from different abilities multiply, but speed never falls below 20\% of normal.
\item \textbf{Out-of-combat healing.} After 5 ticks out of combat (3 for the Rogue), an agent regains 4\% of its maximum health per tick. Dealing or taking damage puts an agent in combat, and so does healing, or being healed, while the healed agent is in combat.
\end{itemize}

\begin{table}[h]
\centering
\caption{Class values. Radii, ranges, sight, and speed are in map units; out-of-combat healing is a share of maximum health; ``--'' means none. An ally target includes the agent itself.}
\label{tab:classes}
\begin{tabular}{lrrrrr}
\toprule
 & Mage & Warrior & Hunter & Rogue & Priest \\
\midrule
Maximum health & 80 & 200 & 100 & 100 & 100 \\
Body radius & 0.5 & 0.5 & 0.5 & 0.5 & 0.5 \\
Speed (per tick) & 1.0 & 1.0 & 1.0 & 1.3 & 1.0 \\
Sight radius & 6 & 6 & 6.5 & 6 & 6 \\
Basic target & enemy & enemy & enemy & enemy & ally \\
Basic range & 3 & 1.5 & 3.5 & 1.5 & 3 \\
Basic damage & 13 & 8 & 6 & 12 & -- \\
Basic healing & -- & -- & -- & -- & 8 \\
Ultimate & Burst & Charge & Trap & Poison & Salvation \\
Ultimate target & self & enemy & enemy & enemy & ally \\
Ultimate range & self & 5.75 & 3 & 1.5 & 5.75 \\
Ultimate damage & -- & 20 & 10 & 36 & -- \\
Ultimate healing & -- & -- & -- & -- & 200 \\
Ultimate cooldown (ticks) & 30 & 30 & 30 & 30 & 30 \\
Out-of-combat delay (ticks) & 5 & 5 & 5 & 3 & 5 \\
Out-of-combat healing (per tick) & 4\% & 4\% & 4\% & 4\% & 4\% \\
\bottomrule
\end{tabular}
\end{table}

\FloatBarrier
\subsection{Maps}
\label{supp:s3.3}
\label{sec:supp-game-maps}

Figures~\ref{fig:supp-atlas} and~\ref{fig:supp-atlas-2} show all 52 maps, drawn from the map files. Each picture gives the map's number and split: the curriculum (maps 0--11, also used in training), training (12--41), validation (42--46), or test (47--51). The two spawn areas sit at the left and right ends. The built-in curriculum that uses maps 0--11 also grows teams from 1v1 to 5v5, and researchers can write their own curriculum stages that mix team sizes, maps, and winning scores.
\input{supplement/atlas}

\FloatBarrier
\section{Engine, API, and Testing}
\label{supp:s4}
\label{sec:engine}

Section~\ref{sec:game-api} of the paper describes the API. As in JaxMARL, \texttt{make}, \texttt{reset}, and \texttt{step} take and return the same things in the same order, and agents and their action and observation spaces are listed the same way. Listing~\ref{lst:supp-api} shows the full loop the paper describes: it runs 1,024 games of Team Deathmatch in parallel with random legal actions, compiled with \texttt{jit} and \texttt{lax.scan}, and a finished game restarts in place. This section then describes the engine under the API, how we test both, and how exactly games repeat.

\begin{lstlisting}[float=h, language=Python, caption={Running 1,024 games of \tdm{} in parallel with random legal actions. A finished game restarts in place, so the batch always holds 1,024 live games.}, label={lst:supp-api}]
import jax
import marl_battlegrounds as marl_bgs

env = marl_bgs.make("tdm", map_id=0, num_envs=1024)
obs, state = env.reset(jax.random.key(0))
def step(carry, _):
    state, key = carry
    key, k_act, k_step, k_reset = jax.random.split(key, 4)
    # random legal actions; swap in your method
    actions = env.sample_actions(k_act, state)
    obs, state, reward, done, info = env.step(k_step, state, actions)
    obs, state = env.reset_done(k_reset, state)  # restart finished games only
    return (state, key), reward
run = jax.jit(lambda s, k: jax.lax.scan(step, (s, k), None, length=128))
(state, key), rewards = run(state, jax.random.key(1))
\end{lstlisting}

\FloatBarrier
\subsection{Engine}
\label{supp:s4.1}
\label{sec:engine-core}

The engine is written from scratch in JAX. Each game's state is an explicit JAX tree with fixed array shapes: every game has ten agent slots, and the slots a roster does not use are padded. The engine's functions are pure, so \texttt{jit}, \texttt{vmap}, and \texttt{lax.scan} compose, and a finished game restarts in place without touching the others. Each rule has one owner: the engine alone defines the rules, observations, action masks, and rewards, and the environment around it handles batching, episode IDs, and optional metrics and replays without re-implementing any rule. The host checks configurations and authored game states before they run and rejects invalid ones with an error. Changing ordinary values, such as the map or the rosters, reuses the compiled program.

\FloatBarrier
\subsection{Tests}
\label{supp:s4.2}
\label{sec:engine-tests}

Table~\ref{tab:tests} counts our 9,583 Python tests by what they check. We counted them with \texttt{pytest --collect-only} and grouped them by test file. The clients' JavaScript unit and browser tests are not included.

\begin{table}[h]
\centering
\caption{Python tests by what they check, grouped by test file.}
\label{tab:tests}
\begin{tabular}{lr}
\toprule
Group & Tests \\
\midrule
\textbf{Engine} & \textbf{1,269} \\
\quad Configuration and game-state validation & 412 \\
\quad Combat and abilities & 271 \\
\quad Reset and step, transition facts, and action mappings & 201 \\
\quad Movement, collisions, and line of sight & 184 \\
\quad Observations, action masks, and action history & 122 \\
\quad Deaths, respawns, scoring, and game endings & 79 \\
\textbf{API} & \textbf{300} \\
\quad Systems and permitted inputs & 159 \\
\quad Environment calls & 91 \\
\quad Runnable examples & 50 \\
\textbf{Tools and content} & \textbf{8,014} \\
\quad Training tools and baselines & 2,682 \\
\quad Clients and viewers & 2,211 \\
\quad Evaluation, metrics, and replays & 1,210 \\
\quad Configuration, command line, and packaging & 634 \\
\quad Tournament & 603 \\
\quad Scenarios and scripted opponents & 366 \\
\quad LLM API & 153 \\
\quad Maps and packaged content & 88 \\
\quad Test tooling & 67 \\
\addlinespace
\textbf{Total} & \textbf{9,583} \\
\bottomrule
\end{tabular}
\end{table}

\FloatBarrier
\subsection{Correctness Checks}
\label{supp:s4.3}
\label{sec:engine-checks}

Besides checking single rules, the tests check that the parts agree with each other:
\begin{itemize}
\item \textbf{Public calls match the engine.} Games played through the public environment are compared with games played through the engine directly, batched games with games run under an outside \texttt{vmap}, and \texttt{lax.scan} loops with explicit steps.
\item \textbf{Fairness.} Collisions resolve the same way whatever order the agents are listed in. On all 52 maps, the left and right halves mirror each other exactly; a test checks this from both teams' sides.
\item \textbf{Information limits.} Each actor's input, its LLM prompt, and its point-of-view replay are checked to hold only what that actor may see.
\item \textbf{Scenario solutions.} Every packaged solution is replayed turn by turn. For Scenarios 3, 6, 7, and 8, variants that change one key move are checked to change the outcome.
\item \textbf{Metrics.} Metric columns are checked against the facts the engine reports for each transition.
\item \textbf{Baselines.} The PPO-family and QMIX learners are compared with independent calculations from the unchanged Mava code, and PQN-VDN with JaxMARL's.
\item \textbf{Command line and Python.} The same short games give the same results through both.
\item \textbf{Saved files.} Replays and results are read back and checked, including malformed and hostile files.
\end{itemize}

\FloatBarrier
\subsection{Continuous Integration}
\label{supp:s4.4}
\label{sec:engine-ci}

On every pull request and every push to the main branch, hosted runners run all of the Python tests in 19 parallel shards, Ruff format and lint checks, strict Pyright type checking, and the clients' static checks and browser tests.

\FloatBarrier
\subsection{Repeatability}
\label{supp:s4.5}
\label{sec:engine-repeat}

The engine is deterministic: the same seed, policies, and settings give the same game. In XLA's deterministic GPU mode (\texttt{XLA\_FLAGS=--xla\_gpu\_deterministic\_ops=true}), two runs of a 7,200-game round robin on test maps 47--51 (nine entrants, 200 games per pairing, 128 games in parallel on one RTX~5090, JAX 0.10.1, CUDA~13) matched in all 7,200 games at no measurable time cost, and recording replays or setting metrics to none or full changed no game.

\FloatBarrier
\section{Evaluation}
\label{supp:s5}
\label{sec:supp-eval}

Section~\ref{sec:evaluation} of the paper describes the two canonical evaluation modes. This section gives the setup of each endgame scenario, the tournament's full rules, how its seeds work, and what full metrics for a whole field cost.

\FloatBarrier
\subsection{Endgame Scenarios}
\label{supp:s5.1}
\label{sec:supp-scenarios}

Table~\ref{tab:supp-scenarios} lists each scenario's challenge, living rosters, clock, score, controller, and the length of its verified winning line. Table~\ref{tab:scenarios} of the paper shows which team skills each one tests. Each scenario loads as an exact snapshot of a game in progress. The eight scenarios have nine verified winning lines: Scenario~1 has two, both 3 ticks long, and every other scenario has one. Our website steps through a verified solution to every scenario.

\begin{table}[h]
\caption{The eight endgame scenarios. Team A v B: the living agents at the start (M Mage, W Warrior, H Hunter, R Rogue, P Priest); each team's other class slots start dead and return in the next respawn wave. Ticks: ticks left. Score: Team A against Team B; 20 points win. Opp.: the reactive controller. Sol.: ticks in the verified winning line.}
\label{tab:supp-scenarios}
\centering\small
\setlength{\tabcolsep}{3pt}
\begin{tabular}{rlccccc}
\toprule
ID & Challenge & Team A v B & Ticks & Score & Opp. & Sol. \\
\midrule
1 & Turn a doomed fight & HRP v MRP & 5 & 18--17 & TDM-Alpha & 3 \\
2 & Trust the healer & WP v MR & 5 & 18--19 & TDM-Alpha & 4 \\
3 & Moving cover & WH v R & 10 & 19--19 & TDM-Beta & 8 \\
4 & Finish before the stun & WR v MWP & 5 & 18--19 & TDM-Alpha & 4 \\
5 & Protect the healer & MWP v MRP & 5 & 17--19 & TDM-Beta & 5 \\
6 & Keep the trap & MHP v MWR & 5 & 17--19 & TDM-Alpha & 5 \\
7 & Retreat, then turn & MWP v MHR & 5 & 17--19 & TDM-Alpha & 4 \\
8 & Protect the bodyguard & WRP v RP & 5 & 18--19 & TDM-Beta & 5 \\
\bottomrule
\end{tabular}
\end{table}

\paragraph{Season~0 runs.} Section~\ref{sec:exp-scenarios} of the paper played each member of the Big Nine, and Random as a reference, as Team~A in every scenario: 256 games per scenario from seed 20261006, 128 games at a time, in XLA's deterministic mode. All 20,480 games took 297 seconds on one RTX~5090, loading and compiling every System included, and replaying them with full metrics took 367 seconds and gave the same outcomes. A two-sided Fisher's exact test compares each System's wins in each scenario with Random's. Table~\ref{tab:scenario-results} gives every system's wins, draws, and losses in each scenario.

\input{tables/scenario_results}

\FloatBarrier
\subsection{Tournament Rules}
\label{supp:s5.2}
\label{sec:supp-rules}

Section~\ref{sec:eval-tournament} of the paper summarizes the official tournament. Table~\ref{tab:supp-rules} gives the essentials of each rule in Version~1 of its rulebook, which governs Team Deathmatch from Season~1, starting in January~2027. The full rulebook, with submission details and the reasons behind each number, is part of the code release. Rules change only between seasons, announced at least 30 days before a season's cutoff, and each monthly table names the rulebook version it used. From Season~1, a new table is published on the first day of each month, and submissions for it close 72 hours before. The maintainers audit each trained policy, for example to check that its run never touched the test maps. Every entrant goes into one general pool of opponents, and each season's Big Nine also gets its own folder.

\begin{table}[h]
\caption{The tournament rules, Version~1, numbered as in the rulebook.}
\label{tab:supp-rules}
\centering\small
\begin{tabular}{@{}p{0.2\linewidth}p{0.76\linewidth}@{}}
\toprule
Rule & Essentials \\
\midrule
1. Who may enter & Any System that runs as pure JAX and exports to the portable file format: learned, scripted, search-based, hybrid, or built on a language model. The table shows each entry's type and whether it uses an outside pretrained model. Season~0's scripted opponents are reference opponents, exempt from Rules~2 and~3. \\
\addlinespace
2. Budget and machine & At most 255 million game steps and 8 hours, validation games included; the run stops at the first limit reached. One RTX~5090 (32\,GB), a 16-core AMD Ryzen~9 9950X3D, 48\,GB of RAM, and 50\,GB of disk, with no network. \\
\addlinespace
3. What a run may use & Released \bgs{} models are free to use as opponents or teammates; used as starting weights, as a frozen part of the entry, or as a teacher, their training steps count toward the budget. Outside pretrained models must have been public for at least 90 days, made by others, never trained on \bgs{}, and licensed for release. Nothing else learned from \bgs{} outside the run is allowed. Maps 0--46 are free to use; the run never touches test maps 47--51 or the scenarios, and nothing is chosen by results on them. \\
\addlinespace
4. Play limits & Playing itself with 512 games at once on one RTX~5090, an entry runs at least 5,000 game steps per second and uses at most 10\,GB of GPU memory. \\
\addlinespace
5. Information & Teammates may message each other one tick late; within a tick, each actor decides alone. Memory starts fresh every game, and weights never change during evaluation. Each agent sees its own sight plus its living teammates' sight, and no System reads hidden game data. \\
\addlinespace
6. Submitting & One submission per team per season, and one seat per team: a team's new System replaces its member only if it rates higher. Everything, admitted or not, is open-sourced under Apache-2.0 and released into the pool of opponents. \\
\addlinespace
7. The maintainers' run & The maintainers run each training script once, on an RTX~5090 with nothing else on it, and enter the run's last saved choice. Submissions are checked one at a time, in the order they arrive. \\
\addlinespace
8. Evaluation & The challenger plays Team~A in 200 mirrored 5v5 games against each member, 40 on each test map, with spawn ends swapped between paired games; first to 20 points wins, and a game still open at tick 300 is a draw. Games run in XLA's deterministic mode with a seed drawn at random when the challenger's games are played, after its training run, and published with its results. Full metrics are recorded from the first tick. Scenario results never feed Elo. \\
\addlinespace
9. Ratings & One joint Elo fit of the field plus the challenger, with an average of 1,200 and 95\% intervals. A challenger enters only if its Elo is higher than the lowest member's, and that member leaves. Every result is published, admitted or not. Tiers (Table~\ref{tab:tiers}) are labels beside the rating and never change rank or admission. \\
\addlinespace
10. Fair play & A rule's purpose binds too: getting around a limit breaks it. The maintainers' own entries follow every rule. A rule break found later removes the entry and bars its people for 12 seasons. \\
\bottomrule
\end{tabular}
\end{table}

\paragraph{Tiers.} As in ranked online games, each Elo also maps to one of eight tiers (Table~\ref{tab:tiers}), mainly so that policies of similar strength can be told apart at a glance. Each tier below Grandmaster splits into five divisions, 5 (lowest) to 1 (highest), 40 Elo apart. In Season~0, for example, four of our baselines share the Gold tier, and their divisions run from Gold~2 to Gold~4. Gold starts at the field's average Elo of 1,200, so tiers are relative: recurrent IPPO's Platinum~4 (1,441) means it would win about 80\% of decisive games against a policy of average strength. In general, a policy rated $R$ wins $1/(1 + 10^{-(R - 1{,}200)/400})$ of its decisive games, those not drawn, against a policy rated 1,200; this is the win chance in Table~\ref{tab:tiers}, and a 200-Elo lead gives about 76\%. Tiers never change rank or admission, and any claimed move between divisions is reported with its Elo gain and interval.

\input{tables/tiers}

\FloatBarrier
\subsection{Tournament Seeds}
\label{supp:s5.3}
\label{sec:supp-seeds}

Each tournament has one root seed (20261004 for Season~0's round robin). The schedule itself uses no randomness: for every pairing and test map it lists the pairs of games, and the two games of a pair share a map and a seed number but start from exchanged spawn ends. The seed number is the pair's number, unique in the tournament; Season~0's 7,200 games used 3,600 seed numbers, each in exactly two games.

The game's rules involve no chance: the simulator keeps reset and step keys in its interface, as JAX environments do, but ignores them, so the only random numbers in a game are the ones Systems draw. Every key is the root seed folded in turn with its use (actions or memory start), the game's seed number, and the tick. Action keys are then folded with the team, or with the agent's slot when the team acts through one policy call per agent. The keys therefore depend only on which game, tick, and team or agent they serve, not on how many games run in parallel or in which order, and the two games of a pair draw the same numbers for every action. Memory-start keys also fold in the game's own number, so a System whose memory starts at random, such as a pool choosing its member, starts the two games of a pair independently; for every other System, the only designed difference between the two games is which team starts where. A challenger's games come from a fixed list, numbered after the field's own, so every challenger meets each member on the same maps, spawn ends, and seed numbers.

From Season~1, each challenger's root seed is drawn at random when its games are played. The maintainers run each training script before its games, so no team can train on the seed, and the seed is published with the challenger's results, so anyone can replay its games. Games between two Systems that never sample do not depend on the seed at all; five of Season~0's nine members never sample, and for those games Rule~3's ban on the test maps is the protection. Each 95\% Elo interval is the middle 95\% of 5,000 refits; each refit draws whole spawn-swapped game pairs again at random within every pairing and map, from a published seed. Training seeds are separate: each team's training script sets its own.

\FloatBarrier
\subsection{Full Metrics for a Whole Field}
\label{supp:s5.4}
\label{sec:supp-field-metrics}

Section~\ref{sec:exp-teamwork} of the paper compares the whole Season~0 field with one \texttt{system\_metrics} call, which reads every entrant's games from its own side. Recording full metrics for all 7,200 games took 834 seconds and 392~MB of disk on one RTX~5090 with 128 games in parallel, against 488 seconds and 40~MB with priority metrics only; all 7,200 games repeated exactly. The analytical Replay Viewer searches all 11,192 metric columns.

\FloatBarrier
\section{Training the Baselines}
\label{supp:s6}
\label{sec:supp-training}

Section~\ref{sec:exp-setup} of the paper summarizes how we trained the six baselines. Every baseline can share one actor network across all agents, one per class, or none, and its training runs can be configured with Hydra~\cite{yadan2019hydra} recipes and resumed. This section gives the reward, the opponents, validation and checkpoint selection, the search, each baseline's final settings, and the compute used. The public record of the search, with every run and its results, is part of the code release.

\FloatBarrier
\subsection{Reward}
\label{supp:s6.1}
\label{sec:supp-reward}

Every Team~A agent receives the game's reward, $+1$ for a win, $-1$ for a loss, and 0 for a draw, plus the combat and support reward in Table~\ref{tab:supp-reward}. The built-in score shaping is off. We designed these terms with the full metrics on recurrent MAPPO: without the Red Zone terms, teams learned to hide at home, and paying for all healing, rather than only for useful healing, rewarded a Priest for healing allies who did not need it. The pooled 250-million-step recurrent MAPPO model on the validation panel was trained with the same reward.

\begin{table}[h]
\caption{The combat and support reward. Shared terms pay every Team~A agent the same amount; the other terms pay only the agent named.}
\label{tab:supp-reward}
\centering\small
\begin{tabular}{p{0.2\linewidth}lp{0.5\linewidth}}
\toprule
Term & Amount & Who receives it \\
\midrule
Kill credit & $+0.05$ per enemy death & Each agent whose attack landed on the enemy on the tick it died, and any Priest whose useful healing reached one of them on that tick \\
Red Zone kill & $+0.10$ more & The same agents, when the enemy dies in its own Red Zone \\
Damage & $+0.001$ per health point & The attacker; damage beyond the enemy's remaining health does not pay \\
Useful healing & $+0.001$ per health point & The Priest; healing beyond an ally's maximum health does not pay \\
Teammate death & $-0.05$ per death & Every agent (shared) \\
Red Zone death & $-0.10$ more & Every agent (shared), when the teammate dies in its own Red Zone \\
Healing range & $+0.001$ per pair per tick & Every agent (shared), for each living Priest and ally within the Priest's Basic range, both outside their own Red Zone \\
Own Red Zone & $-0.002$ per tick & Each living agent standing in its own Red Zone \\
\bottomrule
\end{tabular}
\end{table}

Recurrent MAPPO, recurrent IPPO, and feedforward IPPO learn from the per-agent form: each agent receives the shared terms and its own terms. QMIX and PQN-VDN learn one team value, and feedforward MAPPO's critic predicts one value for all five agents, so these three use the team form: every agent receives the shared terms once plus the team's total of every agent's own terms. Averaged over the team, each event then pays the team what it pays the agent it happens to in the per-agent form; a kill credit, for example, is worth $+0.05$ to the team, not $+0.01$. QMIX and PQN-VDN multiply the whole reward by their reward scale before learning. Logged rewards, results, and selection stay in game units.

\FloatBarrier
\subsection{Opponents}
\label{supp:s6.2}
\label{sec:supp-opponents}

Each new training game draws its opponent: the learner's current weights (60\%); one of its 12 newest saved copies, uniformly, with one copy saved every 1,376,256 game steps (20\%; the current weights until the first copy exists); six fixed recurrent MAPPO models trained for 25 to 40 million game steps, one with this reward and five with variants of it (15\%, 2.5\% each); and the scripted TDM-Alpha and TDM-Beta (5\%, 2.5\% each). PPO opponents sample their actions; the Q methods' current weights and copies act greedily except for their saved exploration rate. TDM-Gamma and the 250-million-step panel model are never training opponents. Training maps 0--41 are drawn uniformly; the curriculum was not used.

\FloatBarrier
\subsection{Validation and Checkpoint Selection}
\label{supp:s6.3}
\label{sec:supp-selection}

The validation panel is TDM-Alpha, TDM-Beta, TDM-Gamma, and the recurrent MAPPO model trained for 250,675,200 game steps, on maps 42, 43, 44, and 46. Map 45 is left out because the scripted opponents' pathing fails there, so wins on it would measure that failure rather than play. A routine check, at step 0 and after every 10\% of training, plays 8 games per map and opponent (128 games). The two best routine checks (step 0 excluded) and the final checkpoint then play 16 new games per map and opponent (256 games), and the one with the highest point margin represents the run. Point margin is our points minus the opponent's points per game, averaged within each map and opponent pair, with the 16 pairs weighted equally. Kills and deaths are reported beside every margin, because a team that avoids fights can score a deceptively good margin. Figure~\ref{fig:supp-curves} reads these routine checks over each baseline's whole lineage, from step~0 of its 40-million-step run to the end of its continuation: point margin and kills per game over the check's 128 games. Figure~\ref{fig:supp-wins} charts the share of its 96 games against the three scripted opponents that the baseline won: at 40 million steps no baseline wins half of these games, and each first does so after 51 to 166 million steps. On point margin against them, every baseline still lost at 40 million steps, by 1.7 (QMIX) to 12.5 (feedforward IPPO) points per game.

\begin{figure}[h]
\centering
\input{figures/curves}
\caption{Training curves of each baseline's chosen run at every routine validation check (Section~\ref{sec:exp-setup} of the paper): point margin (left) and kills (right) per game. Dotted line: the end of the 40-million-step search.}
\label{fig:supp-curves}
\end{figure}

\begin{figure}[h]
\centering
\input{figures/curves_wins}
\caption{The share of each routine validation check's 96 games against the three scripted opponents that each baseline won, one run per baseline. Solid line: half; dotted line: the end of the 40-million-step search.}
\label{fig:supp-wins}
\end{figure}

\FloatBarrier
\subsection{Search}
\label{supp:s6.4}
\label{sec:supp-search}

Two studies ran 84 runs of 40,108,032 game steps each, every recipe on the same three seeds (12259871, 20261001, and 20261002). Study~1 (54 runs) tested QMIX's rollout length and learning rate, PQN-VDN's batch size and rollout length, with optimizer steps per new game step held fixed, and one or two learning-rate changes for each PPO method. Study~2 (30 runs) changed one setting at a time around each Q method's best setting from Study~1. A change replaced its base only if its mean margin over the three seeds beat the base's by more than 2 points, it won on at least 2 of the 3 seeds, and, in Study~2, it averaged at least 2 kills per game against each scripted opponent. No PPO change passed. For both Q methods, scaling the reward by 10 passed with the largest gain; for QMIX, faster target copies and twice the training per game step passed too. Feedforward MAPPO trained one 40-million-step run on the team form of the reward with the same protocol.

Study~3 continued one 40-million-step run per method to 250,675,200 game steps: for five methods, the strongest single run of the search on margin, kills, and deaths together, and for feedforward MAPPO, its team-reward run. A continuation keeps its parent's learner, optimizer, replay, ongoing games, memory, and random streams. It checks every 5\% of the added steps, and the parent's checkpoints stay eligible for selection. PQN-VDN's learning rate, which had decayed to almost 0 by 40 million steps, restarts at $2.1 \times 10^{-4}$ and decays linearly to 250 million steps; the other methods keep their constant rates.

\begin{table}[h]
\caption{Settings of each baseline's Season~0 policy. Games: parallel games per update; T: rollout length in ticks. All six share their weights across the five agents and use the left spawn frame.}
\label{tab:supp-final}
\centering\small
\setlength{\tabcolsep}{3pt}
\begin{tabular}{lllcccr}
\toprule
Baseline & Value & Reward & Games & T & Learning rate & Checkpoint \\
\midrule
Recurrent IPPO & Local critic per agent & Per-agent & 512 & 32 & $10^{-4}$, critic $2.5 \times 10^{-4}$ & 250,675,200 \\
PQN-VDN & VDN sum of Q-values & Team, $\times 10$ & 768 & 128 & $2.5 \times 10^{-4}$, decaying & 240,156,672 \\
Recurrent MAPPO & Centralized critic & Per-agent & 512 & 32 & $10^{-4}$, critic $2.5 \times 10^{-4}$ & 219,103,232 \\
QMIX & QMIX mixer & Team & 768 & 32 & $10^{-4}$ & 219,095,040 \\
Feedforward IPPO & Local critic per agent & Per-agent & 512 & 32 & $10^{-4}$, critic $2.5 \times 10^{-4}$ & 219,103,232 \\
Feedforward MAPPO & Centralized critic & Team & 512 & 32 & $3 \times 10^{-4}$, critic $2.5 \times 10^{-4}$ & 208,568,320 \\
\bottomrule
\end{tabular}
\end{table}

Table~\ref{tab:supp-final} gives each baseline's main settings and the checkpoint that entered the tournament. The PPO baselines use discount 0.999, GAE $\lambda = 0.95$, clip 0.2, entropy 0.01, 4 epochs, 2 minibatches, value normalization, and an input scale of 0.01. QMIX uses discount 0.99, a hard target copy every 128 updates, exploration from 1.0 to 0.05 over 2 million steps, and a replay of 132,096 rows. PQN-VDN uses discount 0.99, TD $\lambda = 0.85$, BatchNorm, and exploration from 1.0 to 0.01 over its first 2.2 million steps.

\FloatBarrier
\subsection{Compute}
\label{supp:s6.5}
\label{sec:supp-compute}

Training took 101.6 GPU-hours on two RTX~5090s, one run per card: 44.8 for Study~1, 32.8 for Study~2, 0.6 for feedforward MAPPO's team-reward run, and 23.4 for the six continuations.

\paragraph{Timing.} The simulator's speed (Figure~\ref{fig:supp-speed}) is timed with random legal actions in 500-tick rollouts with automatic resets on test map 47, compiled into one program with \texttt{jit} and \texttt{lax.scan}: after one call to compile, five timed calls, of which we report the median. All timings use one RTX~5090 alone, JAX~0.10.1, and memory preallocation off, so memory readings show real use. Recurrent MAPPO's training speed is the median of the 37 updates after warm-up, out of 40 timed updates of 1,024 games by 32 ticks in pure self-play: 29,878 game steps per second, with 5.0\,GB of peak JAX memory (an 8.6\,GB allocator pool) and 4.8\,GB of peak RAM. The Season~0 runs trained 512 games at a time for the PPO baselines, with at most 3.1\,GB of JAX memory in use, and 768 for QMIX and PQN-VDN, with at most 15.2\,GB.

\begin{figure}[h]
  \centering
  \begin{tikzpicture}
    \begin{axis}[
      width=0.95\linewidth, height=3.4cm,
      xmode=log, ymode=log, log basis x=2,
      xlabel={Parallel 5v5 games}, ylabel={Game steps per second},
      xtick={32,64,128,512,1024}, xticklabels={32,64,128,512,1024},
      grid=major, grid style={gray!25},
      tick label style={font=\scriptsize}, label style={font=\small}]
      \addplot+[mark=*, thick] coordinates {
        (32,4919) (64,10122) (128,17760) (512,39921) (1024,50416)};
    \end{axis}
  \end{tikzpicture}
  \caption{Simulator throughput on one RTX~5090: random legal actions on test map 47, compiled 500-tick rollouts, compilation excluded.}
  \label{fig:supp-speed}
\end{figure}

\FloatBarrier
\section{The LLM Team}
\label{supp:s7}
\label{sec:supp-llm}

This section backs up the paper's subsection ``An LLM Team'': the setup of the LLM team, the full prompt that one agent reads, the team's results against each member of the Big Nine, and how its teamwork compares with its opponents'. One public experiment stage ran both experiments with the ordinary \nolinkurl{marl_bgs.evaluate} and \nolinkurl{marl_bgs.evaluate_scenarios} calls.

\FloatBarrier
\subsection{Setup}
\label{supp:s7.1}
\label{sec:supp-llm-setup}

Table~\ref{tab:supp-llm-setup} lists the settings. The team used the LLM API's defaults except for the fallback rule, which lets a game go on after a bad reply. The API can also add each actor's own earlier turns to its prompt; our team kept none. Researchers can swap in their own prompt, request, and reply format, turn on a model's thinking, and set the sampling, the reply length, and how many earlier turns each prompt keeps. One call turns any Python function from the agents' permitted views to actions into a System, so a model can also run any other way, such as through a provider's own API or on the researcher's own machine, still with our text view of the game and our legality check.

\begin{table}[h]
\caption{Settings of the LLM team.}
\label{tab:supp-llm-setup}
\centering\small
\begin{tabular}{@{}p{0.16\linewidth}p{0.8\linewidth}@{}}
\toprule
Setting & Value \\
\midrule
Model & Qwen3.5-9B: the original 16-bit weights at the revision pinned in our recipe. vLLM 0.30.0 converts them to 8-bit floating point (FP8) when it loads them. \\
Server & One server on its own RTX~5090 for each experiment, the two running at the same time, each using at most half its card's memory; a context of 16,384 tokens; up to 16 requests at once; prefix caching off; seed 0. \\
Requests & One call per agent per tick, when the agent has more than one legal choice. Temperature 0, top-p 1, seed 0, thinking off, at most 64 reply tokens, and a strict JSON schema with two fields, \texttt{move} and \texttt{combat}, whose values must be listed action names. \\
View & Text version \texttt{actor-text-v1}, in the frame where the agent's own spawn is on the left. Only the agent's own permitted view; no earlier turns. \\
Failures & A reply that is cut off, breaks the format, or picks a masked action becomes standing still with no attack, and is counted. \\
Games & Against the Big Nine: two games per member on each test map, spawn ends swapped, a 300-tick limit, seed 20261007. Scenarios: 256 games each, 128 side by side, seed 20261006. Full metrics in both. \\
Devices & The model on the GPU. The simulator and the Big Nine's policies on the CPU, because the model sets the pace; the rules are the same on either device. The two experiments shared the CPU during the scenarios' 22 minutes. \\
\bottomrule
\end{tabular}
\end{table}

\FloatBarrier
\subsection{The Prompt}
\label{supp:s7.2}
\label{sec:supp-llm-prompt}

Every prompt has the same parts: the public rules and abilities, the map's obstacles, the agent's own permitted view of units, sight, statuses, and scores, the previous actions it saw, and its legal menu, followed by the reply instruction. Below is the complete prompt that Team~A's Mage reads at tick 12 of a game between TDM-Alpha and TDM-Beta on test map 47: 14,303 characters in 101 lines, from \nolinkurl{llm.format_actor_view}. Nothing is shortened. Qwen3.5-9B's prompts against the Big Nine had the same format and averaged 4,038 tokens. For this turn, the legal reply
\begin{center}\texttt{\{"move":"east","combat":"enemy\_2\_basic"\}}\end{center}
would move the Mage east and use its Basic on enemy~2.

\begin{lstlisting}
MARL-BGs actor-text-v1; Frame: left
Rules: Choose one move and one complete combat pair from the legal lists. All actors commit both choices from the same decision. Damage and healing are summed before health clipping and death. Combat reads starting positions and statuses; Charge relocation precedes the chosen ordinary movement. New statuses first control the next decision. Legal choices can have no effect because of game physics. Previous actions describe accepted choices, not hits.
Abilities: Basic and Ultimate are alternatives. Priest Basic heals an ally, including self, and grants a temporary speed floor. Other Basics damage an enemy; Hunter Basic also slows it. Mage Ultimate gives self Burst damage amplification without a target. Warrior Ultimate damages, slows and stuns an enemy and moves the Warrior toward it. Hunter Ultimate damages and traps an enemy. Rogue Ultimate damages and applies slow, stun and reduced healing. Priest Ultimate heals an ally. capability_* fields describe what a unit can apply; other status fields describe effects currently on it. Use the displayed amounts, ranges and durations. Ultimate starts its full cooldown.
Sight and control: Direct sight needs a living observer and target, range and a clear line through static obstacles; units do not block sight. Shared sight does not expand your legal combat choices. Shielded enemies are hidden; shielded units cannot use or receive targeted combat. Stun prevents movement and combat. Different slows multiply with minimum speed fraction 0.2; Priest's speed floor can raise this. Existing effects age once per tick; reapplication refreshes durations. Positive damage breaks an existing Hunter Trap in the next state, but does not remove a Trap first applied that tick. Poison reduces healing and recovery; it does not damage each tick.
Team effects: Mage auras increase outgoing damage; Warrior auras reduce incoming damage. Living unshielded allies, including the emitter, benefit within aura range without a sight requirement. Matching auras multiply up to two emitters' strength. Recovery needs a countdown already at zero and no combat participation that tick; healing a unit already in combat counts as participation.
Units: Positions, radii, widths and heights are map units; movement speeds are map units per tick; angles are radians; health, damage and healing use health units. Durations, cooldowns and clocks count ticks. Multipliers and fractions are dimensionless. North increases y; east increases x. Diagonal directions have unit length too. Sight and interaction radii use center-to-center distance, including their boundary.
Geometry: Obstacle type 1 is a circular pillar and 2 is a rectangular wall. x/y give the center; radius describes a pillar, and width/height are full wall lengths. theta rotates a wall counterclockwise from the x axis. Only active obstacles participate. Bodies have their displayed radius; collisions and map boundaries can shorten or redirect movement.
Roster: Ally/enemy numbers are stable rows 0..4. Class IDs are 1=Mage, 2=Warrior, 3=Hunter, 4=Rogue, 5=Priest; 0 means unused. Public alive flags reveal no hidden position. Source permission does not mean the source currently sees a unit. Respawn clock zero means a wave is due at transition end; a dead actor cannot act before it respawns.
Scoring: In Team Deathmatch a death gives the opponent 1 point, or 2 in the victim's own Red Zone, using its starting position. The zone is the full-height strip beside that team's spawn edge, including its boundary. Depth is in context; zero disables it. A team wins only when it reaches the threshold with more points than the opponent. A simultaneous threshold tie, or the horizon without a threshold winner, is a draw.
Encoding: Unlisted numeric fields and padded rows are positive zero. Zero means a whole positive-zero row. Equal unit definitions share an ID; visibility and source assignments remain separate. Current objectives have 8 rows of 12 columns; unexpected nonzero reserved values are printed. Hidden means no direct visible unit row; Not seen means no previous accepted action was supplied. All numbers preserve their original values.
Map:
Obstacle 0: type=1 x=4 y=7.5 radius=0.5 active=1
Obstacle 1: type=1 x=10 y=9.5 radius=0.5 active=1
Obstacle 2: type=1 x=6 y=0.5 radius=0.5 active=1
Obstacle 3: type=1 x=10 y=0.5 radius=0.5 active=1
Obstacle 4: type=1 x=14 y=0.5 radius=0.5 active=1
Obstacle 5: type=1 x=16 y=2.5 radius=0.5 active=1
Obstacle 6: type=1 x=16 y=7.5 radius=0.5 active=1
Obstacle 7: type=1 x=7.8 y=2 radius=0.5 active=1
Obstacle 8: type=1 x=12.2 y=8 radius=0.5 active=1
Obstacle 9: type=1 x=6 y=9.5 radius=0.5 active=1
Obstacle 10: type=1 x=12.2 y=2 radius=0.5 active=1
Obstacle 11: type=1 x=7.8 y=8 radius=0.5 active=1
Obstacle 12: type=1 x=14 y=9.5 radius=0.5 active=1
Obstacle 13: type=1 x=4 y=2.5 radius=0.5 active=1
Obstacle 14: type=2 x=8 y=5 width=0.5 height=1.5 active=1
Obstacle 15: type=2 x=12 y=5 width=0.5 height=1.5 active=1
Obstacle 16: type=2 x=6 y=6.5 width=1 height=0.5 active=1
Obstacle 17: type=2 x=14 y=6.5 width=1 height=0.5 active=1
Obstacle 18: type=2 x=6 y=3.5 width=1 height=0.5 active=1
Obstacle 19: type=2 x=14 y=3.5 width=1 height=0.5 active=1
Obstacle 20: type=2 x=2 y=7.9 width=0.01 height=2 active=1
Obstacle 21: type=2 x=3 y=5 width=0.01 height=2 active=1
Obstacle 22: type=2 x=2 y=2.1 width=0.01 height=2 active=1
Obstacle 23: type=2 x=18 y=7.9 width=0.01 height=2 active=1
Obstacle 24: type=2 x=17 y=5 width=0.01 height=2 active=1
Obstacle 25: type=2 x=18 y=2.1 width=0.01 height=2 active=1
Obstacle 26: type=2 x=10 y=1.5 width=0.01 height=1 active=1
Obstacle 27: type=2 x=10 y=8.5 width=0.01 height=1 active=1
Obstacle 28: type=2 x=3 y=5 width=1 height=1 theta=0.7853982 active=1
Obstacle 29: type=2 x=17 y=5 width=1 height=1 theta=0.7853982 active=1
Obstacle 30: type=1 x=10 y=2.5 radius=0.5 active=1
Obstacle 31: type=1 x=10 y=7.5 radius=0.5 active=1
Context: current_timestep=12 episode_horizon=300 map_width=20 map_height=10 ally_team_size=5 enemy_team_size=5 is_tdm=1 tdm_score_threshold=20 tdm_red_zone_depth=5
Self row: ally_0
Objectives: Zero
Units (unlisted fields are zero):
unit_0: x=8.313756 y=3.0042372 radius=0.5 active=1 alive=1 class_id=1 base_movement_speed=1 effective_movement_speed=1 observation_radius=6 basic_interaction_radius=3 current_health=80 max_health=80 slow_warrior_charge_multiplier=1 slow_hunter_basic_multiplier=1 slow_rogue_poison_multiplier=1 anti_heal_rogue_poison_multiplier=1 damage_amplification_mage_aura_multiplier=1.15 damage_mitigation_warrior_aura_multiplier=1 capability_basic_damage=13 capability_ultimate_cooldown_duration=30 capability_damage_amplification_mage_burst_duration=5 capability_damage_amplification_mage_burst_multiplier=1.5 capability_damage_amplification_mage_aura_radius=2 capability_damage_amplification_mage_aura_multiplier=1.15 capability_out_of_combat_delay_steps=5 capability_out_of_combat_health_regen_fraction_per_step=0.04
unit_1: x=5.885046 y=1.4976649 radius=0.5 active=1 alive=1 class_id=2 base_movement_speed=1 effective_movement_speed=1 observation_radius=6 basic_interaction_radius=1.5 ultimate_interaction_radius=5.75 current_health=200 max_health=200 slow_warrior_charge_multiplier=1 slow_hunter_basic_multiplier=1 slow_rogue_poison_multiplier=1 anti_heal_rogue_poison_multiplier=1 damage_amplification_mage_aura_multiplier=1 damage_mitigation_warrior_aura_multiplier=0.85 capability_basic_damage=8 capability_ultimate_cooldown_duration=30 capability_slow_warrior_charge_duration=5 capability_slow_warrior_charge_multiplier=0.5 capability_stun_warrior_charge_duration=1 capability_damage_mitigation_warrior_aura_radius=2 capability_damage_mitigation_warrior_aura_multiplier=0.85 capability_ultimate_damage=20 capability_out_of_combat_delay_steps=5 capability_out_of_combat_health_regen_fraction_per_step=0.04
unit_2: x=9.048565 y=3.7180176 radius=0.5 active=1 alive=1 class_id=3 base_movement_speed=1 effective_movement_speed=1 observation_radius=6.5 basic_interaction_radius=3.5 ultimate_interaction_radius=3 current_health=100 max_health=100 slow_warrior_charge_multiplier=1 slow_hunter_basic_multiplier=1 slow_rogue_poison_multiplier=1 anti_heal_rogue_poison_multiplier=1 damage_amplification_mage_aura_multiplier=1.15 damage_mitigation_warrior_aura_multiplier=1 capability_basic_damage=6 capability_ultimate_cooldown_duration=30 capability_slow_hunter_basic_duration=1 capability_slow_hunter_basic_multiplier=0.85 capability_stun_hunter_trap_duration=4 capability_ultimate_damage=10 capability_out_of_combat_delay_steps=5 capability_out_of_combat_health_regen_fraction_per_step=0.04
unit_3: x=7.534366 y=3.6408086 radius=0.5 active=1 alive=1 class_id=4 base_movement_speed=1.3 effective_movement_speed=1.3 observation_radius=6 basic_interaction_radius=1.5 ultimate_interaction_radius=1.5 current_health=100 max_health=100 slow_warrior_charge_multiplier=1 slow_hunter_basic_multiplier=1 slow_rogue_poison_multiplier=1 anti_heal_rogue_poison_multiplier=1 damage_amplification_mage_aura_multiplier=1.15 damage_mitigation_warrior_aura_multiplier=1 capability_basic_damage=12 capability_ultimate_cooldown_duration=30 capability_slow_rogue_poison_duration=5 capability_slow_rogue_poison_multiplier=0.5 capability_stun_rogue_poison_duration=1 capability_anti_heal_rogue_poison_duration=4 capability_anti_heal_rogue_poison_multiplier=0.5 capability_ultimate_damage=36 capability_out_of_combat_delay_steps=3 capability_out_of_combat_health_regen_fraction_per_step=0.04
unit_4: x=11.712831 y=3.0268838 radius=0.5 is_enemy=1 active=1 alive=1 class_id=1 base_movement_speed=1 effective_movement_speed=1 observation_radius=6 basic_interaction_radius=3 current_health=80 max_health=80 slow_warrior_charge_multiplier=1 slow_hunter_basic_multiplier=1 slow_rogue_poison_multiplier=1 anti_heal_rogue_poison_multiplier=1 damage_amplification_mage_aura_multiplier=1.15 damage_mitigation_warrior_aura_multiplier=1 capability_basic_damage=13 capability_ultimate_cooldown_duration=30 capability_damage_amplification_mage_burst_duration=5 capability_damage_amplification_mage_burst_multiplier=1.5 capability_damage_amplification_mage_aura_radius=2 capability_damage_amplification_mage_aura_multiplier=1.15 capability_out_of_combat_delay_steps=5 capability_out_of_combat_health_regen_fraction_per_step=0.04
unit_5: x=11.0110035 y=3.7395618 radius=0.5 is_enemy=1 active=1 alive=1 class_id=3 base_movement_speed=1 effective_movement_speed=1 observation_radius=6.5 basic_interaction_radius=3.5 ultimate_interaction_radius=3 current_health=100 max_health=100 slow_warrior_charge_multiplier=1 slow_hunter_basic_multiplier=1 slow_rogue_poison_multiplier=1 anti_heal_rogue_poison_multiplier=1 damage_amplification_mage_aura_multiplier=1.15 damage_mitigation_warrior_aura_multiplier=1 capability_basic_damage=6 capability_ultimate_cooldown_duration=30 capability_slow_hunter_basic_duration=1 capability_slow_hunter_basic_multiplier=0.85 capability_stun_hunter_trap_duration=4 capability_ultimate_damage=10 capability_out_of_combat_delay_steps=5 capability_out_of_combat_health_regen_fraction_per_step=0.04
unit_6: x=4.826409 y=3.5277374 radius=0.5 active=1 alive=1 class_id=5 base_movement_speed=1 effective_movement_speed=1 observation_radius=6 basic_interaction_radius=3 ultimate_interaction_radius=5.75 current_health=100 max_health=100 slow_warrior_charge_multiplier=1 slow_hunter_basic_multiplier=1 slow_rogue_poison_multiplier=1 anti_heal_rogue_poison_multiplier=1 slow_floor_priest_blessing_of_freedom_duration=1 slow_floor_priest_blessing_of_freedom_fraction=0.85 damage_amplification_mage_aura_multiplier=1 damage_mitigation_warrior_aura_multiplier=1 capability_basic_healing=8 capability_ultimate_cooldown_duration=30 capability_slow_floor_priest_blessing_of_freedom_duration=1 capability_slow_floor_priest_blessing_of_freedom_fraction=0.85 capability_ultimate_healing=200 capability_out_of_combat_delay_steps=5 capability_out_of_combat_health_regen_fraction_per_step=0.04
Self: unit_0
ally_0: Visible unit_0
ally_1: Visible unit_1
ally_2: Visible unit_2
ally_3: Visible unit_3
ally_4: Hidden Zero
enemy_0: Visible unit_4
enemy_1: Hidden Zero
enemy_2: Visible unit_5
enemy_3: Hidden Zero
enemy_4: Hidden Zero
Source ally_0: Unavailable; No visible rows
Source ally_0 objectives: Zero
Source ally_1: Allowed; ally_0=Visible unit_0; ally_1=Visible unit_1; ally_2=Visible unit_2; ally_3=Visible unit_3; ally_4=Visible unit_6
Source ally_1 objectives: Zero
Source ally_2: Allowed; ally_0=Visible unit_0; ally_1=Visible unit_1; ally_2=Visible unit_2; ally_3=Visible unit_3; enemy_0=Visible unit_4; enemy_2=Visible unit_5
Source ally_2 objectives: Zero
Source ally_3: Allowed; ally_0=Visible unit_0; ally_1=Visible unit_1; ally_2=Visible unit_2; ally_3=Visible unit_3; enemy_0=Visible unit_4; enemy_2=Visible unit_5
Source ally_3 objectives: Zero
Source ally_4: Allowed; ally_1=Visible unit_1; ally_4=Visible unit_6
Source ally_4 objectives: Zero
Unlisted source candidates: Hidden Zero
Spawn lifecycle (team axes: own team, opponent):
Spawn pads: [[[0.5,1],[0.5,3],[0.5,5],[0.5,7],[0.5,9]],[[19.5,1],[19.5,3],[19.5,5],[19.5,7],[19.5,9]]]
Shield ticks: [[0,0,0,0,0],[0,0,0,0,0]]
Configured shield ticks: 3
Shield speed: 2
Wave periods: [5,5]
Wave countdowns: [2,2]
Active roster: [[1,1,1,1,1],[1,1,1,1,1]]
Alive roster: [[1,1,1,1,1],[1,1,1,1,1]]
Class IDs: [[1,2,3,4,5],[1,2,3,4,5]]
Previous accepted actions (targets already use your roster names):
ally_0: move=east target=target_none ultimate=No
ally_1: move=east target=target_none ultimate=No
ally_2: move=east target=target_none ultimate=No
ally_3: move=southeast target=target_none ultimate=No
ally_4: Not seen
enemy_0: move=west target=target_none ultimate=No
enemy_1: Not seen
enemy_2: move=west target=target_none ultimate=No
enemy_3: Not seen
enemy_4: Not seen
Legal moves: stay, north, south, east, west, northeast, northwest, southeast, southwest
Legal combat pairs: no_combat, ultimate, enemy_2_basic
Combat: no_combat means no target and no Ultimate; ultimate means an untargeted Ultimate. Target names keep their roster meaning. The target and Ultimate marginal masks are the unions of the listed complete pairs.
Reply with exactly {"move":"<legal move>","combat":"<legal combat>"}.
\end{lstlisting}

\FloatBarrier
\subsection{Results Against Each Member of the Big Nine}
\label{supp:s7.3}
\label{sec:supp-llm-results}

Table~\ref{tab:supp-llm-field} gives the team's results against each member, ten games each. It made 62,629 calls: each read 4,038 tokens and wrote 12.1 on average, and 99.97\% of the replies were legal actions: 16 were cut off at the 64-token reply limit while padding the JSON with blank space, and 2 chose a masked action. A call took 1.67 seconds at the median, a tick of two games side by side took 1.70 seconds, and all 90 games took 3.5 hours. In the scenarios, the team made 9,610 calls, of which one reply was cut off the same way; each read 3,011 tokens and wrote 12.2 on average, and the 2,048 games took 22 minutes. There, 128 games shared 16 request slots, so a call waited 19.9 seconds at the median.

\begin{table}[h]
\caption{Results of the LLM team (Qwen3.5-9B, Team~A) against each member of the Big Nine: ten games each, two per test map. Margin, kills, deaths, and ticks are averages per game; Legal is the share of model replies that were legal actions; Seconds is each pair's wall time divided by its two games, which played side by side.}
\label{tab:supp-llm-field}
\centering\small
\begin{tabular}{@{}lcrrrrrr@{}}
\toprule
Opponent & W--D--L & Margin & Kills & Deaths & Ticks & Legal & Seconds \\
\midrule
Rec.~IPPO & 0--0--10 & $-20.7$ & 0.1 & 11.2 & 87.9 & 99.98\% & 87.5 \\
PQN-VDN & 0--2--8 & $-16.4$ & 0.0 & 9.3 & 150.5 & 100.00\% & 150.8 \\
Rec.~MAPPO & 0--0--10 & $-20.5$ & 0.0 & 11.2 & 111.4 & 100.00\% & 109.8 \\
QMIX & 0--2--8 & $-16.0$ & 0.4 & 9.4 & 148.9 & 100.00\% & 140.5 \\
FF~IPPO & 0--2--8 & $-18.0$ & 0.0 & 9.5 & 151.7 & 99.93\% & 143.6 \\
FF~MAPPO & 0--0--10 & $-20.8$ & 0.0 & 12.0 & 152.7 & 99.96\% & 149.8 \\
TDM-Beta & 0--3--7 & $-15.1$ & 0.0 & 10.9 & 190.1 & 99.97\% & 172.1 \\
TDM-Gamma & 0--0--10 & $-20.5$ & 0.0 & 12.1 & 101.1 & 99.98\% & 95.5 \\
TDM-Alpha & 0--4--6 & $-14.2$ & 0.0 & 11.7 & 223.9 & 99.95\% & 200.8 \\
\bottomrule
\end{tabular}
\end{table}

\FloatBarrier
\subsection{Teamwork Compared With Its Opponents}
\label{supp:s7.4}
\label{sec:supp-llm-teamwork}

The same games show how the team's coordination differs from its opponents'. Table~\ref{tab:supp-llm-teamwork} compares the two teams in each game, per opponent and over all 90 games. Its replies explain part of the gap: in 94.7\% of its legal replies against the Big Nine the model chose no attack, and 98.8\% of its moves went east, toward the enemy's side in its own frame (61.8\%), or north (37.0\%). In the scenarios, 84.8\% of its replies chose no attack.

\begin{table}[h]
\caption{Teamwork of the LLM team (LLM) and its opponent (Opp.) in the same games, averaged per game: Basic uses; damage done; ticks in which two or more teammates dealt damage (Together); kills shared by two or more agents (Shared); mean distance between living teammates in units (Spread); and deaths inside the team's own Red Zone, each worth 2 points to the enemy (Home deaths).}
\label{tab:supp-llm-teamwork}
\centering\small
\setlength{\tabcolsep}{3pt}
\begin{tabular}{@{}lrrrrrrrrrrrr@{}}
\toprule
& \multicolumn{2}{c}{Basic uses} & \multicolumn{2}{c}{Damage} & \multicolumn{2}{c}{Together} & \multicolumn{2}{c}{Shared} & \multicolumn{2}{c}{Spread} & \multicolumn{2}{c}{Home deaths} \\
Opponent & LLM & Opp. & LLM & Opp. & LLM & Opp. & LLM & Opp. & LLM & Opp. & LLM & Opp. \\
\midrule
Rec.~IPPO & 32.1 & 215.3 & 157 & 1,510 & 1.4 & 43.1 & 0.0 & 9.5 & 6.9 & 2.2 & 9.6 & 0.0 \\
PQN-VDN & 31.5 & 256.5 & 163 & 1,317 & 1.1 & 43.8 & 0.0 & 7.9 & 7.6 & 2.4 & 7.1 & 0.0 \\
Rec.~MAPPO & 40.5 & 259.8 & 149 & 1,706 & 1.3 & 51.6 & 0.0 & 8.9 & 7.5 & 2.5 & 9.3 & 0.0 \\
QMIX & 36.3 & 243.4 & 192 & 1,382 & 1.6 & 45.0 & 0.0 & 7.5 & 6.3 & 3.5 & 7.0 & 0.0 \\
FF~IPPO & 39.2 & 267.5 & 141 & 1,395 & 1.6 & 42.7 & 0.0 & 7.9 & 7.7 & 2.5 & 8.5 & 0.0 \\
FF~MAPPO & 47.0 & 302.0 & 204 & 1,739 & 0.9 & 56.2 & 0.0 & 8.4 & 8.2 & 3.8 & 8.8 & 0.0 \\
TDM-Beta & 17.2 & 302.3 & 106 & 1,425 & 0.8 & 38.5 & 0.0 & 8.5 & 5.9 & 2.7 & 4.2 & 0.0 \\
TDM-Gamma & 24.8 & 244.8 & 131 & 1,701 & 1.1 & 47.4 & 0.0 & 9.0 & 6.1 & 3.9 & 8.4 & 0.0 \\
TDM-Alpha & 12.2 & 348.5 & 125 & 1,557 & 0.1 & 43.3 & 0.0 & 8.7 & 4.9 & 2.8 & 2.5 & 0.0 \\
\addlinespace
All & 31.2 & 271.1 & 152 & 1,526 & 1.1 & 45.7 & 0.0 & 8.5 & 6.8 & 2.9 & 7.3 & 0.0 \\
\bottomrule
\end{tabular}
\end{table}

\FloatBarrier
\section{How Agents Cooperate}
\label{supp:s8}
\label{sec:supp-teamwork}

Section~\ref{sec:exp-teamwork} of the paper asks which teamwork goes with strength across the Season~0 field. The tournament saves only results, points, kills, and deaths, so we replayed all 7,200 games from the saved schedule and seeds, recording full metrics for every game in XLA's deterministic mode; every game repeated exactly (Section~\ref{sec:supp-field-metrics}). One \texttt{system\_metrics} call reads each entrant's 1,600 games from its own side and pools them. Counts become rates per 100 ticks, because an entrant's average game lasts 205 to 252 ticks, and shares and averages pool all games. The 95\% intervals in Figure~\ref{fig:teamwork} of the paper resample each entrant's spawn-swapped game pairs within each opponent and map, 2,000 times.

For each of the 145 team-level metrics, $\rho$ is Spearman's rank correlation with Elo over the nine entrants, and its exact $p$ is the share of all 362,880 orderings of the nine ratings whose correlation is at least as strong. Testing the 138 metrics that vary invites chance hits, so Table~\ref{tab:supp-teamwork-tests} also gives Holm's correction over the 138 and the Westfall-Young max-statistic correction over the 137 with a value for every entrant; both raise the bar for the screen as a whole. The six measures of Figure~\ref{fig:teamwork} of the paper are derived from these metrics and tested the same way.

\begin{table}[h]
\caption{Team measures and their tests across the nine Season~0 entrants. *: per 100 ticks; the others are shares or averages. Holm and Max-$t$: $p$ after Holm's and the Westfall-Young max-statistic corrections (bold: below 0.05).}
\label{tab:supp-teamwork-tests}
\centering\small
\input{supplement/teamwork_tests}
\end{table}

\FloatBarrier
\section{Zero-Shot Coordination}
\label{supp:s9}
\label{sec:supp-zsc}

Section~\ref{sec:exp-zsc} of the paper puts a stranger into recurrent IPPO's team one class at a time. Each of the 31 teams played every member of the Season~0 field 200 times on test maps 47--51, 40 per map with spawn ends swapped within each pair, with full metrics and the same seed, 20261007, for every team, so game~$n$ of every team starts from the same opponent, map, spawn end, and seed. The 55,800 games took 39 minutes on one RTX~5090, 128 at a time, in XLA's deterministic mode. Each team is compared with recurrent IPPO alone game start by game start, and the 95\% intervals of Figure~\ref{fig:supp-strangers} resample whole seed pairs within each opponent, 2,000 times from seed 20261007. Figure~\ref{fig:supp-strangers} charts every stranger's cost.

\begin{figure}[h]
  \centering
  \input{figures/strangers}
  \caption{Point margin change per game when another method's agent plays one class of recurrent IPPO, against recurrent IPPO alone, with 95\% intervals. Within each class, strangers run from the highest-rated (top) down; the gray bar is their mean. Under each class name: the cost of a random agent. Dashed line: recurrent IPPO's own agent.}
  \label{fig:supp-strangers}
\end{figure}

Distances are in map units, between living teammates, weighted by the ticks both were alive. A stranger's distance is its mean distance to its four teammates. Its home spacing is the same measure for the same class in its own method's Season~0 games, read from the full-metrics replay of Section~\ref{sec:supp-teamwork}; how far it strays is its distance minus its home spacing. Recurrent IPPO's own agents match their home spacing within 0.03 map units. Within each class, the link between distance and cost is the rank correlation over the five learned strangers, averaged over the five classes; its exact $p$ comes from every way of shuffling the strangers within every class (Figure~\ref{fig:supp-zsc-fit}). The cost holds across opponents: 44 of the 45 changes of a learned method against a single opponent, averaged over its five classes, are below zero (Table~\ref{tab:supp-zsc-opponents}).

\paragraph{Fit across methods.} Strength alone does not decide fit: recurrent MAPPO, whose actor shares recurrent IPPO's network, inputs, reward, and memory but trained with a centralized critic, fits best in every class but the Priest, and four of the five methods' costs follow their Season~0 Elo, but PQN-VDN, second in Season~0, ties for the largest cost. Across methods, cost follows how far a stranger strays from the spacing its class keeps in its own Season~0 team: recurrent MAPPO's strangers stand 0.08 map units closer, QMIX's 0.02 farther, feedforward IPPO's 0.30, feedforward MAPPO's 0.58, and PQN-VDN's 1.37, the same order as their costs; PQN-VDN fields the tightest team in Season~0, yet its agents keep close only to their own kind.

\begin{figure}[h]
\centering
\input{supplement/zsc_fit}
\caption{Each of the 25 learned strangers: how much farther from its four teammates it stands than the agent it replaced, against its margin change, both measured from their class's mean. $\rho$: the mean of the five within-class rank correlations, with the exact $p$ of shuffling strangers within every class.}
\label{fig:supp-zsc-fit}
\end{figure}

\begin{table}[h]
\caption{Point margin per game of recurrent IPPO alone against each Season~0 member, and the change when a stranger plays one class, averaged over its five classes (200 games per opponent and team).}
\label{tab:supp-zsc-opponents}
\centering\small
\input{supplement/zsc_opponents}
\end{table}

\FloatBarrier
\section{The Expert Human's Games}
\label{supp:s10}
\label{sec:supp-expert}

Section~\ref{sec:exp-human} of the paper reports six games by the Expert Human in the BattleClient, in Agent POV, under the tournament's rules: first to 20 points, and a draw at tick 300. The BattleClient is turn-based, so the Expert Human could think as long as they liked before each move. The BattleClient also supports studies of how agents fare against players of different skill. Against recurrent IPPO, they controlled all five Team~A agents on test maps 47, 48, and 49, with Team~A starting on the left, the right, and the left (seeds 20261007--20261009). Alongside it, they controlled the Mage while recurrent IPPO controlled the other four Team~A agents, against PQN-VDN on the same maps and spawn ends (seeds 20261013--20261015). The seeds were declared before play.

Each saved game is measured with the same full metrics as any evaluation, and the analysis reads every Ultimate, hit, heal, and death from the replays tick by tick, with the game's own record of which actions each agent could take. Table~\ref{tab:expert} compares the two teams in the same three games, so each row answers ``the AI was just losing'' in one of three ways: a test against luck in the same games, each side against its own normal play, or each choice against what was possible at that moment. The three games lasted 233, 161, and 157 ticks. An ability is \emph{ready}, and an enemy \emph{in reach}, when the game's action mask allows the action: the agent is alive and not stunned, and the target is in range and in sight. The luck level for Ultimates used together slides each agent's Ultimate times by one random number of ticks within its game, 2,000 times from seed 20261016; none of the 2,000 put 89\% of the Expert Human's Ultimates together, while 77\% put at least recurrent IPPO's 57\% together. A Mage's rate on other ticks is its damage per tick on all its other living ticks, and a Burst's window is the tick it is pressed and its boosted ticks. The random pick is the share of attacks that would hit that enemy if each attacker picked evenly among the enemies it could attack, counting only attacks with at least two to choose from. For the alongside games, each game's margin is set beside recurrent IPPO's own 20 Season~0 games against PQN-VDN on the same map and spawn end.

\input{tables/expert}

{\small

%% file: supplement/table1_check.tex
A script fetched every source and searched it for its quote. Of 920 quotes, 917 were found (2 of them only after ignoring punctuation, spacing and line breaks, which PDF text extraction and web-page markup change). The other 3 could not be found this way: two pages refuse scripts (one is the error page of a dead download link, itself the evidence), and one quote sits inside a compressed StarCraft II map file. These were checked by hand. A further 2 entries record searches that found nothing, as evidence that a feature is absent. A found quote shows only that the text exists; each mark was then reviewed by hand against the rules in Section~\ref{sec:rules}.

%% file: supplement/table1_rules.tex
\begin{longtable}{>{\raggedright\arraybackslash}p{0.16\textwidth}>{\raggedright\arraybackslash}p{0.26\textwidth}>{\raggedright\arraybackslash}p{0.26\textwidth}>{\raggedright\arraybackslash}p{0.20\textwidth}}
\caption{The rule for each mark in Table~\ref{tab:comparison} of the paper, in the paper's row order.}\label{tab:rules}\\
\toprule Feature & \cmark{} yes & \pmark{} partial & \xmark{} no \\ \midrule \endfirsthead
\toprule Feature & \cmark{} yes & \pmark{} partial & \xmark{} no \\ \midrule \endhead
\bottomrule \endlastfoot
\multicolumn{4}{l}{\textit{Game}} \\
JAX / GPU-native simulator & The simulator itself runs batched on accelerators (JAX, PyTorch or CUDA). & Only part of the simulation runs on accelerators. & CPU engine. \\
\addlinespace
Classes with distinct abilities & Agent types (classes, unit types, heroes, roles) have different sets of actions or abilities, so they play differently. & Types differ only in stats (health, speed, range, damage), or distinct abilities appear only in some tasks or maps. & All agents have the same actions and stats. \\
\addlinespace
Healing, auras, crowd control, ultimates & Agents can use at least two of the four kinds, as abilities. & One kind. & None. Consumable items such as potions do not count. \\
\addlinespace
Line of sight and obstacle cover & Obstacles (walls, terrain, bushes) block sight in the standard game: a unit behind an obstacle is hidden. & Obstacles block sight only in some tasks or maps, or they block movement or attacks but not sight. & No obstacle blocks sight, movement or attacks. A vision radius alone does not count, and the arena's outer boundary is not an obstacle. \\
\addlinespace
Body blocking & An agent's body stops another agent from passing through it, including collision physics that bodies cannot pass through. & Blocking is off in some standard tasks, only some ground units or abilities block, or bodies can be pushed through or jumped over. & Agents pass through each other. \\
\addlinespace
\multicolumn{4}{l}{\textit{Evaluation}} \\
Built-in train/validation/test maps & Separate sets of maps or levels for training, validation and testing ship with the benchmark. & Some held-out maps or levels, without a full three-way split. & None. Held-out partners are not held-out maps. \\
\addlinespace
Scenarios with verified solutions & Authored scenarios ship, and every one has a supplied script, policy or documented line shown to meet its success condition. & Authored scenarios ship, but not every one has a verified solution. & None. \\
\addlinespace
Strong scripted opponents & Scripted or built-in AI opponents ship, and a source shows them stronger than at least some learned baselines. & Scripted or built-in AI opponents ship, without that evidence. & None. \\
\addlinespace
Tournament ratings with confidence intervals & Built-in tools compute ratings with uncertainty. & Ratings without uncertainty, or ratings only in a paper or competition. & None. \\
\addlinespace
Recurring official tournament & The benchmark hosts an official competition or leaderboard on a regular, recurring schedule as of September 2026. & Not used. & Otherwise, including past competitions that stopped. \\
\addlinespace
Released agent population & Official trained checkpoints are released and download on the check date. & Not used. & No official checkpoints. Training code alone, third-party checkpoints and dead links do not count. \\
\addlinespace
Who-did-what-to-whom metrics & The benchmark reports source-to-recipient statistics for at least three kinds of interaction (for example damage, healing and kills). & At least one reported statistic or recorded event names both source and recipient. & None. Raw game state and live API queries do not count. \\
\addlinespace
\multicolumn{4}{l}{\textit{Research tools}} \\
Built-in baseline learners & The official release (main or official companion repository) provides scripts to train baselines. & Only a minimal official example. & None. \\
\addlinespace
Built-in curriculum tools & A curriculum tool ships: staged training, automatic task or difficulty progression, or a curriculum API (the environment takes a schedule, stage or progress signal that changes the task as training goes on). & Difficulty or reward settings, or an ordered set of difficulty levels, that the researcher changes by hand. & None. A collection of unrelated tasks does not count. \\
\addlinespace
Built-in reward shaping tools & A researcher's own reward function is accepted and applied in the built-in training pipeline. & Fixed built-in reward options only (named modes, flags, weights). & None: changing the reward needs the researcher's own wrapper. \\
\addlinespace
Built-in competitive self-play tools & Official training code manages the opponents: it updates them, or keeps and samples past ones. & A learnable or fixed saved opponent can be loaded, but the researcher writes the update or selection loop. & Only built-in AI opponents, or cooperative only. \\
\addlinespace
Built-in cross-play tools & Tools evaluate independently trained agents or teams with or against each other (cross-play matrices, tournaments between trained agents). & Possible only with the researcher's own code, or only inside a competition server. & None. \\
\addlinespace
Built-in zero-shot coordination tools & The evaluation tools run a team made of any mix of systems, and a pool of downloadable trained agents serves as unseen partners. & Only one of the two. & Neither. \\
\addlinespace
Built-in LLM interface & An official interface lets LLM or text agents play: text observations and action parsing. & Official text observations only. & None. \\
\addlinespace
Human play client & An official way for a human to play beside or against other agents. & Limited: debugging only, the human cannot play with or against other controlled agents, or the official launcher does not run. & None. \\
\addlinespace
Replay viewer & Games are recorded and played back in a viewer. & Live rendering or video export only. & None. \\
\addlinespace
Built-in per-slot mixed teams & Learned, scripted, LLM, hybrid and human controllers can each fill any slot out of the box. & Some of these types, or only with the researcher's own code. & None. \\
\addlinespace
\end{longtable}

\noindent Notes on single rows:
\begin{itemize}
\item \textbf{Healing, auras, crowd control, ultimates.} Only mechanics an agent is the source of count: an ability that heals another agent, an aura, crowd control (stun, slow, root, silence, knockback and the like), or an ultimate (a strong ability on a long cooldown). Passive regeneration, healing at map places or pickups, and environment effects do not count.
\item \textbf{Line of sight and obstacle cover.} Line of sight and cover are not the same as a limited vision radius.
\item \textbf{Body blocking.} Marks were graded from each benchmark's movement and collision code. Units built to fly over others, such as StarCraft air units, do not lower the mark.
\item \textbf{Scenarios with verified solutions.} A scenario is a designed situation for agents that tests a named skill and has a clear success condition: a designed start, where small random jitter is allowed, and a pass or fail goal. It counts even when it is the benchmark's only task. A full open-ended match, a task scored only by its reward, a random layout, a seed choice, or a web tutorial that teaches human players is not a scenario. Yes needs every qualifying scenario to have a verified solution.
\item \textbf{Strong scripted opponents.} Sources report win rates, where a game that times out counts as not won. Yes needs a source showing the opponent stronger than some learned baselines: for example, learned baselines winning under half their games against it on at least one standard task, or failing to learn to play successfully against it.
\item \textbf{Tournament ratings with confidence intervals.} A score with confidence intervals is not a rating. Ratings are skill ratings computed from games between entrants, such as Elo or TrueSkill.
\item \textbf{Recurring official tournament.} An annual series counts as current between editions while it keeps running each year.
\item \textbf{Who-did-what-to-whom metrics.} A recorded event that names both source and recipient (for example a combat log entry) gives partial; yes needs reported statistics for at least three kinds of interaction.
\item \textbf{Built-in curriculum tools.} A curriculum API counts even when no built-in trainer uses it. A curriculum API changes the task of an existing environment from a stage or progress value during training; choosing a different preset or setting when creating the environment is partial.
\item \textbf{Built-in zero-shot coordination tools.} Both parts are required for yes: mixed-team evaluation and a downloadable pool of trained partners.
\item \textbf{Built-in reward shaping tools.} A dense reward built into a task is not a shaping tool.
\end{itemize}

%% file: supplement/table1_scope.tex
\begin{longtable}{>{\raggedright\arraybackslash}p{0.18\textwidth}>{\raggedright\arraybackslash}p{0.74\textwidth}}
\caption{What was graded for each compared benchmark. Every source link is in \texttt{table1\_evidence.csv}.}\label{tab:scope}\\
\toprule Benchmark & What was graded \\ \midrule \endfirsthead
\toprule Benchmark & What was graded \\ \midrule \endhead
\bottomrule \endlastfoot
SMAX~\cite{rutherford2024jaxmarl} & The JaxMARL paper and the official JaxMARL repository, which ships SMAX. \\
Gigastep~\cite{lechner2023gigastep} & The benchmark's NeurIPS paper and its supplementary material, and the official repository, including its third-party-learning folder, which ships the trained agent pool and the human-play script. \\
Lux S3~\cite{tao2024luxs3} & The official repository (Lux-Design-S3) and GitHub organization, Kaggle's environment package, which hosts the competition's starter bot, and the NeurIPS 2024 competition page. One competitor's public code was checked only to confirm that no official training code or agents exist. \\
VMAS~\cite{bettini2022vmas} & The benchmark's paper and official repository, and BenchMARL, the training library that the VMAS authors released as its companion. \\
MAgent2~\cite{zheng2018magent} & Farama's MAgent2, the original MAgent (geek-ai/MAgent), the MAgent paper, and its NeurIPS 2017 demo page, checked for competitions. \\
MPE~\cite{mordatch2018emergence,lowe2017maddpg} & The MADDPG paper, the original OpenAI release (openai/multiagent-particle-envs), its companion training code (openai/maddpg), and Farama's MPE2 with its PettingZoo documentation. JaxMARL's JAX version of MPE was checked only to confirm that it is a separate project. \\
SMAC~\cite{samvelyan2019smac} & The benchmark's paper, its official repository and that repository's release files; PyMARL, the training framework released with SMAC; and the SMACv2 paper. Third-party papers and checkpoints were checked only to confirm what the release lacks, and a StarCraft~II community wiki documents how the game engine blocks units. \\
SMACv2~\cite{ellis2023smacv2} & The benchmark's paper, official repository and project website, and pymarl2, the companion repository that SMACv2's official guide links (its smacv2-feature-inferrability branch holds the trained QMIX checkpoints). \\
GRF~\cite{kurach2020grf} & The benchmark's paper and official repository, Google's download links for its trained checkpoints, archived pages of the official GRF League, and the Kaggle competition page. \\
HoK Arena~\cite{wei2022hok} & The benchmark's paper and its official hok\_env release (1v1 in the paper, 3v3 in a later release); Tencent's Hokoff release and paper; Tencent's AI Arena documentation and competition pages; an earlier Tencent paper on the full game; and official Honor of Kings game pages, with one game-news article, which document how the game itself handles body blocking. \\
Mini HoK~\cite{liu2025minihok} & The benchmark's paper and official repository (tencent-ailab/marl-mini); official Honor of Kings hero and news pages, for the heroes' skills and for blocking; and a news report and Tencent's Kaiwu competition pages, checked for competitions. \\
Dota 2~\cite{berner2019dota} & Dota~2 has no released benchmark: OpenAI released game replays, but never its training code or OpenAI Five's weights. Graded instead are the OpenAI Five paper, an archived OpenAI blog post, and the game as Valve ships it: Valve's official hero data, the archived Valve Developer Community page on bot scripting, Valve's gameplay issue tracker, and a public mirror of the game's own files. Archived pages of a community wiki document how the game handles vision, unit size and bots, and two third-party projects were checked only to confirm that no official learner or agents exist. \\
Overcooked~\cite{carroll2019overcooked} & The benchmark's paper and the overcooked\_ai repository, including its deprecated human\_aware\_rl training code and its demo server. \\
OvercookedV2~\cite{gessler2025overcookedv2} & The benchmark's paper, the JaxMARL environment and the authors' experiment code (overcookedv2/experiments). \\
MP 2.0~\cite{agapiou2022meltingpot2} & The Melting Pot 2.0 paper, the meltingpot repository, the official Melting Pot Contest 2023 repository (rstrivedi/Melting-Pot-Contest-2023), which uses its own frozen Melting Pot version, and the contest's NeurIPS report. \\
Hanabi~\cite{bard2020hanabi} & The benchmark's paper and official repository (hanabi-learning-environment). A third-party paper and Meta's separate checkpoints were checked only to confirm what the release lacks. \\
\end{longtable}

%% file: supplement/atlas.tex
\begin{figure}[p]
\centering
\includegraphics[width=\textwidth]{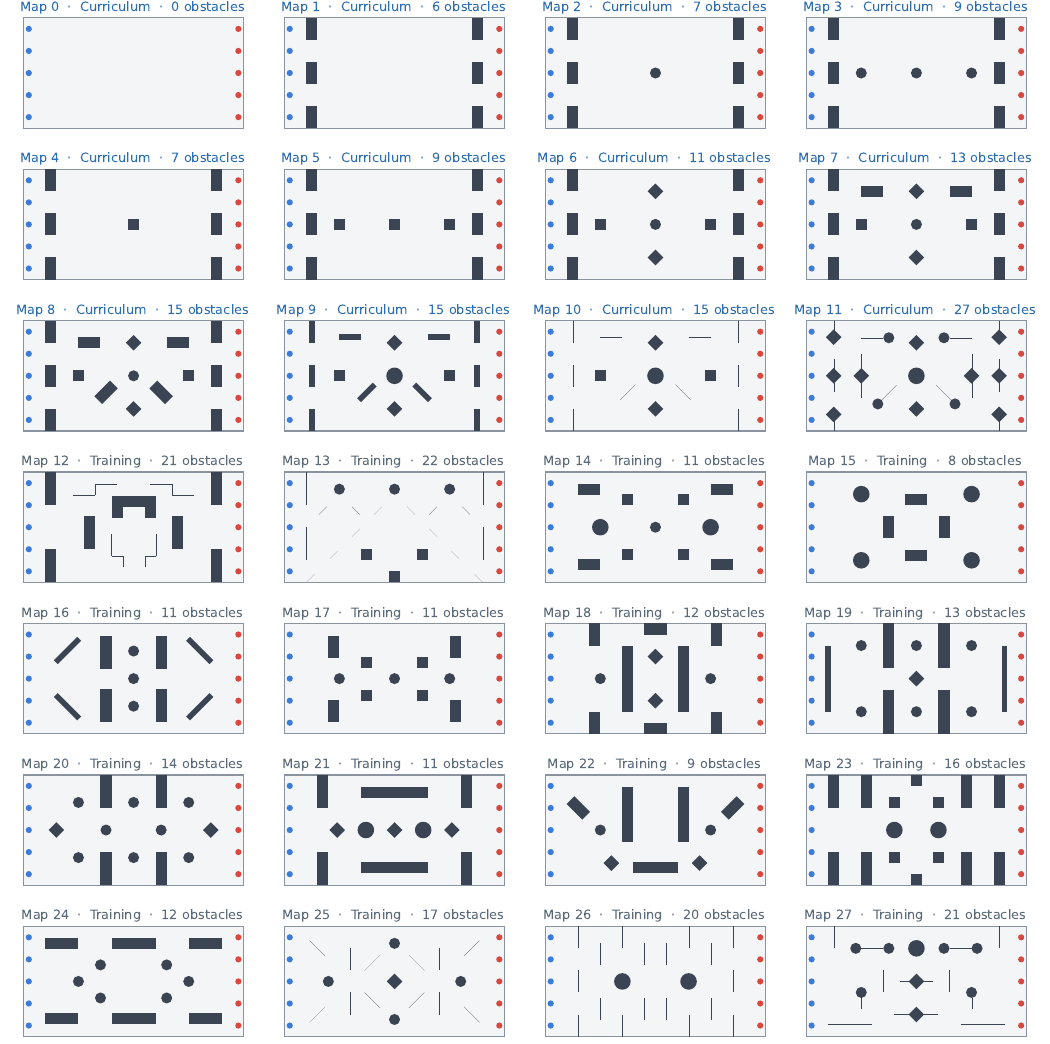}
\caption{The maps of Team Deathmatch, drawn from the map files (maps 0--27). Each title gives the map's number, its split, and its number of obstacles. Dark shapes are pillars and walls; blue and red dots are the two teams' spawn pads. Every map is 20 by 10 map units.}
\label{fig:supp-atlas}
\end{figure}
\begin{figure}[p]
\centering
\includegraphics[width=\textwidth]{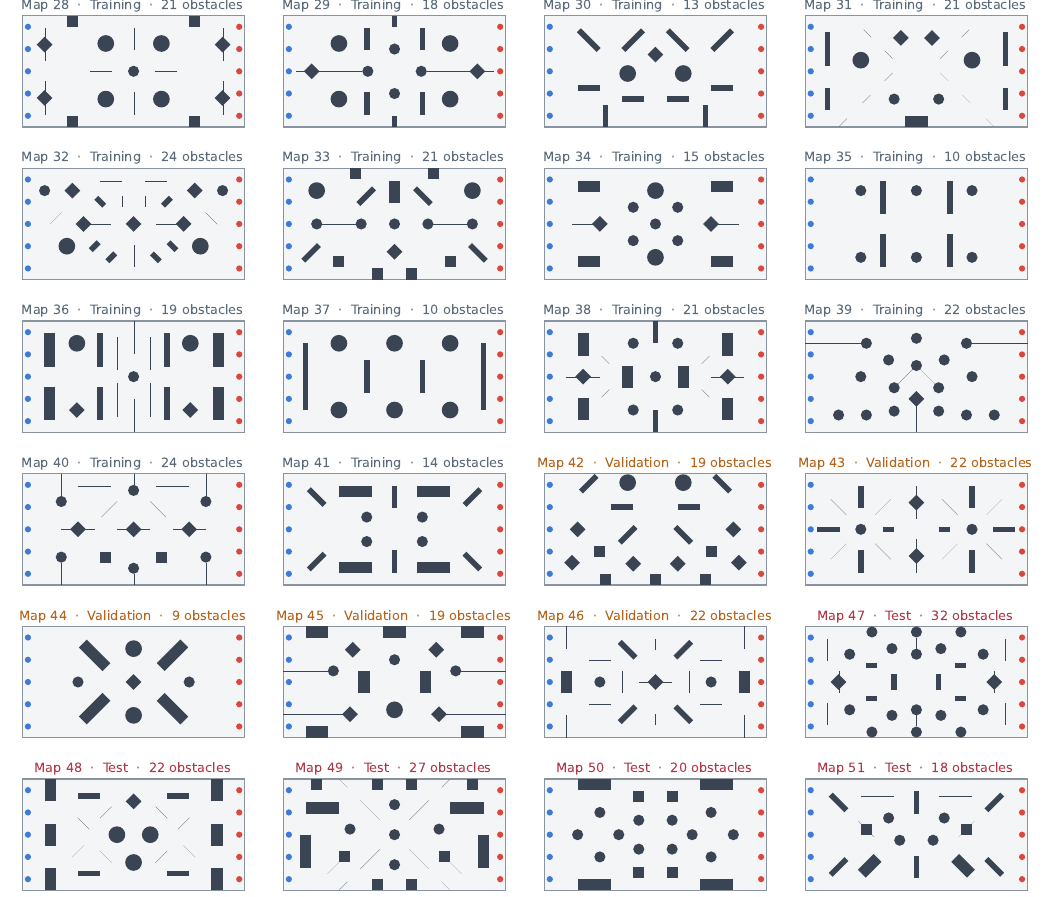}
\caption{The maps of Team Deathmatch, continued (maps 28--51).}
\label{fig:supp-atlas-2}
\end{figure}

%% file: tables/scenario_results.tex
\begin{table}[h]
\caption{Scenario results with each system as Team~A: wins--draws--losses in 256 games per scenario. Every system loses all games of Scenarios 5--8, which share a column. Wins and Win~\% cover all eight scenarios. Rec.: recurrent; FF: feedforward; Qwen3.5-9B: the LLM team (Section~7.6 of the paper).}
\label{tab:scenario-results}
\centering\footnotesize
\setlength{\tabcolsep}{2.6pt}
\begin{tabular}{@{}lcccccrr@{}}
\toprule
Team~A & S1 & S2 & S3 & S4 & S5--S8 & Wins & Win~\% \\
\midrule
Rec.~IPPO & 7--160--89 & 0--0--256 & 0--0--256 & 0--3--253 & 0--0--256 & 7 & 0.34\% \\
PQN-VDN & 0--256--0 & 0--0--256 & 0--0--256 & 0--0--256 & 0--0--256 & 0 & 0.00\% \\
Rec.~MAPPO & 3--200--53 & 0--0--256 & 0--0--256 & 0--2--254 & 0--0--256 & 3 & 0.15\% \\
QMIX & 0--0--256 & 0--0--256 & 0--0--256 & 0--0--256 & 0--0--256 & 0 & 0.00\% \\
FF~IPPO & 8--148--100 & 1--3--252 & 2--1--253 & 0--21--235 & 0--0--256 & 11 & 0.54\% \\
FF~MAPPO & 1--208--47 & 0--0--256 & 1--3--252 & 0--1--255 & 0--0--256 & 2 & 0.10\% \\
TDM-Beta & 0--256--0 & 0--0--256 & 0--0--256 & 0--0--256 & 0--0--256 & 0 & 0.00\% \\
TDM-Gamma & 0--256--0 & 0--0--256 & 0--0--256 & 0--0--256 & 0--0--256 & 0 & 0.00\% \\
TDM-Alpha & 0--256--0 & 0--0--256 & 0--0--256 & 0--0--256 & 0--0--256 & 0 & 0.00\% \\
\addlinespace
Random & 2--92--162 & 0--3--253 & 0--0--256 & 0--17--239 & 0--0--256 & 2 & 0.10\% \\
Qwen3.5-9B & 0--3--253 & 0--0--256 & 0--0--256 & 0--33--223 & 0--0--256 & 0 & 0.00\% \\
\bottomrule
\end{tabular}
\end{table}

%% file: tables/tiers.tex
\begin{table}[h]
\caption{Leaderboard tiers. Each tier below Grandmaster splits into five divisions, 40 Elo apart, numbered 5 (lowest) to 1 (highest). Win chance: the share of decisive games, those not drawn, that a policy with this Elo wins against a policy of the field's average Elo, 1,200.}
\label{tab:tiers}
\centering\footnotesize
\begin{tabular}{@{}lrr@{\hspace{1.4em}}lrr@{}}
\toprule
Tier & Elo & Win chance & Tier & Elo & Win chance \\
\midrule
Grandmaster & 2,000+ & 99.0\%+ & Gold & 1,200--1,399 & 50.0--76.0\% \\
Master & 1,800--1,999 & 96.9--99.0\% & Silver & 1,000--1,199 & 24.0--50.0\% \\
Diamond & 1,600--1,799 & 90.9--96.9\% & Bronze & 800--999 & 9.1--24.0\% \\
Platinum & 1,400--1,599 & 76.0--90.9\% & Iron & below 800 & below 9.1\% \\
\bottomrule
\end{tabular}
\end{table}

%% file: figures/curves.tex
\definecolor{crvIppo}{HTML}{0072B2}\definecolor{crvMappo}{HTML}{009E73}%
\definecolor{crvPqn}{HTML}{D55E00}\definecolor{crvQmix}{HTML}{CC79A7}%
\pgfplotstableread[col sep=comma]{figures/training_curves.csv}\curvesdata
\pgfplotsset{
  rows/.style 2 args={skip coords between index={0}{#1}, skip coords between index={#2}{186}},
  curves panel/.style={scale only axis, width=3.05cm, height=2.6cm, xmin=0, xmax=2.6e8,
    xtick={0,5e7,1e8,1.5e8,2e8,2.5e8}, xticklabels={0,50,100,150,200,250}, scaled x ticks=false,
    xlabel={Game steps trained (millions)}, grid=major, grid style={gray!20},
    tick label style={font=\tiny}, label style={font=\scriptsize}, title style={font=\scriptsize, yshift=-1.2ex},
    every axis plot/.append style={line width=0.6pt, mark size=0.9pt, mark repeat=3,
      mark options={solid, line width=0.4pt}}},
  rec ippo/.style={crvIppo, mark=*}, ff ippo/.style={crvIppo, dashed, mark=o},
  rec mappo/.style={crvMappo, mark=square*}, ff mappo/.style={crvMappo, dashed, mark=square},
  pqn vdn/.style={crvPqn, mark=diamond*}, qmix/.style={crvQmix, mark=triangle*},
}
\newcommand{\curvesplots}[1]{%
  \addplot[rec ippo, forget plot, rows={0}{31}] table[x=env_steps, y=#1] {\curvesdata};
  \addplot[pqn vdn, forget plot, rows={31}{62}] table[x=env_steps, y=#1] {\curvesdata};
  \addplot[rec mappo, forget plot, rows={62}{93}] table[x=env_steps, y=#1] {\curvesdata};
  \addplot[qmix, forget plot, rows={93}{124}] table[x=env_steps, y=#1] {\curvesdata};
  \addplot[ff ippo, forget plot, rows={124}{155}] table[x=env_steps, y=#1] {\curvesdata};
  \addplot[ff mappo, forget plot, rows={155}{186}] table[x=env_steps, y=#1] {\curvesdata};
  \draw[gray, densely dotted] ({axis cs:40108032,0}|-{rel axis cs:0,0}) -- ({axis cs:40108032,0}|-{rel axis cs:0,1});}
\begin{tikzpicture}
\begin{axis}[curves panel, name=margin, title={Point margin per game}, ymin=-21, ymax=9]
  \curvesplots{margin}
  \draw[black!70, line width=0.4pt] ({rel axis cs:0,0}|-{axis cs:0,0}) -- ({rel axis cs:1,0}|-{axis cs:0,0});
\end{axis}
\begin{axis}[curves panel, name=kills, at={(margin.east)}, anchor=west, xshift=0.62cm, title={Kills per game},
  ymin=0, ymax=21, legend to name=curveslegend, legend columns=3, legend cell align=left,
  legend style={font=\tiny, draw=none, /tikz/every even column/.append style={column sep=6pt}}]
  \curvesplots{kills}
  \addlegendimage{rec ippo}\addlegendentry{Rec.\ IPPO}
  \addlegendimage{pqn vdn}\addlegendentry{PQN-VDN}
  \addlegendimage{rec mappo}\addlegendentry{Rec.\ MAPPO}
  \addlegendimage{qmix}\addlegendentry{QMIX}
  \addlegendimage{ff ippo}\addlegendentry{FF IPPO}
  \addlegendimage{ff mappo}\addlegendentry{FF MAPPO}
\end{axis}
\path (margin.outer south east) -- (kills.outer south west)
  node[midway, below=-1pt] {\pgfplotslegendfromname{curveslegend}};
\end{tikzpicture}

%% file: figures/curves_wins.tex
\definecolor{crvIppo}{HTML}{0072B2}\definecolor{crvMappo}{HTML}{009E73}%
\definecolor{crvPqn}{HTML}{D55E00}\definecolor{crvQmix}{HTML}{CC79A7}%
\pgfplotstableread[col sep=comma]{figures/training_curves_scripted.csv}\scripteddata
\pgfplotsset{
  rows/.style 2 args={skip coords between index={0}{#1}, skip coords between index={#2}{186}},
  wins panel/.style={scale only axis, width=11cm, height=4cm, xmin=0, xmax=2.6e8,
    xtick={0,5e7,1e8,1.5e8,2e8,2.5e8}, xticklabels={0,50,100,150,200,250}, scaled x ticks=false,
    xlabel={Game steps trained (millions)}, ylabel={Games won against scripted opponents (\%)},
    grid=major, grid style={gray!20}, tick label style={font=\scriptsize}, label style={font=\small},
    every axis plot/.append style={line width=0.6pt, mark size=1pt, mark repeat=3,
      mark options={solid, line width=0.4pt}}},
  rec ippo/.style={crvIppo, mark=*}, ff ippo/.style={crvIppo, dashed, mark=o},
  rec mappo/.style={crvMappo, mark=square*}, ff mappo/.style={crvMappo, dashed, mark=square},
  pqn vdn/.style={crvPqn, mark=diamond*}, qmix/.style={crvQmix, mark=triangle*},
}
\begin{tikzpicture}
\begin{axis}[wins panel, ymin=0, ymax=102, ytick={0,25,50,75,100},
  legend columns=6, legend cell align=left, legend to name=winslegend,
  legend style={font=\scriptsize, draw=none, /tikz/every even column/.append style={column sep=8pt}}]
  \addplot[rec ippo, rows={0}{31}] table[x=env_steps, y expr=\thisrow{win_share}*100] {\scripteddata};
  \addlegendentry{Rec.\ IPPO}
  \addplot[pqn vdn, rows={31}{62}] table[x=env_steps, y expr=\thisrow{win_share}*100] {\scripteddata};
  \addlegendentry{PQN-VDN}
  \addplot[rec mappo, rows={62}{93}] table[x=env_steps, y expr=\thisrow{win_share}*100] {\scripteddata};
  \addlegendentry{Rec.\ MAPPO}
  \addplot[qmix, rows={93}{124}] table[x=env_steps, y expr=\thisrow{win_share}*100] {\scripteddata};
  \addlegendentry{QMIX}
  \addplot[ff ippo, rows={124}{155}] table[x=env_steps, y expr=\thisrow{win_share}*100] {\scripteddata};
  \addlegendentry{FF IPPO}
  \addplot[ff mappo, rows={155}{186}] table[x=env_steps, y expr=\thisrow{win_share}*100] {\scripteddata};
  \addlegendentry{FF MAPPO}
  \draw[black!70, line width=0.4pt] ({rel axis cs:0,0}|-{axis cs:0,50}) -- ({rel axis cs:1,0}|-{axis cs:0,50});
  \draw[gray, densely dotted] ({axis cs:40108032,0}|-{rel axis cs:0,0}) -- ({axis cs:40108032,0}|-{rel axis cs:0,1});
\end{axis}
\end{tikzpicture}\\[2pt]
\pgfplotslegendfromname{winslegend}

%% file: supplement/teamwork_tests.tex
\begin{tabular}{@{}llrrrr@{}}
\toprule
 & Measure & $\rho$ & $p$ & Holm & Max-$t$ \\
\midrule
Formation & Teammate distance & $-0.90$ & $0.002$ & 0.27 & 0.12 \\
 & Time inside a Mage aura & $+0.88$ & $0.003$ & 0.39 & 0.17 \\
 & Time inside a Warrior aura & $+0.87$ & $0.005$ & 0.55 & 0.23 \\
\addlinespace[2pt]
Aura payoff & Damage Mage auras added* & $+0.92$ & $0.001$ & 0.18 & 0.087 \\
\addlinespace[2pt]
Support & Useful share of Basic healing & $+0.90$ & $0.002$ & 0.27 & 0.12 \\
\addlinespace[2pt]
Setups & Damage to Hunter-slowed enemies* & $+0.88$ & $0.003$ & 0.39 & 0.17 \\
 & Kills with two or more helpers & $+0.85$ & $0.006$ & 0.71 & 0.28 \\
 & Ticks with two or more attackers* & $-0.02$ & $0.982$ & 1 & 1 \\
\addlinespace[2pt]
Abilities & Share of kills a Warrior Ultimate helped & $-0.95$ & $3.5\times 10^{-4}$ & \textbf{0.049} & \textbf{0.029} \\
 & Ultimates used* & $-0.13$ & $0.744$ & 1 & 1 \\
 & Basic abilities used* & $-0.17$ & $0.678$ & 1 & 1 \\
 & Kills a Warrior's Basic helped* & $+0.37$ & $0.336$ & 1 & 1 \\
\addlinespace[2pt]
Results & Kills* & $+0.02$ & $0.982$ & 1 & 1 \\
 & Deaths* & $-0.83$ & $0.008$ & 0.89 & 0.35 \\
\bottomrule
\end{tabular}

%% file: figures/strangers.tex
\definecolor{zIppo}{HTML}{0072B2}\definecolor{zMappo}{HTML}{009E73}%
\definecolor{zPqn}{HTML}{D55E00}\definecolor{zQmix}{HTML}{CC79A7}%
\begin{tikzpicture}
\begin{axis}[set layers, width=\linewidth, height=5.4cm, xmin=-8, xmax=0.6, ymin=-0.55, ymax=4.55, y dir=reverse,
  ytick={0,1,2,3,4}, yticklabels={{Mage\\[-1pt]{\tiny\color{black!55}random $-16.9$}},{Warrior\\[-1pt]{\tiny\color{black!55}random $-8.2$}},{Hunter\\[-1pt]{\tiny\color{black!55}random $-7.9$}},{Rogue\\[-1pt]{\tiny\color{black!55}random $-5.0$}},{Priest\\[-1pt]{\tiny\color{black!55}random $-14.5$}}}, yticklabel style={align=right}, ytick style={draw=none},
  xtick={-8,-6,-4,-2,0}, xmajorgrids, grid style={gray!15},
  xlabel={Point margin change per game, against recurrent IPPO alone},
  label style={font=\scriptsize}, tick label style={font=\scriptsize},
  legend style={font=\tiny, at={(0.5,1.02)}, anchor=south, legend columns=3, draw=none,
    /tikz/every even column/.append style={column sep=4pt}}, legend cell align=left]
  \draw[black!45, dashed] (axis cs:0,-0.55) -- (axis cs:0,4.55);
  \draw[black!55, line width=0.4pt] ({rel axis cs:0,0}|-{axis cs:0,0.5}) -- ({rel axis cs:1,0}|-{axis cs:0,0.5});
  \draw[black!55, line width=0.4pt] ({rel axis cs:0,0}|-{axis cs:0,1.5}) -- ({rel axis cs:1,0}|-{axis cs:0,1.5});
  \draw[black!55, line width=0.4pt] ({rel axis cs:0,0}|-{axis cs:0,2.5}) -- ({rel axis cs:1,0}|-{axis cs:0,2.5});
  \draw[black!55, line width=0.4pt] ({rel axis cs:0,0}|-{axis cs:0,3.5}) -- ({rel axis cs:1,0}|-{axis cs:0,3.5});
  \addplot[black!38, line width=1.6pt, forget plot, on layer=axis background] coordinates {(-4.782,-0.40) (-4.782,0.40)};
  \addplot[black!38, line width=1.6pt, forget plot, on layer=axis background] coordinates {(-2.529,0.60) (-2.529,1.40)};
  \addplot[black!38, line width=1.6pt, forget plot, on layer=axis background] coordinates {(-2.323,1.60) (-2.323,2.40)};
  \addplot[black!38, line width=1.6pt, forget plot, on layer=axis background] coordinates {(-1.620,2.60) (-1.620,3.40)};
  \addplot[black!38, line width=1.6pt, forget plot, on layer=axis background] coordinates {(-3.411,3.60) (-3.411,4.40)};
  \addplot[only marks, mark=diamond*, mark size=1.7pt, zPqn, mark options={fill=zPqn, line width=0.6pt},
    error bars/.cd, x dir=both, x explicit, error bar style={zPqn, line width=0.5pt}] coordinates {
    (-6.763,-0.260) -= (0.481,0) += (0.483,0)
    (-3.752,0.740) -= (0.419,0) += (0.445,0)
    (-3.549,1.740) -= (0.391,0) += (0.397,0)
    (-2.041,2.740) -= (0.318,0) += (0.342,0)
    (-4.437,3.740) -= (0.451,0) += (0.491,0)
  };
  \addlegendentry{PQN-VDN (1,358 Elo)}
  \addplot[only marks, mark=square*, mark size=1.7pt, zMappo, mark options={fill=zMappo, line width=0.6pt},
    error bars/.cd, x dir=both, x explicit, error bar style={zMappo, line width=0.5pt}] coordinates {
    (-1.542,-0.130) -= (0.332,0) += (0.333,0)
    (-0.879,0.870) -= (0.327,0) += (0.329,0)
    (-0.761,1.870) -= (0.311,0) += (0.327,0)
    (-0.219,2.870) -= (0.306,0) += (0.304,0)
    (-2.431,3.870) -= (0.316,0) += (0.311,0)
  };
  \addlegendentry{Rec.\ MAPPO (1,304 Elo)}
  \addplot[only marks, mark=triangle*, mark size=1.7pt, zQmix, mark options={fill=zQmix, line width=0.6pt},
    error bars/.cd, x dir=both, x explicit, error bar style={zQmix, line width=0.5pt}] coordinates {
    (-3.261,0.000) -= (0.310,0) += (0.308,0)
    (-1.404,1.000) -= (0.316,0) += (0.322,0)
    (-2.298,2.000) -= (0.282,0) += (0.283,0)
    (-1.942,3.000) -= (0.339,0) += (0.328,0)
    (-1.451,4.000) -= (0.302,0) += (0.307,0)
  };
  \addlegendentry{QMIX (1,300 Elo)}
  \addplot[only marks, mark=o, mark size=1.7pt, zIppo, mark options={fill=white, line width=0.6pt},
    error bars/.cd, x dir=both, x explicit, error bar style={zIppo, line width=0.5pt}] coordinates {
    (-5.184,0.130) -= (0.345,0) += (0.337,0)
    (-3.256,1.130) -= (0.329,0) += (0.357,0)
    (-2.125,2.130) -= (0.301,0) += (0.292,0)
    (-1.642,3.130) -= (0.292,0) += (0.299,0)
    (-3.965,4.130) -= (0.323,0) += (0.327,0)
  };
  \addlegendentry{FF IPPO (1,240 Elo)}
  \addplot[only marks, mark=square, mark size=1.7pt, zMappo, mark options={fill=white, line width=0.6pt},
    error bars/.cd, x dir=both, x explicit, error bar style={zMappo, line width=0.5pt}] coordinates {
    (-7.162,0.260) -= (0.418,0) += (0.401,0)
    (-3.354,1.260) -= (0.347,0) += (0.351,0)
    (-2.881,2.260) -= (0.324,0) += (0.331,0)
    (-2.253,3.260) -= (0.336,0) += (0.329,0)
    (-4.772,4.260) -= (0.382,0) += (0.346,0)
  };
  \addlegendentry{FF MAPPO (1,140 Elo)}
  \addlegendimage{only marks, mark=|, mark size=3.6pt, black!38, mark options={line width=1.6pt}}
  \addlegendentry{Mean of the five}
\end{axis}
\end{tikzpicture}

%% file: supplement/zsc_fit.tex
\begin{tikzpicture}
\begin{axis}[width=\linewidth, height=5.2cm, xmin=-0.8, xmax=0.9, ymin=-3.5, ymax=3.5,
  xlabel={Distance added, minus its class's mean (map units)}, ylabel={Margin change, minus its class's mean},
  label style={font=\scriptsize}, tick label style={font=\tiny},
  title={Within class: mean $\rho = -0.84$, exact $p = 2 \times 10^{-5}$}, title style={font=\scriptsize},
  legend style={font=\tiny, at={(0.98,0.98)}, anchor=north east, draw=none}, legend cell align=left]
  \addplot[only marks, mark=*, mark size=1.6pt, black] coordinates {(-0.434,3.240) (-0.365,1.650) (-0.580,1.562) (-0.152,1.401) (-0.353,0.980)};
  \addlegendentry{Rec.\ MAPPO}
  \addplot[only marks, mark=square*, mark size=1.6pt, black] coordinates {(-0.279,1.522) (-0.145,1.125) (-0.311,0.025) (-0.337,-0.323) (-0.434,1.960)};
  \addlegendentry{QMIX}
  \addplot[only marks, mark=triangle*, mark size=1.6pt, black] coordinates {(-0.129,-0.402) (-0.126,-0.726) (-0.267,0.198) (-0.327,-0.023) (-0.080,-0.554)};
  \addlegendentry{FF IPPO}
  \addplot[only marks, mark=diamond*, mark size=1.6pt, black] coordinates {(0.670,-2.380) (0.546,-0.825) (0.592,-0.558) (0.065,-0.634) (0.710,-1.361)};
  \addlegendentry{FF MAPPO}
  \addplot[only marks, mark=x, mark size=2pt, black] coordinates {(0.173,-1.981) (0.090,-1.223) (0.565,-1.226) (0.751,-0.422) (0.157,-1.026)};
  \addlegendentry{PQN-VDN}
\end{axis}
\end{tikzpicture}

%% file: supplement/zsc_opponents.tex
\begin{tabular}{@{}lrrrrrrrrr@{}}
\toprule
 & \rotatebox{60}{Recurrent IPPO} & \rotatebox{60}{PQN-VDN} & \rotatebox{60}{Rec.\ MAPPO} & \rotatebox{60}{QMIX} & \rotatebox{60}{FF IPPO} & \rotatebox{60}{FF MAPPO} & \rotatebox{60}{TDM-Beta} & \rotatebox{60}{TDM-Gamma} & \rotatebox{60}{TDM-Alpha} \\
\midrule
Rec.\ IPPO alone & $+0.1$ & $+3.2$ & $+0.6$ & $+2.6$ & $+2.9$ & $+7.0$ & $+8.0$ & $+7.6$ & $+7.7$ \\
\addlinespace
Rec.\ MAPPO & $-1.2$ & $-0.4$ & $-0.4$ & $-1.3$ & $+0.1$ & $-0.3$ & $-2.2$ & $-2.7$ & $-2.2$ \\
QMIX & $-2.4$ & $-2.3$ & $-1.6$ & $-1.7$ & $-3.4$ & $-2.2$ & $-1.8$ & $-1.9$ & $-1.3$ \\
FF IPPO & $-3.1$ & $-1.7$ & $-3.5$ & $-2.5$ & $-3.2$ & $-2.5$ & $-4.4$ & $-4.5$ & $-3.7$ \\
FF MAPPO & $-3.9$ & $-5.7$ & $-4.6$ & $-4.1$ & $-6.3$ & $-1.8$ & $-3.7$ & $-3.6$ & $-3.0$ \\
PQN-VDN & $-3.7$ & $-2.9$ & $-3.4$ & $-3.2$ & $-3.8$ & $-2.9$ & $-5.7$ & $-6.1$ & $-5.3$ \\
Random & $-9.7$ & $-8.4$ & $-10.7$ & $-9.8$ & $-9.1$ & $-7.7$ & $-13.0$ & $-13.3$ & $-12.7$ \\
\bottomrule
\end{tabular}

%% file: tables/expert.tex
\begin{table}[t]
  \caption{What the Expert Human, controlling all five agents, did that recurrent IPPO did not, in their three games on maps 47--49. Counts are summed over the games, and every row points the same way in each game. \emph{Ready} and \emph{in reach} mean the game allowed the action on that tick. Last two rows: brackets give the share under a random pick.}
  \label{tab:expert}
  \centering\footnotesize
  \setlength{\tabcolsep}{2.6pt}
  \begin{tabular}{@{}lrr@{}}
  \toprule
  Measure & Expert Human & Rec.~IPPO \\
  \midrule
  Ultimates used the first tick ready & 25\% (16/63) & 51\% (34/67) \\
  Ticks a ready Charge waited (median) & 3 & 0 \\
  Damage of the opening Burst & 76 to 95 & 0 to 22 \\
  \addlinespace
  Bursts with an enemy in reach & 14/15 & 0/16 \\
  Mage damage in Burst vs.\ other ticks & 2.9$\times$ & 0.8$\times$ \\
  \addlinespace
  Ultimates within 4 ticks of an ally's (luck: ${\approx}$60\%) & 89\% (56/63) & 57\% (38/67) \\
  Mage Bursts within 4 ticks of a Charge & 15/15 & 3/16 \\
  Allies with their Ultimate also ready (mean) & 1.2 & 0.6 \\
  \addlinespace
  Basic heals to a hurt ally, when one was in reach & 94\% & 81\% \\
  Allies at 35\% health or less healed in 2 ticks & 74\% (32/43) & 49\% (23/47) \\
  \addlinespace
  Attacks on the enemy Priest in reach & 40\% (37\%) & 5\% (32\%) \\
  Attacks on the enemy Rogue in reach & 41\% (39\%) & 88\% (37\%) \\
  \bottomrule
  \end{tabular}
\end{table}